%% file: root.tex
\documentclass[lettersize,journal]{IEEEtran}
\usepackage{amsmath,amsfonts}
\usepackage{algorithmic}
\usepackage{algorithm}
\usepackage{array}
\usepackage[caption=false,font=normalsize,labelfont=sf,textfont=sf]{subfig}
\usepackage{textcomp}
\usepackage{stfloats}
\usepackage{url}
\usepackage{verbatim}
\usepackage{graphicx}
\usepackage{cite}
\usepackage{xcolor}

\usepackage{subcaption}
\usepackage{amsmath}
\usepackage[hidelinks]{hyperref}
\usepackage{cleveref}
\usepackage{booktabs}
\usepackage{makecell}
\usepackage{multirow}
\usepackage{arydshln}
\usepackage{pifont}
\newcommand{\cmark}{\ding{51}}  

\begin{document}

\title{RoEL: Robust Event-based 3D Line Reconstruction}

\author{Gwangtak Bae, Jaeho Shin, Seunggu Kang, Junho Kim, Ayoung Kim,~\IEEEmembership{Senior Member,~IEEE}, \\ and Young Min Kim,~\IEEEmembership{Member,~IEEE}

\thanks{This work was partly supported by Institute of Information \& communications Technology Planning \& Evaluation (IITP) grant funded by the Korea government(MSIT) [Artificial Intelligence Graduate School Program (Seoul National University)], Creative-Pioneering Researchers Program through Seoul National University, and the BK21 FOUR program of the Education and Research Program for Future ICT Pioneers, Seoul National University in 2026. \textit{(Corresponding author: Young Min Kim)}}
\thanks{G. Bae, J. Kim, and Y. Kim are with the Dept. of Electrical and Computer Engineering, Seoul National University (\texttt{tak3452@snu.ac.kr}, \texttt{82magnolia@snu.ac.kr}, \texttt{youngmin.kim@snu.ac.kr}).}%
\thanks{J. Shin was with the Seoul National University during the course of this research. He is now with the University of Michigan (\texttt{jaehos@umich.edu}).}%
\thanks{S. Kang is with the Dept. of Future Automotive Mobility, Seoul National University (\texttt{jrkangsg@snu.ac.kr}).}%
\thanks{A. Kim is with the Dept. of Mechanical Engineering, Seoul National University (\texttt{ayoungk@snu.ac.kr}).}
}

\markboth{IEEE TRANSACTIONS ON ROBOTICS, PREPRINT VERSION, ACCEPTED, FEB 2026}%
{Shell \MakeLowercase{\textit{et al.}}: A Sample Article Using IEEEtran.cls for IEEE Journals}

\maketitle

\input{sec/00_abstract}    


\input{sec/01_intro}
\input{sec/02_related_works}
\input{sec/03_method}

\input{sec/04_experiments}

\input{sec/05_analysis}
\input{sec/06_limitation}
\input{sec/07_conclusion}

\bibliographystyle{IEEEtran}
\bibliography{references}

\newpage

\begin{IEEEbiography}[{\includegraphics[width=1in,height=1.25in,clip,keepaspectratio]{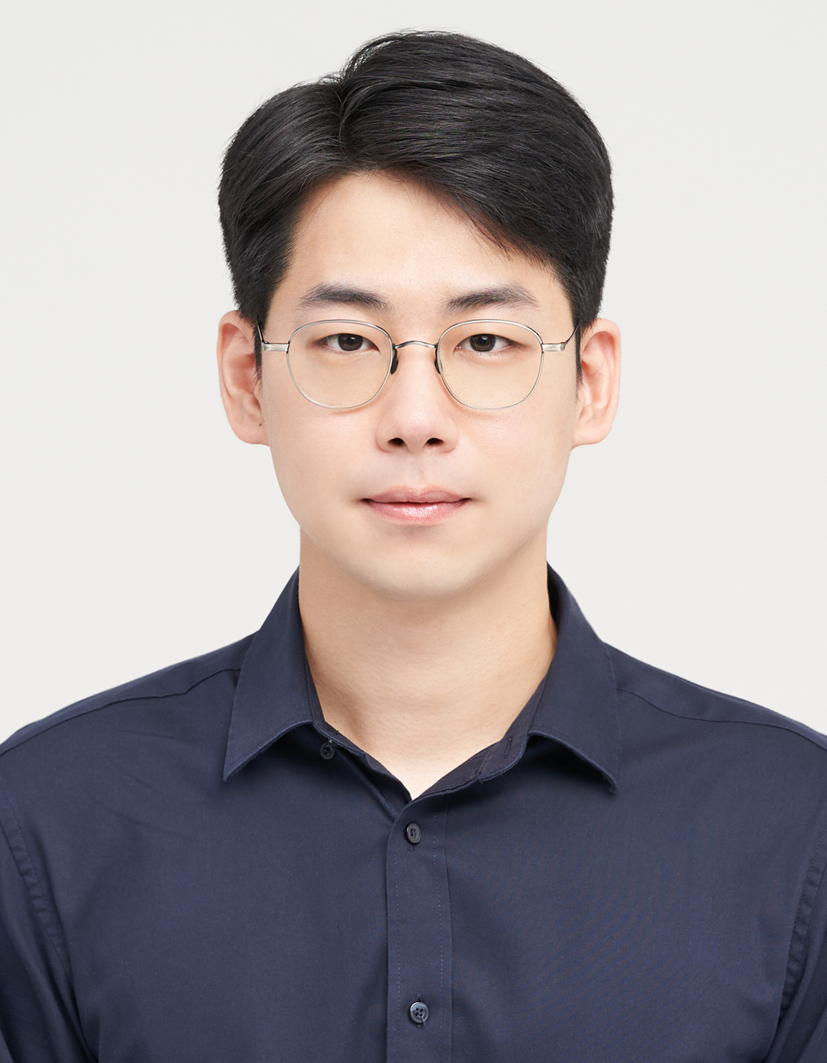}}]{Gwangtak Bae}%
received the B.S. degree in Electrical Engineering from the Korea Advanced Institute of Science and Technology (KAIST) in 2019. He was a Research Officer at the Agency for Defense Development (ADD). He is currently a Ph.D. candidate in the Department of Electrical and Computer Engineering at Seoul National University (SNU). His research interests include 3D reconstruction, localization, and event-based vision, focusing on robust computer vision methods for robotic applications.
\end{IEEEbiography}%

\begin{IEEEbiography}[{\includegraphics[width=1in,height=1.25in,clip,keepaspectratio]{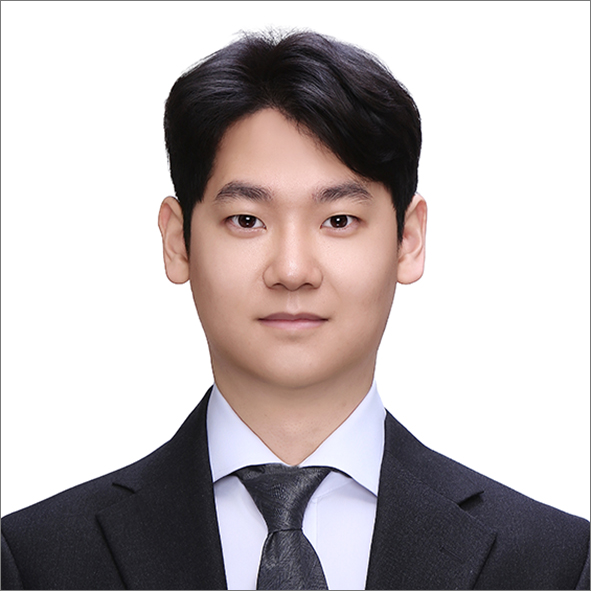}}]{Jaeho Shin}%
received B.S. and M.S. degrees in mechanical engineering from Seoul National University (SNU) in 2023 and 2025. He is currently a Ph.D. student in robotics department at University of Michigan. His research interest lies in Simultaneous Localization and Mapping (SLAM), manifold and optimization, which involve scalable and trustworthy state estimation for robotic application. \end{IEEEbiography}%

\begin{IEEEbiography}
[{\includegraphics[width=1in,height=1.25in,clip,keepaspectratio]{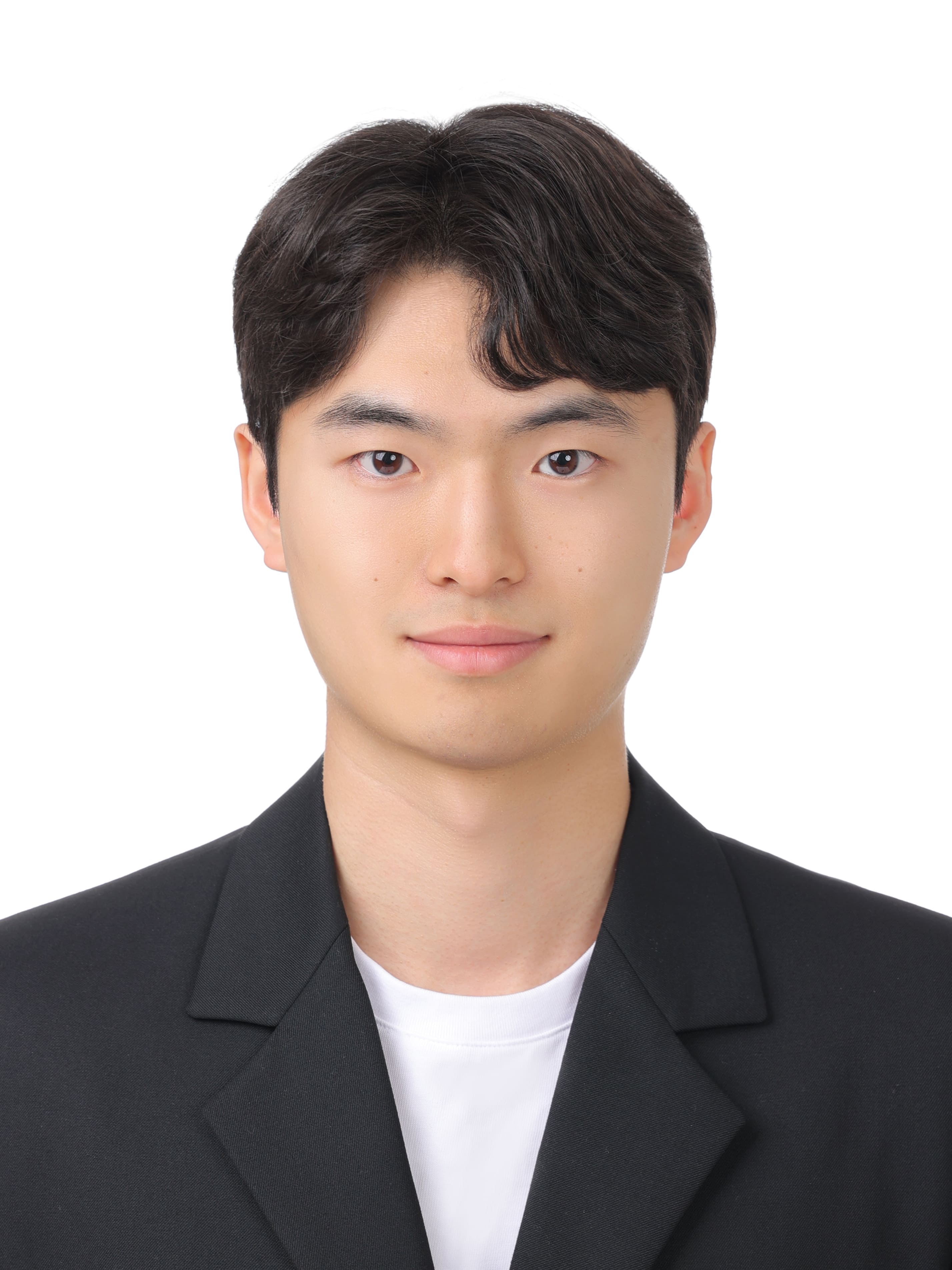}}]{Seunggu Kang}%
received the B.S. degree in mechanical engineering from Hanyang University in 2025. He is currently a M.S. candidate in the Department of Future Automotive Mobility at Seoul National University (SNU). His research interests include 3D geometric reconstruction, scene understanding, and HD map construction for autonomous driving systems.
\end{IEEEbiography}

\begin{IEEEbiography}[{\includegraphics[width=1in,height=1.25in,clip,keepaspectratio]{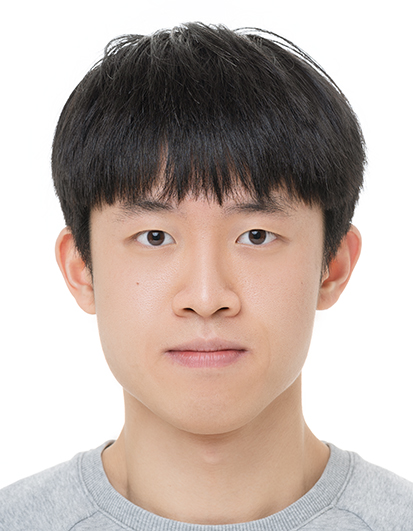}}]{Junho Kim} received the bachelor's degree from Seoul National University and the Ph.D. degree from the same university in 2025.
He is currently a postdoctoral scholar at Seoul National University.
He has published papers at prestigious venues including CVPR, ICCV, and ECCV.
His research interests include 3D reconstruction, scene understanding, and event-based vision.
\end{IEEEbiography}

\begin{IEEEbiography}[{\includegraphics[width=1in,height=1.25in,clip,keepaspectratio]{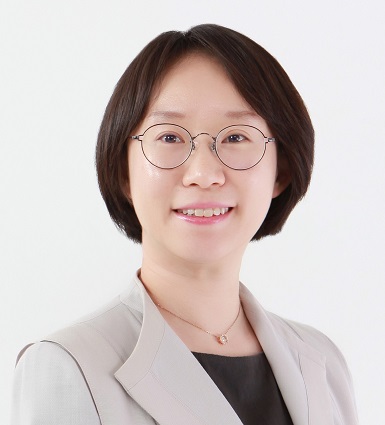}}]{Ayoung Kim}%
(S'08--M'13--S'23) received B.S. and M.S. degrees in mechanical engineering from SNU in 2005 and 2007, and an M.S. degree in electrical engineering and a Ph.D. degree in mechanical engineering from the University of Michigan (UM), Ann Arbor, in 2011 and 2012. 
She was an associate professor at Korea Advanced Institute of Science and Technology (KAIST) from 2014 to 2021. Currently, she is an associate professor at SNU.
\end{IEEEbiography}%

\begin{IEEEbiography}[{\includegraphics[width=1in,height=1.25in,clip,keepaspectratio]{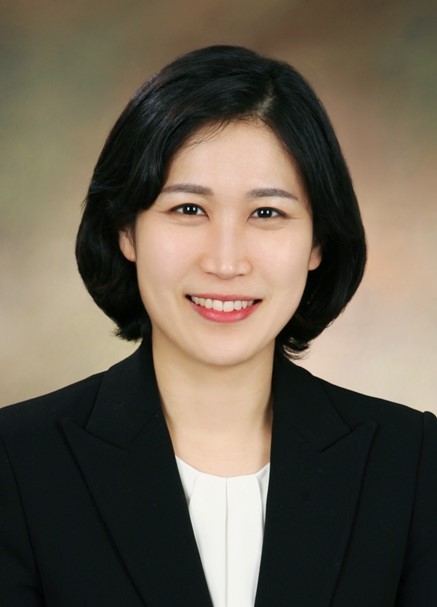}}]{Young Min Kim}%
is an Associate Professor in the Department of Electrical and Computer Engineering at Seoul National University, Seoul, Korea, where she is leading a 3D vision lab. She received a B.S. from Seoul National University in 2006 and an M.S. and Ph.D. in electrical engineering from Stanford University in 2008 and 2013, respectively. Before joining SNU, she was a Senior Research Scientist at the Korea Institute of Science and Technology (KIST). She serves as an area chair in CVPR, ICCV, ECCV, ACCV, program committee for Pacific Graphics, AAAI, and technical papers committee for SIGGRAPH and SIGGRAPH Asia. She is also a program chair for 3DV 2026. Her research interest lies in 3D vision, where she combines computer vision, graphics, and robotics algorithms to solve practical problems. 
\end{IEEEbiography}%

\vfill

\end{document}

%% file: sec/00_abstract.tex
\begin{abstract}
Event cameras in motion tend to detect object boundaries or texture edges, which produce lines of brightness changes, especially in man-made environments.
While lines can constitute a robust intermediate representation that is consistently observed, the sparse nature of lines may lead to drastic deterioration with minor estimation errors.
Only a few previous works, often accompanied by additional sensors, utilize lines to compensate for the severe domain discrepancies of event sensors along with unpredictable noise characteristics.
We propose a method that can stably extract tracks of varying appearances of lines using a clever algorithmic process that observes multiple representations from various time slices of events, compensating for potential adversaries within the event data.
We then propose geometric cost functions that can refine the 3D line maps and camera poses, eliminating projective distortions and depth ambiguities. 
The 3D line maps are highly compact and can be equipped with our proposed cost function, which can be adapted for any observations that can detect and extract line structures or projections of them, including 3D point cloud maps or image observations.
We demonstrate that our formulation is powerful enough to exhibit a significant performance boost in event-based mapping and pose refinement across diverse datasets, and can be flexibly applied to multimodal scenarios.
Our results confirm that the proposed line-based formulation is a robust and effective approach for the practical deployment of event-based perceptual modules. Project page: 
\href{https://gwangtak.github.io/roel/}{https://gwangtak.github.io/roel/}
\end{abstract}

%% file: sec/01_intro.tex
\section{Introduction}

\IEEEPARstart{E}{vent} cameras provide a power-efficient input sensor modality that detects brightness differences with microsecond temporal resolution and a high dynamic range.
Despite the prominent potential advantages of mobile and extreme setups, the effectiveness of event cameras has not yet been demonstrated for downstream vision-based tasks to be widely deployed in practical scenarios.
The high sensitivity of the sensor comes at the cost of measurement and noise characteristics that vary significantly depending on lighting conditions or sensor movements, and can produce diverse or even no outputs when observing the same 3D structure, which further challenges the development of stable algorithms with consistent performance.
Along with the complex measurement characteristics, the output formats differ from those of conventional frame-based cameras, requiring a new set of algorithms rather than benefiting from sophisticated network architectures that utilize abundant datasets from recent advancements in computer vision. 
Na\"ive application of methods developed for frame-based cameras to event data often results in inferior performance~\cite{cao2024embracing, wang2023visevent}.

\begin{figure}
    \centering
    \includegraphics[width=\linewidth]{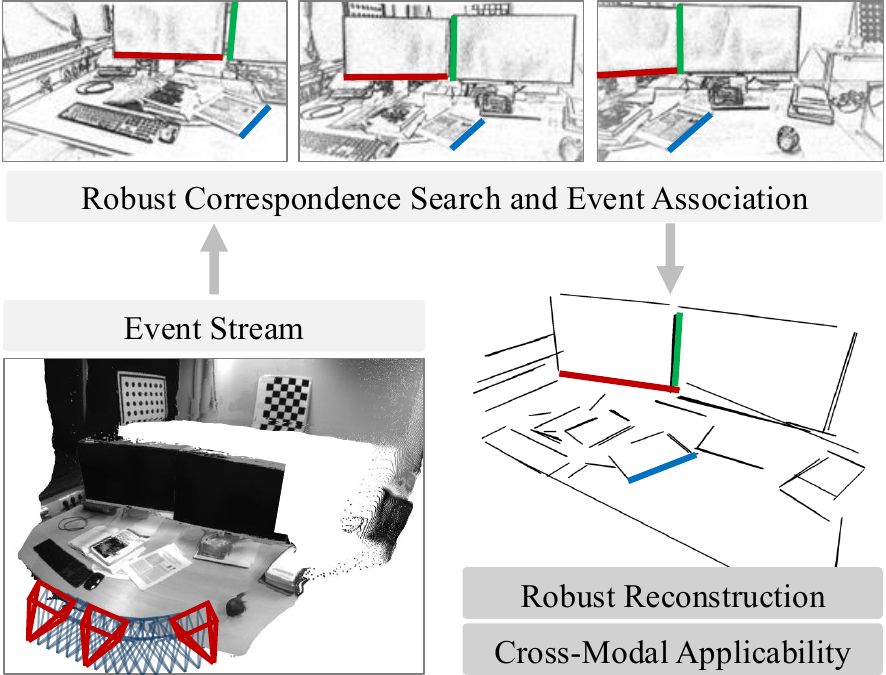}
    \caption{We present RoEL, an event-based 3D line reconstruction pipeline that achieves noise-robust reconstruction by leveraging line correspondences. Our 3D line maps not only provide efficient representations for edge-capturing event cameras, but also enable cross-modal applications such as registration and localization.}
    \label{fig:teaser}
\end{figure}

In this work, we present an event-specific method that construct a sparse and coherent 3D line structure of the indoor environment while simultaneously refining the sensor poses.
Man-made environments often contain a rich number of structural lines, which can be consistently extracted across varying viewpoints and motion directions.
Because of the prevalence of noise in event streams, localization or mapping frameworks can potentially excel with \emph{indirect methods}, which extract and match features, such as corners and lines.
However, due to the unstable performance of existing feature extraction methods, indirect event-based mapping remains underexplored~\cite{sh-ch10-event, gallego2020event}.
In contrast, methods that leverage all events are usually referred to as \emph{direct methods}.
A prior event-based line mapping method, EL-SLAM~\cite{elslam}, is a direct method that lifts 2D lines into 3D using depth maps derived from ray densities~\cite{emvs}. 
As with other direct methods, it is inherently sensitive to noise, resulting in degraded reconstruction quality.
We present, to the best of our knowledge, the first indirect 3D line mapping pipeline for event data. 
We introduce a series of event-specific techniques that enable accurate correspondence search and 3D line reconstruction, and stably achieve noise-robust and reliable 3D line reconstruction, which leads to a performant and practical solution, as illustrated in~\Cref{fig:teaser}.

Our method carefully devises a reliable line extraction and association method tailored to event data, which serves as a crucial part in developing a robust indirect method using lines. 
While fast and accurate line extraction methods exist for images captured by conventional cameras, similar counterparts for event data are less mature, resulting in relatively poor performance in event cameras, as observed in many other computer vision tasks.
Many previous works compile events within a temporal window into a 2D frame with accumulated frames and devise image-based algorithms on them.
The distribution of event signals depends on the complex interplay of camera motion, scene condition, and sensor characteristics, making it challenging to choose the optimal temporal window.
A large window may result in blurry traces of edges, whereas critical structure may not be captured if the window is too small~\cite{esvo, esvo2}.
Furthermore, various accumulation strategies~\cite{cohen2018spatial, park2016performance, alzugaray2018ace, lagorce2016hots, esvo2} exhibit different strengths, making it non-trivial to select the most suitable representation for a given task.
We propose composing several versions of frames that apply multiple temporal windows for multiple representations, and then quickly use a performant 2D line extraction method~\cite{elsed} to acquire a pool of candidate lines.
Starting from the seed lines, we fit piece-wise planar geometry in the space-time volume and select prominent lines from varying densities of events with noise. 
Accompanied by a lightweight event-based point matcher~\cite{minima}, we can also benefit from the generalization of direct methods and establish efficient and accurate temporal correspondence between lines.

We then build a compact 3D line map representation accompanied by elegant geometric loss functions, which can adapt to multi-modal measurements and maps as long as we can extract 2D or 3D lines.
The consistent lines extracted from our frame-based association constitute a selective set of 3D structures represented by two endpoints, dramatically reducing the memory requirement compared to point-based maps.
We adopt the orthonormal representation to optimize 3D infinite lines without overparametrization, and obtain final 3D line segments via endpoint trimming.
Previous works have often projected 3D lines onto 2D image planes to compute distances~\cite{limap, line_map_3d_pp}, as defining the direct 3D distance is hindered by sign ambiguity~\cite{ICCV-2025-Shin}. 
However, discarding the depth information leads to inconsistent distance measures whose magnitudes do not correctly reflect the actual reconstruction errors, especially when the focal length varies.
Instead, we define the cost functions using geodesic distance on the Grassmann manifold~\cite{ICCV-2025-Shin}, which correctly represent errors in 3D space without projection, and use them to align 3D lines with multi-view observations for both extracted lines and the associated inlier events.
The distance metric is adapted to jointly refine the 3D line map and the camera poses.
Furthermore, the deduced geometric constraint enables any lines to serve as effective mid-level representations, universally applicable to other modalities. 
Cross-modal scenarios indicate extensive applicability that may overcome the inherent limitation of purely event-based representations in terms of diverse datasets or practical algorithms.

We validate our method on both a newly constructed synthetic dataset and real-world data, achieving better quantitative and qualitative results.
Furthermore, we demonstrate that our reconstructed line maps serve as effective mid-level representations in cross-modal scenarios, including registration into 3D pointcloud maps and panoramic image localization, and evaluate their performance through task-based assessments in these downstream applications.
In summary, we present the following technical contributions by building a framework for event-based line reconstruction:
\begin{itemize}
\item We present a series of event-specific techniques to robustly and accurately detect, refine, and match line features from diverse measurements.
    \item We introduce a multi-view reconstruction method that estimates 3D lines by jointly using line and event observations with accurate distance measures.
    \item Our line map demonstrates superior performance in event-based mapping and pose refinement, fully leveraging the advantages of indirect methods in noisy and inconsistent event data.
    \item Our reconstructed line maps serve as effective mid-level representations for cross-modal applications, such as registration and localization, further extending their real-world applicability.
\end{itemize}

%% file: sec/02_related_works.tex
\section{Related work}
\subsection{Event-based Mapping}
Due to the high dynamic range and temporal resolution, event cameras have long been sought after for robust mapping.
Existing event-based mapping methods could be classified into direct and indirect methods.

Direct methods accumulate or align all available events without relying on feature extraction~\cite{real_time_3d_event,emvs, evo, elslam, cmax, cmax_slam}. EMVS and EVO~\cite{evo} reconstruct 3D point clouds by accumulating event rays and identifying local maxima in ray densities. CMax-based methods~\cite{cmax, cmax_slam, emba} aim to maximize the alignment of warped event images to construct panoramic 2D maps. Although direct methods are generally applicable, they are inherently sensitive to the noisy nature of event data because all events are treated equally. To mitigate this limitation, hybrid methods have been proposed that incorporate frame-based cameras~\cite{eds, evi-sam} or use stereo event setups~\cite{esvo, esvo2}.
Learning-based approaches typically employ a direct method, learning desired outputs directly from the raw data.
Several works estimate depth from event streams~\cite{e2depth, emdepth, ereformer, unsup_ev}. 
These methods leverage deep learning architectures developed for frame-based vision, and are trained either in a supervised~\cite{e2depth, ereformer} or unsupervised manner~\cite{unsup_ev, emdepth}. However, due to the limited availability of large and diverse event datasets, their generalizability remains restricted compared to their frame-based counterparts.

Indirect methods abstract events into compact geometric primitives and explicitly utilize feature correspondences~\cite{ultimate_slam, pl-evio, esvio, evi-sam}. These methods benefit from mature geometric algorithms~\cite{mvg} and are generally more robust to noise. However, robust feature detection and matching, which are critical to the accuracy of indirect methods, are challenging due to the inherent sparsity, asynchrony, and noise in event data. As a result, prior works often use additional sensors, such as frame-based cameras~\cite{pl-evio, ultimate_slam, evi-sam} or stereo event cameras~\cite{esvio}.

In this work, we present, to the best of our knowledge, the first line mapping method for monocular event data. We propose a series of event-specific techniques that overcome the usual instability.
We represent event data with line features to build 3D line maps motivated by the fact that event cameras primarily respond to edges, and that most edges in man-made environments can be well-approximated by lines. Most closely related to our work, EL-SLAM~\cite{elslam} also reconstructs 3D line maps based on the same motivation. However, it follows a direct approach that lifts 2D line information to 3D using EMVS, and thus suffers from noise sensitivity. Other works also leverage line information in event cameras, but they mainly focus on motion estimation rather than mapping~\cite{pl-evio, idol, eventail1, eventail2, eventail3}.

\begin{figure*}[t]
    \centering
    \includegraphics[width=\linewidth]{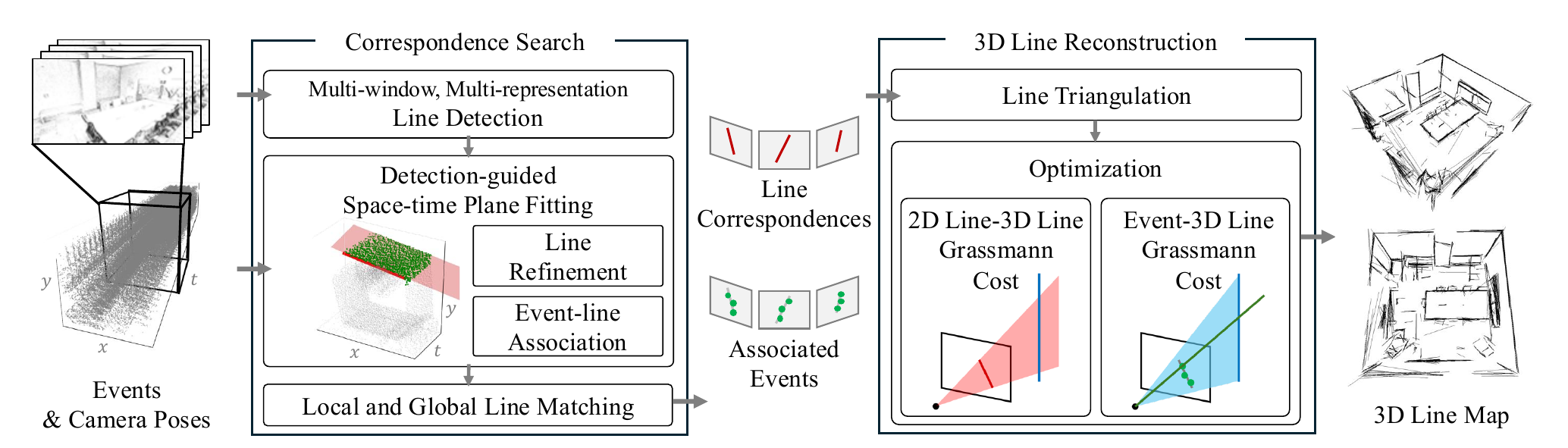}
    \caption{Method overview. Our 3D line mapping pipeline, RoEL, consists of two stages: correspondence search and 3D line reconstruction. The first stage takes events and camera poses as input. Through line detection, plane fitting, and matching processes that take into account the characteristics of event data, it finds line correspondences and identifies the events supporting each line. In the second stage, our method triangulates 2D lines to generate initial 3D lines. We further optimize 3D lines with multi-view observations using cost functions defined in 3D space based on the Grassmann distance. Finally, our method outputs 3D line segments that effectively represent the scene.}
    \label{fig:overview}
\end{figure*}

\subsection{3D Line Reconstruction}
Compared to commonly used 3D representations such as point clouds or meshes, lines can succinctly capture the salient geometry of man-made structures, leading to memory-efficient mapping.
Existing works build 3D line maps by directly processing pre-captured point clouds~\cite{lin2017facet, lu2019fast} or lifting image observations to 3D~\cite{sfm_line_corresp,sfm_line_pioneer}.
The former set of works operate from plane detections on the input point clouds followed by polygonal contour extraction to obtain line segments~\cite{lin2017facet, lu2019fast}.
For the latter set, monocular approaches~\cite{line_map_single,plane_rcnn} train a neural network to regress lines (or planar patches) from image inputs.
The more widely studied multi-view approaches~\cite{line_map_elsr,line_map_pl_slam,line_map_pl_slam_2,limap,sfm_line_corresp,sfm_line_pioneer,sfm_line_relaxed,sfm_line_resection,sfm_line_robust_sfm, line_map_3d, line_map_3d_pp} aggregate 2D line information from images taken at various viewpoints of a scene and lift them to 3D. 
Specifically, these methods often proceed by i) extracting and matching lines from images using off-the-shelf detectors and matchers~\cite{deeplsd,elsed,sold2,gluestick}, ii) triangulating lines using camera pose information~\cite{sfm_line_pioneer, line_map_3d_pp, limap}, and iii) running bundle adjustment between 2D / 3D lines to refine reconstruction quality.
Inspired from these approaches, we propose a correspondence-based 3D line reconstruction pipeline that operates using event camera measurements.
As the working principle of event cameras fundamentally differs from that of frame-based imaging, we propose a stack of event-specific methods to enable robust correspondence search and accurate 3D line reconstruction.

%% file: sec/03_method.tex
\section{Method Overview}
\label{sec:method_overview}
We propose RoEL, a full pipeline for 3D line reconstruction from event data by finding line correspondences.
Given a set of raw events  
\begin{equation}
\mathcal{E} = \{ e_k = (x_k, y_k, t_k, p_k) \}_{k=1}^M,
\end{equation}
where $ (x_k, y_k) $ denotes the pixel location, $ t_k $ the timestamp, and $ p_k \in \{-1, +1\} $ the polarity,  
and a set of known camera poses $ \{P_i\}_{i=1}^N $ with associated timestamps $ \{t^P_i\}_{i=1}^N $,  
our method reconstructs a 3D line-based map of the scene, represented as $ \{ \mathcal{L}_j \}_{j=1}^J. $
As illustrated in~\Cref{fig:overview}, the proposed system is composed of two main stages: correspondence search and 3D line reconstruction.

In the correspondence search stage, we process raw events to obtain line correspondences and associated events with their corresponding lines.
We introduce a series of event-based techniques into each step to enable robust correspondence search for event cameras.
In the first step, we detect 2D line segments using our robust solution for leveraging frame-based line detectors.
Our multi-window, multi-representation line detection addresses the ambiguity of event-to-image conversion and enables stable line extraction.
Second, we fit planes in the space–time volume of events to refine the detected 2D lines and associate inlier events.
By using line detection as a guide for plane fitting, our method achieves more complete and accurate line sets even in complex scene structures.
Third, we perform line matching to cluster 2D lines corresponding to the same 3D line.
Our local line matching strategy, applied between temporally adjacent frames, provides a matching mechanism that is well-suited for temporally dense event streams.

In the 3D line reconstruction stage, we estimate 3D line segments from multi-view 2D lines and their associated event observations.
A key feature of this stage is that our cost functions are defined directly in 3D space using geodesic distances on the Grassmann manifold~\cite{ICCV-2025-Shin}, unlike projection-based cost functions that lose 3D geometric information.
First, we triangulate 2D lines to obtain an initial 3D line map.
Our RANSAC-based triangulation mitigates the impact of outliers introduced during the correspondence search stage.
Second, we refine both 3D lines and camera poses through joint optimization.
In this step, we present two types of cost functions: one that measures consistency between 3D lines and their 2D projections, and another that leverages event observations.
To avoid suboptimal convergence caused by over-parameterization, we represent 3D infinite lines using a minimal orthonormal parameterization.
Finally, we trim the optimized infinite lines into finite segments using their 2D observations, resulting in a robust and accurate 3D line reconstruction.

\input{sec/03-1_front-end}
\input{sec/03-2_back-end}

%% file: sec/03-1_front-end.tex
\section{Correspondence Search}
\label{subsec:front-end}

In this section, we present a series of event-specific methods that enable robust line detection, refinement, and matching.
We also associate events with their corresponding lines, which are later utilized in the subsequent reconstruction stage.
The correspondence search module consists of three stages: a line detection stage that leverages a multi-window, multi-representation strategy to reduce false negatives (\Cref{subsubsec:line_det}), a space-time plane fitting procedure that simultaneously refines detected lines and associates events with them (\Cref{subsubsec:plane_fit}), and an inter-frame line matching strategy to establish consistent correspondences across time (\Cref{subsubsec:line_mat}).

\subsection{Multi-window, Multi-representation Line Detection}
\label{subsubsec:line_det}
Accurate 2D line detection from events is a critical stage, as line features detected at this stage serve as building blocks for subsequent stages.
However, accumulating events into frame-like representations and applying frame-based line detectors is challenging due to the sparse and non-uniform characteristics of event data.
Since event cameras respond to local brightness changes, the density and distribution of events vary significantly depending on scene complexity, camera motion, lighting conditions, and sensor characteristics.
The number of accumulated events significantly affects the appearance of edges: too few leads to missing structures, while too many results in blurred edges.
Moreover, different event accumulation strategies~\cite{cohen2018spatial, park2016performance, alzugaray2018ace, lagorce2016hots, esvo2} have their own strengths and weaknesses that vary depending on scene complexity.

We propose a multi-window, multi-representation line detection strategy to minimize false negatives.
We generate multiple images by varying both the integration windows and the event representations.
Then, we apply a fast and light frame-based line detector~\cite{elsed} to each generated image and aggregate the resulting line candidates.
During aggregation, lines with small 2D perpendicular distances are merged to eliminate redundancy.
2D perpendicular distance is defined as the maximum perpendicular distance from the endpoints to the corresponding infinite line.
This approach enables robust line detection under diverse scene conditions and camera motion.
Although aggregating multi-frame line detections may introduce noisy lines, these inaccuracies can be corrected in subsequent steps.
This strategy allows us to minimize missed detections and thereby enhance the completeness of the reconstructed 3D line map.
We construct image-like event frames at fixed time intervals and perform the above line detection sequentially.

\subsection{Detection-guided Space-time Plane Fitting}
\label{subsubsec:plane_fit}
We propose a geometric procedure based on space-time plane fitting to benefit from both direct and indirect paradigms.
While indirect methods provide robustness by explicitly matching features, they suffer when detection is imperfect.
On the other hand, direct methods operate directly on raw event streams.
They are thus free from errors introduced during feature detection, but they tend to be sensitive to noise and incur significant computational cost.
We selectively extract line-supporting events from the raw stream, preserving the advantages of direct methods while reducing noise and lowering computational cost.
At the same time, we also improve line detection accuracy, which is important for achieving reliable results in indirect reconstruction methods.
Space-time plane fitting allows us to refine the initially detected 2D lines and extract a subset of events associated with each line, thereby obtaining a compact and informative representation of the raw event data.

\begin{figure*}[t]
    \centering
    \includegraphics[width=0.95\linewidth]{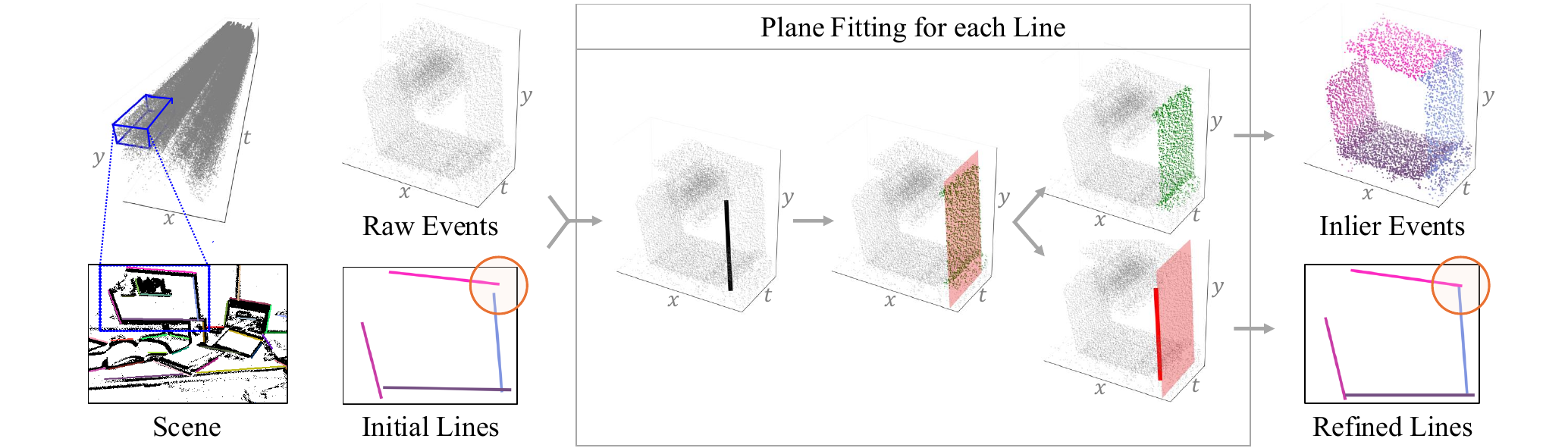}    
    \caption{Detection-guided space-time plane fitting. Through plane fitting for each line, raw events are distilled into inlier events, and noisy initial lines are refined. In the center box, the black line depicts an initial line which is used for candidate events selection for plane fitting, the red plane indicates the fitted plane, the red line represents the refined line, and the green dots denote all inlier events. Each line is associated with its corresponding events, and the inaccurate 2D lines are refined, as highlighted in the orange circle.}
    \label{fig:method_plane_fitting}
\end{figure*}

A moving camera observing a 3D line generates a surface in the $(x, y, t)$ space, and it can be locally approximated as a plane over short time intervals.
As demonstrated in~\Cref{fig:method_plane_fitting}, we treat events as 3D points and apply RANSAC-based~\cite{ransac} plane fitting using a point-to-plane distance metric.

Given a $j^{th}$ detected 2D line segment $\ell_i^j$ in frame $i$, defined by two endpoints $\textbf{p}_1=(x_1,y_1)$ and $\textbf{p}_2=(x_2,y_2)$, we first select candidate events in its spatial neighborhood:
\begin{equation}
\label{eq:neighbor}
\mathcal{E}_i^j = \left\{ e_k \in \mathcal{E} \;\middle|\; (x_k, y_k) \in \mathcal{N}(\ell_i^j) \right\},
\end{equation}
where the spatial neighborhood $\mathcal{N}(\ell_i^j)$ is defined based on the distance from 2D line segments to points.
We then find a plane $\pi_i^j: a_i^jx+b_i^jy+c_i^jt+d_i^j=0$ to these events using RANSAC:
\begin{equation}
\pi_i^j = \arg\min_{\pi} \sum_{e_k \in \mathcal{E}_i^j} d_{\text{plane}}\left((x_k, y_k, t_k), \pi \right)^2,
\end{equation}
where $d_{\text{plane}}$ is the point-to-plane distance. Events that lie within a threshold distance $\tau$ from the fitted plane are considered inliers and are associated with the line:
\begin{equation}
\mathcal{E}_{i, \text{assoc}}^j = \left\{ e_k \in \mathcal{E}_i^j \;\middle|\; d_{\text{plane}}\left((x_k, y_k, t_k), \pi_i^j \right) < \tau \right\}.
\end{equation}
Among these inliers, we select the $N_e$ events whose timestamps are closest to the line observation time $t_i$, following the same motivation as the spatial distance-based filtering strategy in~\cite{liu2024line}.
The selected events are then stored for later use in the 3D line reconstruction stage~(\Cref{subsec:back-end}).
Simultaneously, the refined 2D line is obtained by intersecting the fitted plane with the constant-time slice at $t_i$:
\begin{equation}
\label{eq:slicing}
\tilde{\ell}_i^j = \pi_i^j \cap \{ t = t_i \}.
\end{equation}
Since $\tilde{\ell}_i^j$ is an infinite line, its endpoints are calculated by projecting the endpoints of the initially detected line $\ell_i^j$ onto the $\tilde{\ell}_i^j$.
This improves the spatial accuracy of detected lines, especially since event-accumulated images often produce thick, imprecise edges.
This joint process of refinement and association reduces noise sensitivity and provides an abstract but reliable event subset for downstream optimization.

\subsection{Local and Global Line Matching}
\label{subsubsec:line_mat}
This stage aims to establish correspondences between lines detected in different frames.
To achieve this, we adopt a two-stage matching strategy that combines local and global approaches.
For temporally adjacent frames, we first perform local matching by applying a mutual nearest neighbor search~\cite{mutual_nn} based on the perpendicular distance between lines.
Although the search is applied to all detected lines, the method remains efficient.
This simple and efficient method works well for short-term associations but may fail in the presence of flickering or motion variations, both of which are common in event-based data.
To address these challenges, we incorporate a global matching stage by sampling frames over longer time intervals and applying a modality-invariant matching model~\cite{minima} to obtain point correspondences from event images.
Line correspondences are then inferred when a sufficient number of point correspondences are found around each line.
If two line tracks obtained from local matching are found to correspond under the global model, they are merged accordingly.
This hybrid matching framework produces dense and temporally consistent line tracks, which serve as a reliable foundation for subsequent 3D reconstruction.

%% file: sec/03-2_back-end.tex
\section{3D line reconstruction}
\label{subsec:back-end}

\begin{figure}
    \centering
    \includegraphics[width=0.9\linewidth]{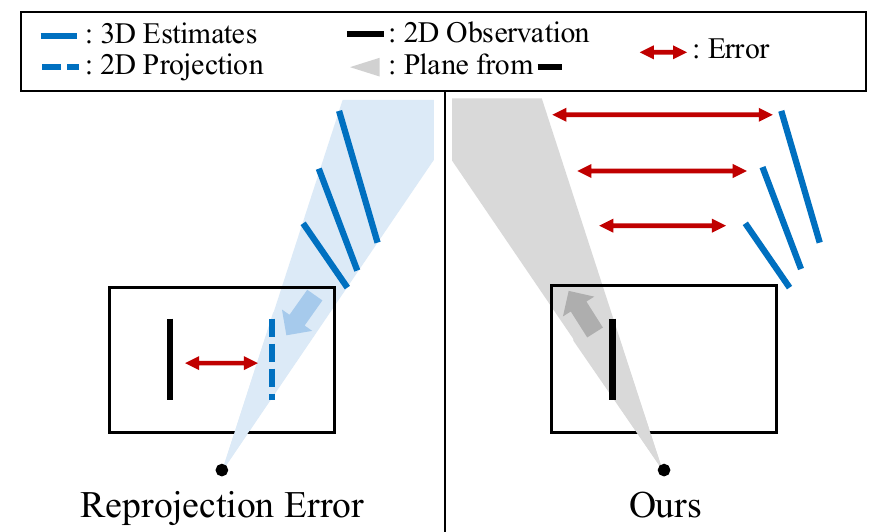}
    \caption{Difference from reprojection error. The two columns illustrate the same error-calculation scenario consisting of 3D line estimates and a 2D observation. In this scenario, different 3D line estimates project to the same 2D location, leading to identical reprojection errors. Our Grassmann-based cost evaluates geometric consistency directly in 3D.}
    \label{fig:grassmann}
\end{figure}

The 3D line reconstruction module constructs a 3D line map based on line correspondences and their associated events.
It also allows for joint pose refinement when the initial estimates are noisy.
The process consists of two main stages: line triangulation and multi-view line optimization.
In the triangulation stage, initial 3D lines are estimated from multi-view 2D line observations (\Cref{subsubsec:triang}).
We adopt a RANSAC-based~\cite{ransac} strategy to ensure robustness during triangulation.
In the optimization stage, 3D lines and camera poses are refined using multi-view 2D lines and their corresponding events (\Cref{subsubsec:opt}).
To evaluate the geometric consistency between 3D lines and 2D observations in $\mathbb{R}^3$, we employ a novel distance metrics defined on the Grassmann manifold, instead of the conventional reprojection error.
As illustrated in~\Cref{fig:grassmann}, when computing the error between 3D estimates and 2D observations, 3D lines that project to the same 2D location yield identical reprojection errors even if they are far apart in 3D space.
In contrast, our cost function, defined on the Grassmann manifold, accurately captures geometric discrepancies directly in 3D space.
We further apply an appropriate line parameterization to ensure stable optimization.
Prior to describing the triangulation and optimization stages, in~\Cref{subsubsec:graff}, we introduce the affine Grassmannian representation used throughout the reconstruction pipeline.

\subsection{Affine Grassmannian}
\label{subsubsec:graff}
In our reconstruction framework, each geometric primitive (\textit{i.e.}, 3D line or 3D plane) is represented as a point on a differentiable manifold known as the affine Grassmannian. The affine Grassmannian, denoted as $\text{Graff}(k,n)$, is defined as the set of all $k$-dimensional affine subspaces in $\mathbb{R}^n$. Letting $\mathbf{A} \in \mathbb{R}^{n \times k}$ denote an orthonormal basis matrix of the linear subspace and $\mathbf{b}_0 \in \mathbb{R}^n$ a displacement vector orthogonal to $\mathbf{A}$, an element on the manifold can be uniquely expressed as $\mathbb{A} + \mathbf{b}_0 := \mathrm{span}(\mathbf{A}) + \mathbf{b}_0$. For the analysis of geodesic distance on the manifold, an element $\mathbb{A} + \mathbf{b}_0$ is first mapped to a higher-dimensional linear subspace $\mathrm{Gr}(k+1, n+1)$ by embedding $z(\cdot)$, where the orthonormal basis matrix $\mathbf{Y}$ is defined as \cite{SIAM-2021-Lim}:
\begin{equation}
\mathbf {Y}_{z({\mathbb{A}+\mathbf{b}_0})} =    
\begin{pmatrix}
\mathbf{A} & \frac{\mathbf{b}_0}{\sqrt{1+\|\mathbf{b}_0\|^2}} \\
0 & \frac{1}{\sqrt{1+\|\mathbf{b}_0\|^2}} \\
\end{pmatrix}.
\end{equation} Then, the affine Grassmannian inherits the metrics of the typical linear subspace, where the geodesic distance between two elements is defined as the squared sum of the principal angles. Let $\mathbf{Y}_1$ and $\mathbf{Y}_2$ be two orthonormal basis matrices, with $\mathbf{Y}_1 \in \mathrm{Gr}(k, n)$ and $\mathbf{Y}_2 \in \mathrm{Gr}(l, n)$ $(k\leq l)$. The principal angles are then given by:
\begin{equation}\label{equ:prinAng}
\theta_i = \cos^{-1}{\sigma_i}, ; i = 1, \dots, k,
\end{equation}
where $\sigma_i$ is the $i^\text{th}$ singular value of $\mathbf{Y}_1^{\top}\mathbf{Y}_2$.

\subsection{Line Triangulation}
\label{subsubsec:triang}
Given a set of corresponding 2D lines $\{\tilde\ell_i\}_{i=1}^{N}$ across $N$ views with known camera poses $\{P_i\}_{i=1}^{N}$, we aim to select the most geometrically consistent 3D line $\mathcal{L}^*$ by evaluating multiple hypotheses, each triangulated from a minimal subset of two views. 
We first randomly sample two indices $p, q$, and triangulate a 3D line hypothesis $\mathcal{L}_{(p,q)}$ from the 2D lines $\tilde\ell_p$ and $\tilde\ell_q$ by finding intersection of two planes back-projected from two lines
\begin{equation}
\mathcal{L}_{(p,q)} = \text{Triangulate}(\tilde\ell_p, P_p, \tilde\ell_q, P_q).
\end{equation}
We then evaluate its geometric consistency with the remaining views, where $k \neq p, q$.
The simplest way to assess consistency between a 3D line and a 2D image line is by computing the distance in two-dimensional space through projection of the 3D line onto the image plane.
However, this projection disregards the 3D line’s configuration within the viewing plane (\textit{i.e.}, even if the 3D line moves arbitrarily within the plane, the consistency score remains unchanged on the image plane).

Therefore, we adopt a method to evaluate consistency as primitives in $\mathbb{R}^3$ by back-projecting the measured 2D line into the 3D plane and computing its proximity to the 3D line.
We embed both the lines and planes into the affine Grassmannian, and the features of the 3D line were shifted to include the origin, ensuring an invariant distance measure \cite{IROS-2022-Lusk}.
As a result, for each view $k$, we compute the Grassmann distance as the geometric error for each pair, which was then used to select inlier views within the RANSAC framework.
We also incorporate the distances defined in \cite{limap}, considering a 2D line as an inlier only if all distances are below their respective thresholds.

Furthermore, by selecting the best candidate $\mathcal{L}^*$ as the initial value, we refine the line parameters by optimizing the sum of the Grassmann distances across all inlier views.
For the total cost function, an optimizable form of the Grassmann distance was adopted, as defined in \cite{ICCV-2025-Shin}:
\begin{equation}\label{equ:TriCost}
E_{Tri}=\sum_{i}\left(\left\|\mathbf{P}_{ \mathbb{
\pi}^i}\mathbf{v}-\mathbf{v}\right\|_2^2 +
\left\|\mathbf{P}_{z(\mathbb{\pi}^i+\mathbf{d}_0^i)}\tilde{\mathbf{c}}_0- \tilde{\mathbf{c}}_0\right\|_2^2\right),
\end{equation} where $\pi^i + \mathbf{d}^i_0$ denotes the viewing plane of the $i^{\text{th}}$ inlier view, $\mathbf{v} + \mathbf{c}_0$ denotes the reconstructed 3D line as the target variables, $\mathbf{P}_{(\cdot)}$ is the projection matrix onto the linear subspace, and $\tilde{(\cdot)}$ represents the normalized homogeneous vector.

\begin{figure}
    \centering
    \includegraphics[width=0.9\linewidth]{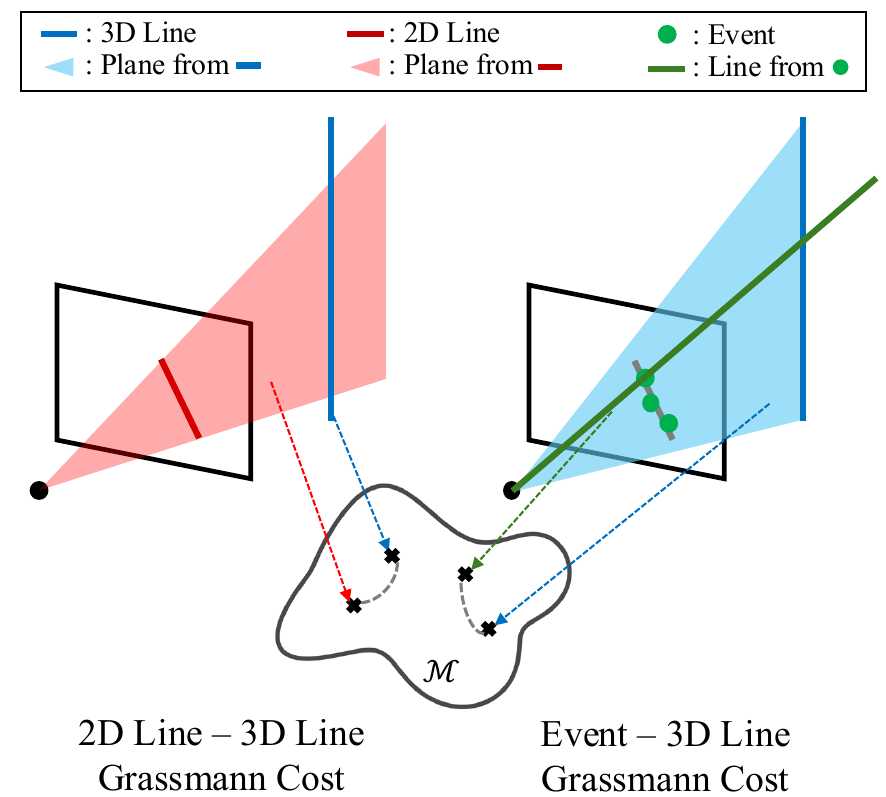}
    \caption{Two different cost functions defined for 2D line observations and 2D event observations. The geodesic distance on the Grassmann manifold serve as the basis for cost functions. The left figure illustrates the cost function defined between a 2D line and a 3D line. We use the plane-to-line Grassmann distance between the red plane, which is back-projected from the 2D line, and the blue 3D line. The right figure shows the cost function defined between an event and a 3D line. We use the plane-to-line Grassmann distance between the green line back-projected from the event and the blue plane connecting the 3D line and the camera center.}
    \label{fig:method_cost}
\end{figure}

\subsection{Multi-view Line Optimization}
\label{subsubsec:opt}

After triangulation, we refine the 3D line map and, optionally, the camera poses with optimization.
This process leverages both multi-view line observations and associated event data to improve reconstruction accuracy and robustness.
Specifically, we jointly optimize the parameters of each 3D line $\mathcal{L^*}_j$ and, if desired, the camera poses $\{P_i\}$ by minimizing a cost composed of two terms:  
the distance between the 3D line and its observed 2D lines, and the distance between the 3D line and its associated events. 

As seen in~\Cref{fig:method_cost}, to align the 3D and 2D lines in the first case, we back-project the 2D line onto the viewing plane within the reference frame using the 2D line parameters and the camera pose.
Similarly, the alignment of the event and the 3D line in the second case was achieved by representing their configuration with a plane and line in $\mathbb{R}^3$.
Specifically, the 3D line was transformed into the camera coordinate system and regarded as the viewing plane, which includes the camera center.
Each event was then back-projected onto a ray that constitutes the 3D line.

As both cases leverage the optimization of plane-to-line distance, we adopted the following optimizable Grassmann distance, adding the camera pose $P = (R, t)$ from Equation \eqref{equ:TriCost} as a variable to represent the error metrics of a single correspondence $(\Pi,\mathcal{L}):=(\pi + \mathbf{d}_0, \mathbf{v} + \mathbf{c}_0)$: \begin{equation}\label{equ:Problem2}
e_{Ref}(\Pi,\mathcal{L},P)=\left(\left\|\mathbf{P}_{ R \cdot \mathbb{
\pi}}\mathbf{v}-\mathbf{v}\right\|_2^2 +
\left\|\mathbf{P}_{z(P\cdot(\mathbb{\pi}+\mathbf{d}_0))}\tilde{\mathbf{c}}_0- \tilde{\mathbf{c}}_0\right\|_2^2\right),
\end{equation} where $R \cdot$ and $P \cdot$ denote the $SO(3)$ and $SE(3)$ group actions on the Grassmann manifold and affine Grassmann manifold, respectively. Let $\mathcal{E}_{i, \text{assoc}}^j$ be the set of events associated with the 2D line $\tilde{\ell}_i^j$. We then define the following total cost function:
\begin{equation}
E = \sum_{i,j} w^j_i \cdot e_{Ref}^2(\tilde\ell_i^j, \mathcal{L}_j, P_i) + \lambda_{\text{event}} \cdot \sum_{e_k \in \mathcal{E}_{i,\text{assoc}}^j} e_{Ref}^2( \mathcal{L}_j,e_k, P_i),
\label{eq:opt_cost}
\end{equation}
where $w^j_i$ represents the 2D line length, which serves to downweight short line segments that are prone to inaccurate detection. The constant $\lambda_\text{event}$ is used to balance the two cost terms.
To ensure numerical stability and avoid overparameterization in 3D line optimization, we adopt the orthonormal representation~\cite{bartoli2005structure}, which provides a minimal 4-DoF parameterization of 3D lines.
While Pl\"{u}cker coordinates~\cite{plucker1828analytisch}, defined by a normalized direction vector $\mathbf{d}$ and a moment vector $\mathbf{m}$, offer convenient geometric operations such as transformation, projection, and intersection, they represent 3D lines with six parameters and one constraint, which can lead to instability during optimization.
To address this, we perform optimization in the orthonormal representation and subsequently convert it back to Pl\"{u}cker coordinates for further processing.
The conversion from Pl\"{u}cker coordinates $(\mathbf{d}, \mathbf{m})$ to the orthonormal representation is as follows:
\begin{equation}
\mathbf{U} = \left( 
\mathbf{d},\ 
\frac{\mathbf{m}}{\|\mathbf{m}\|},\ 
\frac{\mathbf{d} \times \mathbf{m}}{\|\mathbf{d} \times \mathbf{m}\|}
\right) \in SO(3),
\label{eq:ortho_so3}
\end{equation}
\begin{equation}
\mathbf{W} =
\begin{bmatrix}
w_1 & -w_2 \\
w_2 & \ \ w_1
\end{bmatrix}
\in SO(2).
\label{eq:ortho_so2}
\end{equation}
After optimization, 3D lines represented with the orthonormal representation are converted back to Pl\"{u}cker coordinates ($\hat{\mathbf{d}}$, $\hat{\mathbf{m}}$) as follows:
\begin{equation}
w_1 = \frac{1}{\sqrt{1 + \|\mathbf{m}\|^2}}, \quad
w_2 = \frac{\|\mathbf{m}\|}{\sqrt{1 + \|\mathbf{m}\|^2}},
\end{equation}
\begin{equation}
\hat{\mathbf{d}} = \mathbf{u}_1, \quad
\hat{\mathbf{m}} = \frac{w_2}{w_1} \mathbf{u}_2.
\end{equation}
Since inaccurate line tracks from the correspondence search module that contain different lines can lead to unstable optimization, we identify and discard such erroneous clusters when divergence is observed.

After obtaining the optimized Pl\"{u}cker coordinates, we extract a finite 3D line segment by trimming its endpoints.
By following~\cite{zhang2015building}, for each 2D line segment, we construct a perpendicular line at each endpoint and back-project it to obtain a 3D plane.
We then compute the intersection between this plane and the 3D line to estimate its 3D endpoints.
From the multiple 3D endpoint candidates obtained across views, we perform clustering to select the final two endpoints that define the 3D line segment.
We empirically observed that the proposed formulation indeed yields more accurate 3D lines, thereby stabilizing the pipeline.

%% file: sec/04_experiments.tex
\section{Experimental Results}
\label{sec:experiments}

We demonstrate that our method reconstructs a compact and accurate 3D line map from event data and refines camera poses.
Furthermore, we show that the reconstructed line map serves as an effective mid-level representation for cross-modal applications in downstream tasks.
The implementation details are explained in~\Cref{subsec:details}.
Baseline methods and datasets used in our experiments are described in~\Cref{subsec:baselines_datasets}.
Then, we provide both quantitative and qualitative evaluations of reconstruction performance in~\Cref{subsec:recon}.
We evaluate the usefulness of our line map in a cross-modal registration task with RGB-D point maps in~\Cref{subsec:registration}, and in a cross-modal localization task using panoramic RGB images in~\Cref{subsec:localization}.
The effectiveness of pose refinement through joint optimization is evaluated in~\Cref{subsec:pose}.
In~\Cref{subsec:challenging}, we compare our method against RGB-based mapping systems under challenging visual conditions, including motion blur and extreme lighting.

\input{sec/04-1_details}

\input{sec/04-2_baselines}
\input{sec/04-4_recon}

\input{sec/04-6_registration}
\input{sec/04-7_localization}
\input{sec/04-5_pose}
\input{sec/04-8_challenging}

%% file: sec/04-1_details.tex
\subsection{Implementation Details}
\label{subsec:details}
Our pipeline consists of two main stages: correspondence search and 3D line reconstruction.
The parameters for the correspondence search, particularly those related to the event camera’s resolution, are set individually for each dataset.
In contrast, \textit{the parameters for the 3D line reconstruction stage are fixed across all experiments.}
In this subsection, we describe the key implementation details for each stage.
Key parameters are summarized in~\Cref{tab:param}.

\subsubsection{Correspondence Search}
\label{subsubsec:correspondence}
The event frames used for line detection are generated at regular intervals, with the interval chosen to match the temporal resolution of a conventional frame-based camera.
Multi-window, multi-representation line detection is performed using two temporal windows (based on the number of events) and two image representations: binary image~\cite{cohen2018spatial} and timestamp image~\cite{park2016performance}.
For timestamp images, separate images are created per polarity.
During space-time plane fitting, we scale the time axis (in milliseconds) to normalize spatial and temporal scales.
In~\Cref{eq:neighbor}, candidate events are defined as those located within a distance of 10 from the line.
Mutual nearest neighbor line matching~\cite{mutual_nn} is applied between every pair of temporally adjacent frames, while global matching is performed using frames sampled every five timestamps. For the global stage, we obtain and merge point correspondences from both binary and timestamp images.
A 2D line correspondence is established if there are more than 10 point matches within a distance of 10 pixels from the 2D line segment.
To verify the reliability of the obtained line correspondences, we use epipolar line-based matching score~\cite{line_map_3d_pp} for filtering.
Each frame computes line matches with its 20 nearest neighbors in terms of camera center distance.
We parallelize the matching process using multiprocessing.
All components of the correspondence search stage are implemented in Python.

\subsubsection{3D Line Reconstruction}
\label{subsubsec:reconstruction}
In the triangulation stage, candidate 3D lines are hypothesized by selecting pairs of 2D lines from each line cluster obtained in the matching stage.
We consider a 3D line hypothesis as an inlier if~\Cref{equ:TriCost} is below the threshold.
We also use 3D angle distance, 2D angle distance, perpendicular distance, and perspective distance~\cite{limap} for inlier selection.
We discard line clusters with few inliers.
To eliminate redundancy, we apply DBSCAN~\cite{dbscan} clustering to the triangulated lines based on their direction, disparity, and center position.
Within each cluster, the longest line is retained.
We use a fixed number of events per line to allow efficient batch-wise computation in the optimization step.
We represent~\Cref{eq:ortho_so3} in the Lie algebra $\mathfrak{so}(3)$ using the logarithmic map and perform optimization in the tangent space.
The triangulation module is implemented in C++ with Python bindings, while the optimization is implemented in Python using PyTorch.

\input{tables/parameters}

%% file: tables/parameters.tex
\renewcommand{\arraystretch}{1.25}
\begin{table}[t]
\centering
\caption{Key parameters used in our pipeline.}
\label{tab:param}
\large
\resizebox{0.43\textwidth}{!}{
\begin{tabular}{l c}
\hline
\textbf{Parameter} & \textbf{Value} \\
\hline
Inlier selection threshold $\tau$ & 2 \\
Inlier selection $N_e$ & 100 \\
Time-axis scaling for plane fitting & 20 \\
Candidate distance to line in \Cref{eq:neighbor} & 10 \\
Epipolar score threshold & 0.3 \\
Inlier error threshold using~\Cref{equ:TriCost} & 1.4 \\
Min inliers per line cluster & 10 \\
Weighting parameter $\lambda_{\mathrm{event}}$ in~\Cref{eq:opt_cost} & $1\times 10^{4}$ \\
Number of events per line for optimization & 50 \\
\hline
\end{tabular}
}
\end{table}
\renewcommand{\arraystretch}{0.8}

%% file: sec/04-2_baselines.tex
\subsection{Baselines and Datasets}
\label{subsec:baselines_datasets}

\subsubsection{Baselines}
We compare our method with representative baselines across three categories: event-based point mapping~\cite{emvs, evo}, event-based line mapping~\cite{elslam}, and frame-based line mapping~\cite{limap}.
To ensure a fair comparison across all baselines, we used the same camera poses for every method.

EMVS~\cite{emvs} is one of the most widely used event-based 3D reconstruction methods.
It directly back-projects all available events and accumulates them in a Disparity Space Image (DSI), where peaks in ray density indicate probable scene surfaces.
Since EMVS uses a fixed voxel grid and is designed for short sequences, we follow the keyframe selection strategy from EVO~\cite{evo} to build multiple DSIs for long sequences.

EL-SLAM~\cite{elslam} is an event-based line SLAM system that reconstructs a 3D line map and estimates camera poses directly from event data.
Although EL-SLAM can perform camera tracking, we use the same camera poses as ours and compare only the mapping component for fair comparison.
Since the original implementation is not publicly available, we re-implemented the 3D line extraction pipeline based on the descriptions in the original paper.
It actively uses votes and depths of EMVS when detecting 2D lines and obtaining 3D line parameters.
We extract 3D lines at each keyframe and aggregate them to form the full line map.

LIMAP~\cite{limap} is a state-of-the-art frame-based method for 3D line mapping.
It is an indirect approach that explicitly utilizes line correspondences.
We provide LIMAP with event images as input, similar to our method.
We use SOLD2~\cite{sold2} for line detection and matching, as it was reported to yield the best performance in LIMAP.
We increase the number of visual neighbors used in line matching by a factor of three, while keeping all other parameters identical to the original implementation.

\input{tables/setup_dataset}

\subsubsection{Datasets}
We evaluate our method using four datasets: Replica~\cite{replica}, TUM-VIE\cite{tum-vie}, VECtor\cite{vector}, and the dataset from $I^2$-SLAM~\cite{i2slam}.
For synthetic datasets, we use ground-truth camera poses to isolate and evaluate the mapping performance.
Event data are generated using VID2E~\cite{vid2e}.
For real-world datasets, we use camera poses estimated by DEVO~\cite{devo}, a state-of-the-art event-based visual odometry method, to demonstrate the applicability of our approach in real-world environments.
The dataset descriptions and experimental settings are summarized in~\Cref{tab:setup_dataset}.

Replica~\cite{replica} provides high-quality, room-scale indoor scenes and is widely used for evaluating 3D reconstruction methods~\cite{sucar2021imap, nice_slam, monogs} due to its accurate ground-truth mesh.
We use the scenes and trajectories from iMAP~\cite{sucar2021imap}, and simulate event data by rendering dense RGB frames using the Habitat simulator~\cite{savva2019habitat}, with five times higher temporal resolution than iMAP.
Note that iMAP does not provide absolute timestamps, and thus we report the temporal resolution only in relative terms.
Events are then generated using VID2E~\cite{vid2e}.
For trajectory interpolation, we adopt the method from $I^2$-SLAM.
We simulate 20-second sequences at a resolution of 1200\,$\times$\,680 pixels.
Additionally, we evaluate on real-world sequences from TUM-VIE~\cite{tum-vie} and VECtor~\cite{vector} datasets.
We select scenes that contain rich line structures and where DEVO operates reliably.
For TUM-VIE, we use the \texttt{mocap-desk} and \texttt{mocap-desk2} sequences, which were captured with a Prophesee Gen4 sensor at 1280\,$\times$\,720 resolution.
For VECtor, we use the \texttt{desk-fast}, \texttt{sofa-fast}, \texttt{desk-normal}, and \texttt{sofa-normal} sequences, captured with a Prophesee Gen3 sensor at 640\,$\times$\,480 resolution.
For the experiments comparing our method against RGB-based mapping under challenging visual conditions, we follow the data generation procedure of $I^2$-SLAM~\cite{i2slam}.
We simulate RGB images under high-speed motion and underexposure by using the same tone-mapping procedure and scene settings as $I^2$-SLAM.
As in the Replica experiments, we render temporally dense frames and generate the corresponding event data using VID2E.

%% file: tables/setup_dataset.tex
\renewcommand{\arraystretch}{1.25}
\begin{table}[t]
\centering
\caption{Summary of datasets and experimental settings.}
\label{tab:setup_dataset}
\large
\resizebox{0.44\textwidth}{!}{
\begin{tabular}{ccccc}
\toprule
\textbf{Dataset} & \textbf{Sensor} & \textbf{Event} & \textbf{Pose} \\ 
\midrule
Replica~\cite{replica}  & Synthetic         & VID2E~\cite{vid2e}             & GT \\
TUM-VIE~\cite{tum-vie}  & Prophesee Gen4    & Real              & DEVO \\
VECtor~\cite{vector}   & Prophesee Gen3    & Real              & DEVO \\
$I^2$-SLAM~\cite{i2slam}  & Synthetic         & VID2E~\cite{vid2e}             & GT \\
\bottomrule
\end{tabular}
}
\end{table}
\renewcommand{\arraystretch}{0.8}

%% file: sec/04-4_recon.tex
\subsection{Reconstruction}
\label{subsec:recon}

\subsubsection{Setup}
\label{subsubsec:recon_setup}

\input{tables/recon_replica_transpose}
\input{figures/recon_replica}
\input{figures/recon_tum}

We evaluate the reconstruction performance of our method and the baselines on the Replica~\cite{replica}, TUM-VIE~\cite{tum-vie}, and VECtor~\cite{vector} datasets.
Quantitative evaluation is conducted on the Replica and VECtor datasets, which provide ground-truth depth maps.
Our evaluation protocol and metrics follow prior works~\cite{limap, emap, nef, sucar2021imap, nice_slam}.

To compare line maps with point maps, we uniformly sample points along each reconstructed line to generate a point-based map.
We use three evaluation metrics: \textit{Accuracy (mm)}, \textit{Completion (mm)}, and \textit{IoU (Intersection over Union)}.
\textit{Accuracy} and \textit{Completion} measure the distance between the estimated and ground-truth maps~\cite{emap, sucar2021imap, nice_slam}, while \textit{IoU} reflects the balance between precision and recall at a given threshold~\cite{limap, emap, nef}.
For calculating \textit{Completion} and \textit{IoU}, ground-truth edge map $G_{edge}$ is applied as event cameras observe edge regions.

It is the set of ground-truth points observable by an event camera.
To construct a ground-truth edge map, we: (1) extract edges from RGB images using PiDiNet~\cite{pidinet}, (2) filter the edge regions from the ground-truth depth map, and (3) fuse them into a 3D point cloud using ground-truth poses and TSDF-Fusion~\cite{tsdf-fusion, 3dmatch}.
For the \textit{Accuracy} calculation, ground-truth dense map $G_{\text{dense}}$, reconstructed by TSDF-Fusion with depth maps, is used for evaluation.
Both $G_{\text{edge}}$ and $G_{\text{dense}}$ are point cloud map. 

\textit{Accuracy} measures how closely the predicted points approximate the ground-truth surface.
It is defined as the average Euclidean distance from each predicted point to its nearest neighbor in the dense ground-truth point set.
Let $P$ be the set of predicted points and $G_{\text{dense}}$ the set of all ground-truth points. Then, \textit{Accuracy} is computed as:
\begin{equation}
    \textit{Accuracy} = \frac{1}{|P|} \sum_{p \in P} \min_{g \in G_{\text{dense}}} \| p - g \|_2.
\end{equation}

\textit{Completion} measures how well the predicted points cover the ground-truth structure.
It is defined as the average Euclidean distance from each ground-truth edge point to its nearest neighbor in the predicted point set.
Then, \textit{Completion} is computed as:
\begin{equation}
\label{eq:metric_completion}
    \textit{Completion} = \frac{1}{|G_{\text{edge}}|} \sum_{g \in G_{\text{edge}}} \min_{p \in P} \| g - p \|_2.
\end{equation}

\textit{IoU} quantifies the overall geometric agreement between the predicted and ground-truth structures under a given distance threshold $\delta$.
We count the number of predicted points within $\delta$ of any ground-truth point ($c_{\text{pred}}$), and the number of ground-truth edge points within $\tau$ of any predicted point ($c_{\text{gt}}$).
The \textit{IoU} is defined as:
\begin{equation}
    \textit{IoU}_{\delta} = \frac{ \min \left( c_{\text{pred}},\ c_{\text{gt}} \right) }{ |P| + |G| - \max \left( c_{\text{pred}},\ c_{\text{gt}} \right) }.
\end{equation}
This formulation ensures that \textit{IoU} captures both precision and recall aspects of reconstruction quality, providing a balanced measure.

However, unlike in the Replica dataset, we could not obtain an accurate ground-truth edge map for the real-world VECtor dataset, as the RGB images and depth maps are not aligned.
Therefore, we evaluate only \textit{Accuracy} and \textit{Precision} using the dense map, $G_{\text{dense}}$.

Additionally, to evaluate the compactness of each reconstruction, we count the number of geometric entities.
For point-based methods, we report the total number of points; for line-based methods, we report the number of reconstructed lines.

\input{tables/recon_vector_transpose}

\input{figures/recon_vector}

\subsubsection{Results}
\label{subsubsec:recon_results}

Our method achieves the best \textit{Accuracy} and \textit{IoU} scores across all scenes in the qualitative evaluation on the Replica~\cite{replica} dataset, as shown in~\Cref{tab:recon_replica_transpose}.
In contrast, EMVS~\cite{emvs} demonstrates higher \textit{Completion} on all scenes.
While our reconstructed points are more precise, EMVS exhibits broader coverage of ground-truth points.
This discrepancy arises because EMVS tends to produce dense reconstructions that include substantial noise, resulting in many false positives but fewer false negatives.
When evaluating with the \textit{IoU}, which balances the precision and recall aspects of the reconstruction, our method consistently outperforms all baselines across all scenes and distance thresholds.
Despite using only 2,494 line segments on average to represent the scene, our method achieves superior reconstruction quality compared to EMVS, which utilizes on estimate 884,709 points, demonstrating the compactness and efficiency of our line reconstruction.
Other line-based methods, EL-SLAM~\cite{elslam} and LIMAP~\cite{limap}, use a similar number of lines, but underperform compared to ours in all metrics on average.
As illustrated in~\Cref{tab:recon_vector_transpose}, our method also shows higher accuracy and precision on the VECtor~\cite{vector} dataset on average.
This indicates that our predicted points are geometrically close to the ground-truth structure and have a high proportion of true positives.

Qualitative evaluation on the Replica dataset in~\Cref{fig:recon_replica} shows that our method produces cleaner line maps with fewer noisy segments and successfully reconstructs scene structures and objects.
For example, in \texttt{office4}, all ceiling boundaries, desks, and chairs are reconstructed as 3D lines.
In particular, ceiling edges are challenging to reconstruct due to limited diversity of viewing directions, yet our method succeeds in recovering them across multiple scenes.
EMVS, a direct method that reconstructs 3D points by using back-projected ray densities, performs poorly in areas with insufficient ray overlap caused by restricted viewpoints.
Moreover, since EMVS uses all events, it tends to generate a large number of noisy 3D points.
The effect is more severe in cluttered scenes.
For example, around the sofa in \texttt{office3}, noisy points make it difficult to distinguish individual objects.
EL-SLAM also uses EMVS to build a map, but its line extraction step helps reduce some of the noise.
However, as shown in \texttt{office2}, when objects are clustered, EL-SLAM produces many spurious lines. It also fails to reconstruct ceiling boundaries.
LIMAP generates less noise than EMVS and EL-SLAM and recovers objects like tables and sofas with finer details.
Nevertheless, its reconstructions are still noisier and less geometrically accurate than ours.
For example, in \texttt{office3}, imprecise and redundant lines around the door obscure its structure.
Our method benefits from space-time plane fitting for refining 2D lines and combines 2D lines with associated events in the optimization step, resulting in more accurate 3D line reconstruction.

\input{figures/registration}
\input{tables/registration_replica_transpose}

In the TUM-VIE dataset, similar trends to Replica are observed in~\Cref{fig:recon_tum}.
EMVS and EL-SLAM yield less structured outputs than in Replica.
This is partly because we use camera poses estimated by DEVO instead of ground-truth poses.
Another reason is that the camera follows a forward-facing trajectory, which limits viewpoint diversity and reduces ray overlap.
Consequently, both ray density-based methods show lower reconstruction quality.
Compared to LIMAP, our method produces cleaner maps with fewer duplicated segments.
For instance, in \texttt{desk2} our method reconstructs all monitor boundaries with few duplicated lines.
Although objects like books and keyboards on the desk in \texttt{desk2} are missing in LIMAP, they are successfully reconstructed by our method.

Reconstruction results on the VECtor dataset are shown in~\Cref{fig:recon_vector}.
Here, the background represents the ground-truth map, while the black lines indicate the reconstructed map.
VECtor possesses even more challenges than TUM-VIE due to its low resolution and noisy poses.
While EMVS and EL-SLAM reconstruct the two monitors and a laptop, the outputs are noisy and spatially inaccurate.
LIMAP fails to recover parts of the monitor edges and the laptop.
Our method, although containing some redundant lines, reconstructs the monitors and laptop with better accuracy than EMVS and EL-SLAM.
Additionally, the vertical lines behind the desk in our results correspond to wall boundaries, which are also correctly recovered.

%% file: tables/recon_replica_transpose.tex
\begin{table*}[t]
\caption{Quantitative comparison of reconstruction quality on the Replica~\cite{replica} dataset. ↓: lower is better, ↑: higher is better. Bold indicates best. \#Rep. indicates the number of reconstructed lines or points.}
\label{tab:recon_replica_transpose}
\centering
\scriptsize
\resizebox{0.89\textwidth}{!}{
\begin{tabular}{ll c c c c c c c}
\toprule
\textbf{Method} & \textbf{Scene} & \textbf{Acc.↓} & \textbf{Comp.↓} & \textbf{IoU 5↑} & \textbf{IoU 10↑} & \textbf{IoU 20↑} & \makecell{\textbf{\#Rep.}} & \makecell{\textbf{Rep.}} \\
\midrule
\multirow{9}{*}[-0.45ex]{{EL-SLAM}~\cite{elslam}}
& avg     & 248.92 & 153.47 & 0.006 & 0.021 & 0.063 & 1169 & \multirow{9}{*}[-0.45ex]{Line} \\[0.3ex] 
\cdashline{2-8}[0.7pt/2pt]
\addlinespace[0.6ex]  
& \texttt{room0}    & 194.57 & 123.34 & 0.006 & 0.022 & 0.063 &  966 \\
& \texttt{room1}    & 344.48 &  99.50 & 0.004 & 0.014 & 0.048 & 1375 \\
& \texttt{room2}    & 254.05 & 177.93 & 0.007 & 0.026 & 0.074 & 1222 \\
& \texttt{office0}  & 242.14 &  81.87 & 0.009 & 0.028 & 0.081 & 1331 \\
& \texttt{office1}  & 236.25 &  72.84 & 0.006 & 0.023 & 0.071 & 1350 \\
& \texttt{office2}  & 261.58 & 314.53 & 0.004 & 0.017 & 0.046 & 1360 \\
& \texttt{office3}  & 293.97 & 159.63 & 0.006 & 0.019 & 0.058 & 1046 \\
& \texttt{office4}  & 164.28 & 198.10 & 0.004 & 0.016 & 0.060 &  701 \\

\midrule
\multirow{9}{*}[-0.45ex]{{LIMAP}~\cite{limap}}
& avg     & 536.79 & 113.47 & 0.006 & 0.024 & 0.073 & 2136 & \multirow{9}{*}[-0.45ex]{Line} \\[0.3ex] 
\cdashline{2-8}[0.7pt/2pt]
\addlinespace[0.6ex] 
& \texttt{room0}    & 119.48 &  69.94 & 0.005 & 0.022 & 0.071 & 3383 \\
& \texttt{room1}    & 134.92 & 122.79 & 0.006 & 0.025 & 0.077 & 1312 \\
& \texttt{room2}    & 127.26 & 124.18 & 0.007 & 0.025 & 0.069 & 2355 \\
& \texttt{office0}  &  98.62 &  92.98 & 0.006 & 0.026 & 0.088 & 2024 \\
& \texttt{office1}  &  76.60 & 138.55 & 0.009 & 0.035 & 0.103 &  674 \\
& \texttt{office2}  & 3465.06& 122.06 & 0.005 & 0.020 & 0.061 & 2171 \\
& \texttt{office3}  & 119.87 &  78.01 & 0.006 & 0.022 & 0.068 & 3281 \\
& \texttt{office4}  & 152.53 & 159.22 & 0.004 & 0.014 & 0.044 & 1887 \\
\midrule
\multirow{9}{*}[-0.45ex]{{EMVS}~\cite{emvs}}
& avg     & 181.24 &  \textbf{45.68} & 0.009 & 0.027 & 0.062 & 884\,709 & \multirow{9}{*}[-0.45ex]{Point} \\[0.3ex] 
\cdashline{2-8}[0.7pt/2pt]
\addlinespace[0.6ex] 
& \texttt{room0}    & 136.27 &  \textbf{40.18} & 0.010 & 0.031 & 0.075 & 739\,802 \\
& \texttt{room1}    & 258.74 &  \textbf{30.78} & 0.004 & 0.015 & 0.039 & 1\,045\,577 \\
& \texttt{room2}    & 150.90 &  \textbf{69.46} & 0.013 & 0.037 & 0.082 & 681\,203 \\
& \texttt{office0}  & 188.49 &  \textbf{24.82} & 0.011 & 0.035 & 0.079 & 935\,953 \\
& \texttt{office1}  &  99.87 &  \textbf{19.19} & 0.009 & 0.024 & 0.044 & 1\,698\,722 \\
& \texttt{office2}  & 172.17 &  \textbf{84.07} & 0.008 & 0.026 & 0.055 & 843\,780 \\
& \texttt{office3}  & 269.86 &  \textbf{44.43} & 0.009 & 0.029 & 0.070 & 601\,842 \\
& \texttt{office4}  & 173.63 &  \textbf{52.49} & 0.005 & 0.019 & 0.050 & 530\,792 \\
\midrule
\multirow{9}{*}[-0.45ex]{Ours}
& avg     &  \textbf{28.20} &  99.04 & \textbf{0.020} & \textbf{0.055} & \textbf{0.137} & 2494 & \multirow{9}{*}[-0.45ex]{Line} \\[0.3ex] 
\cdashline{2-8}[0.7pt/2pt]
\addlinespace[0.6ex] 
& \texttt{room0}    &  \textbf{38.26} &  82.91 & \textbf{0.014} & \textbf{0.038} & \textbf{0.096} & 3385 \\
& \texttt{room1}    &  \textbf{35.83} &  85.27 & \textbf{0.020} & \textbf{0.055} & \textbf{0.128} & 1977 \\
& \texttt{room2}    &  \textbf{24.34} & 139.20 & \textbf{0.017} & \textbf{0.046} & \textbf{0.132} & 2276 \\
& \texttt{office0}  &  \textbf{26.63} &  87.89 & \textbf{0.024} & \textbf{0.065} & \textbf{0.154} & 2486 \\
& \texttt{office1}  &  \textbf{23.25} & 113.19 & \textbf{0.042} & \textbf{0.105} & \textbf{0.219} &  676 \\
& \texttt{office2}  &  \textbf{23.09} & 160.77 & \textbf{0.020} & \textbf{0.056} & \textbf{0.152} & 2509 \\
& \texttt{office3}  &  \textbf{25.57} &  67.86 & \textbf{0.012} & \textbf{0.032} & \textbf{0.096} & 4155 \\
& \texttt{office4}  &  \textbf{28.63} &  55.20 & \textbf{0.014} & \textbf{0.042} & \textbf{0.120} & 2488 \\
\bottomrule
\end{tabular}
}
\end{table*}

%% file: figures/recon_replica.tex
\begin{figure*}[t]
    \centering
    \renewcommand{\arraystretch}{1.1}
    \begin{tabular}{@{}c@{\,}c@{\,}c@{\,}c@{\,}c@{\,}c@{}}
    
        \parbox[c]{0.02\textwidth}{\centering\rotatebox{90}{\texttt{office3}}} &
        \parbox[c]{0.18\textwidth}{\centering\includegraphics[width=\linewidth]{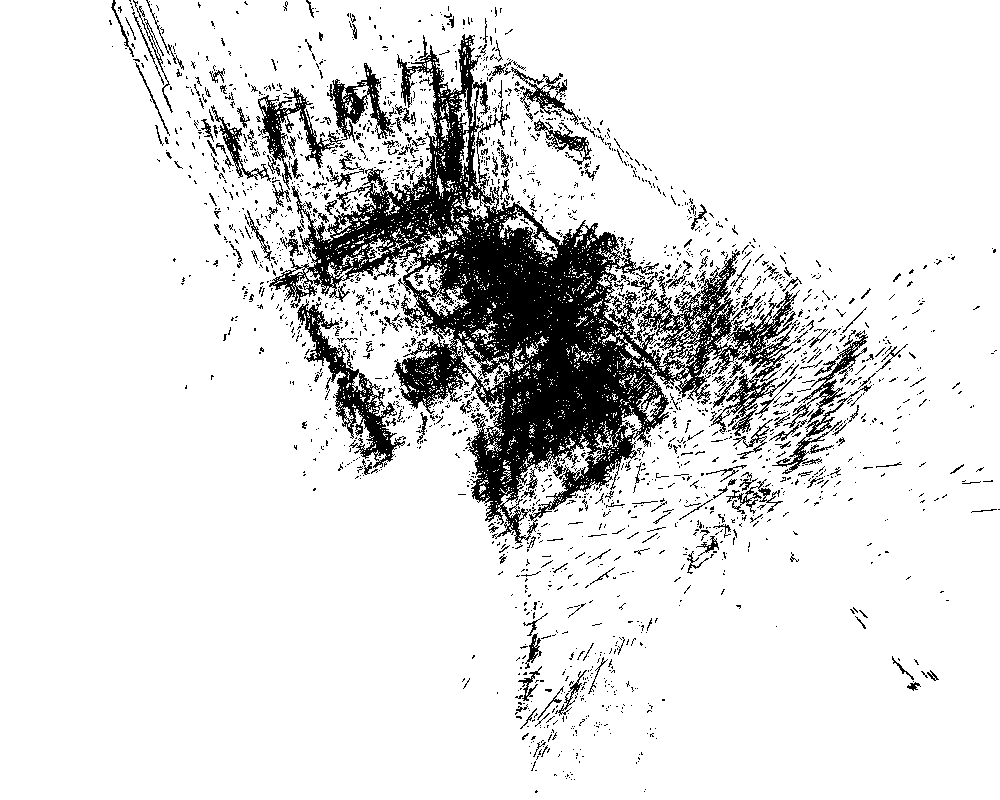}} &
        \parbox[c]{0.18\textwidth}{\centering\includegraphics[width=\linewidth]{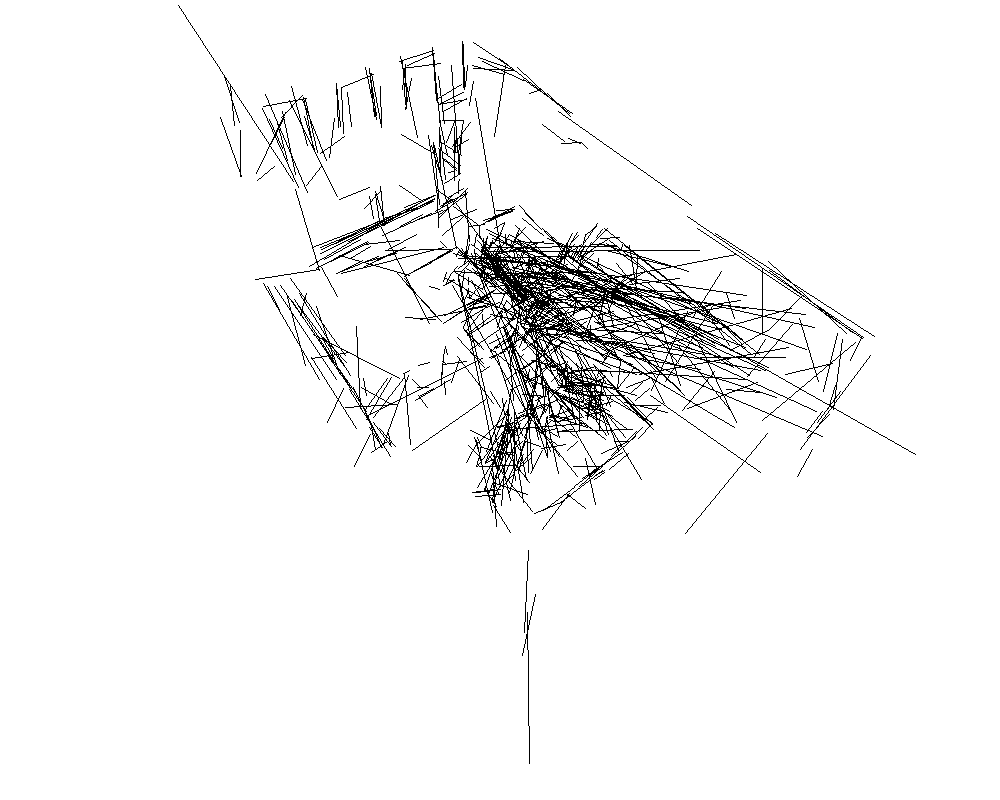}} &
        \parbox[c]{0.18\textwidth}{\centering\includegraphics[width=\linewidth]{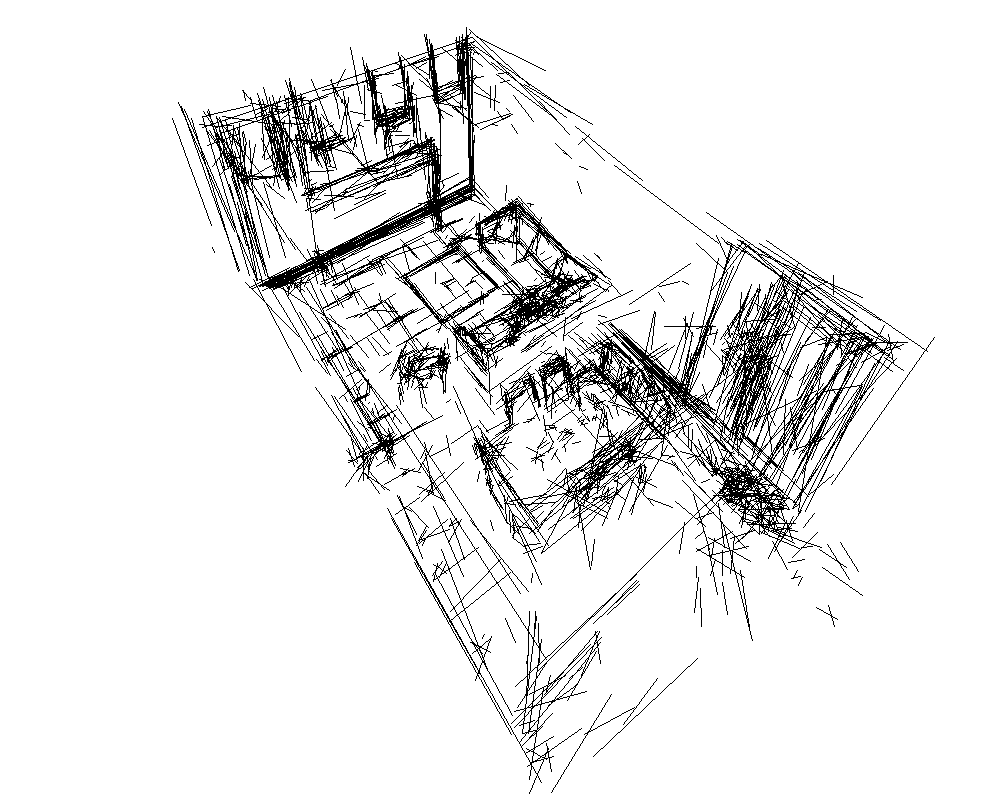}} &
        \parbox[c]{0.18\textwidth}{\centering\includegraphics[width=\linewidth]{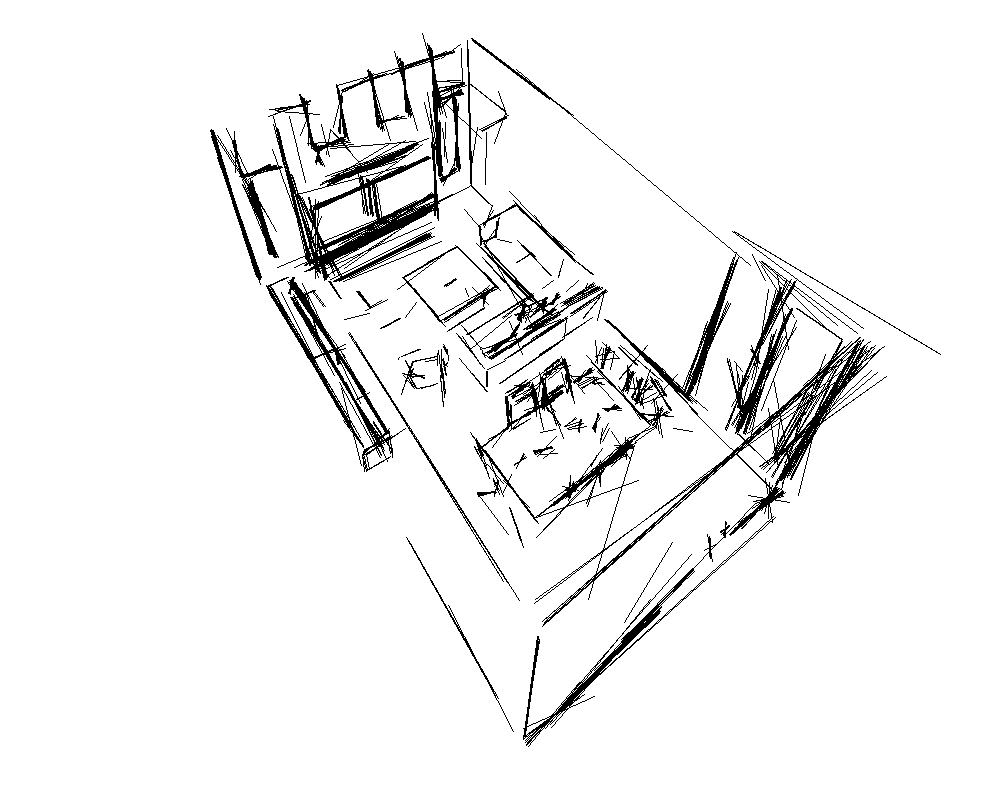}} &
        \parbox[c]{0.18\textwidth}{\centering\includegraphics[width=\linewidth]{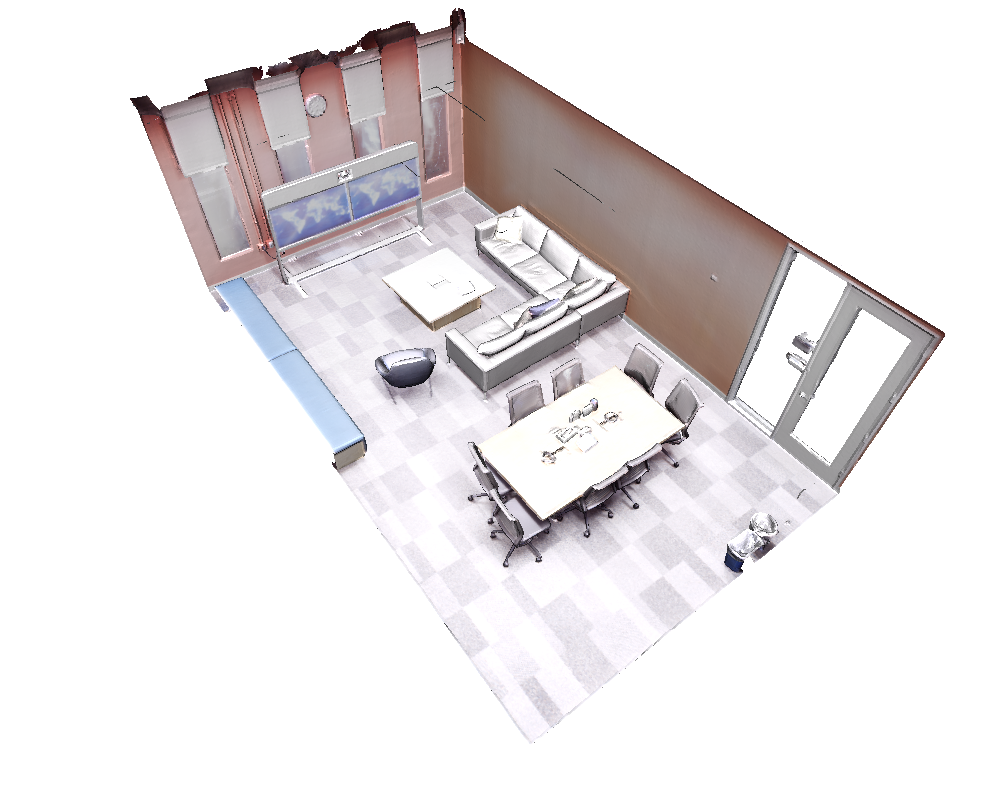}} \\

        \parbox[c]{0.02\textwidth}{\centering\rotatebox{90}{\texttt{office4}}} &
        \parbox[c]{0.18\textwidth}{\centering\includegraphics[width=\linewidth]{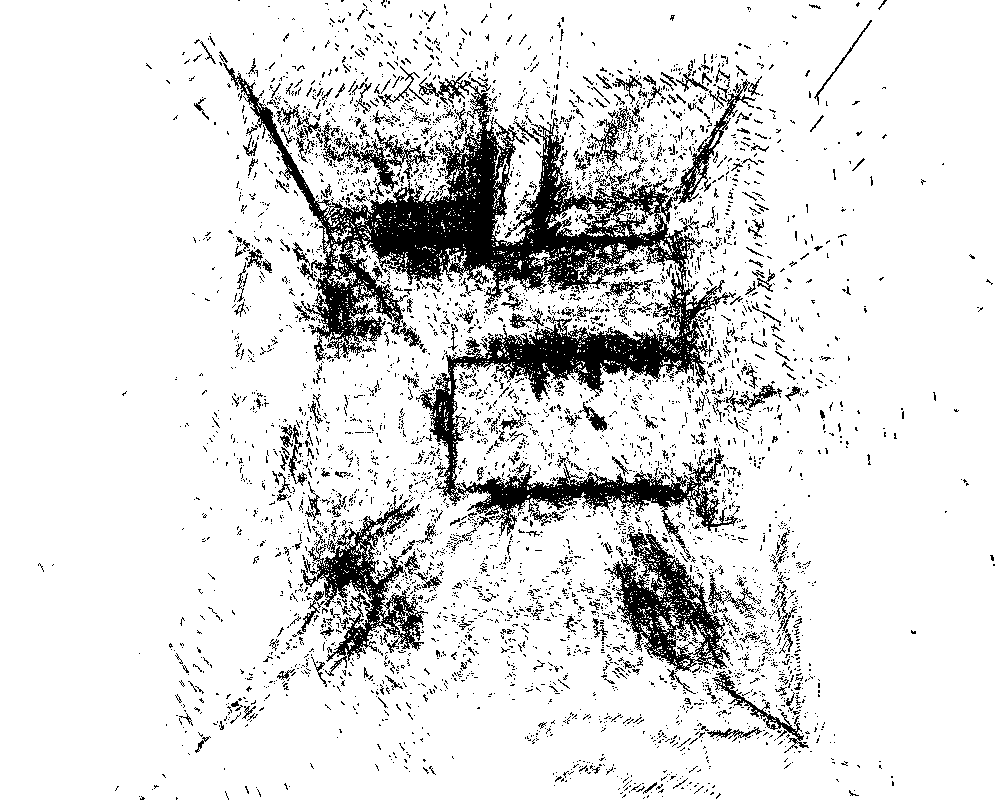}} &
        \parbox[c]{0.18\textwidth}{\centering\includegraphics[width=\linewidth]{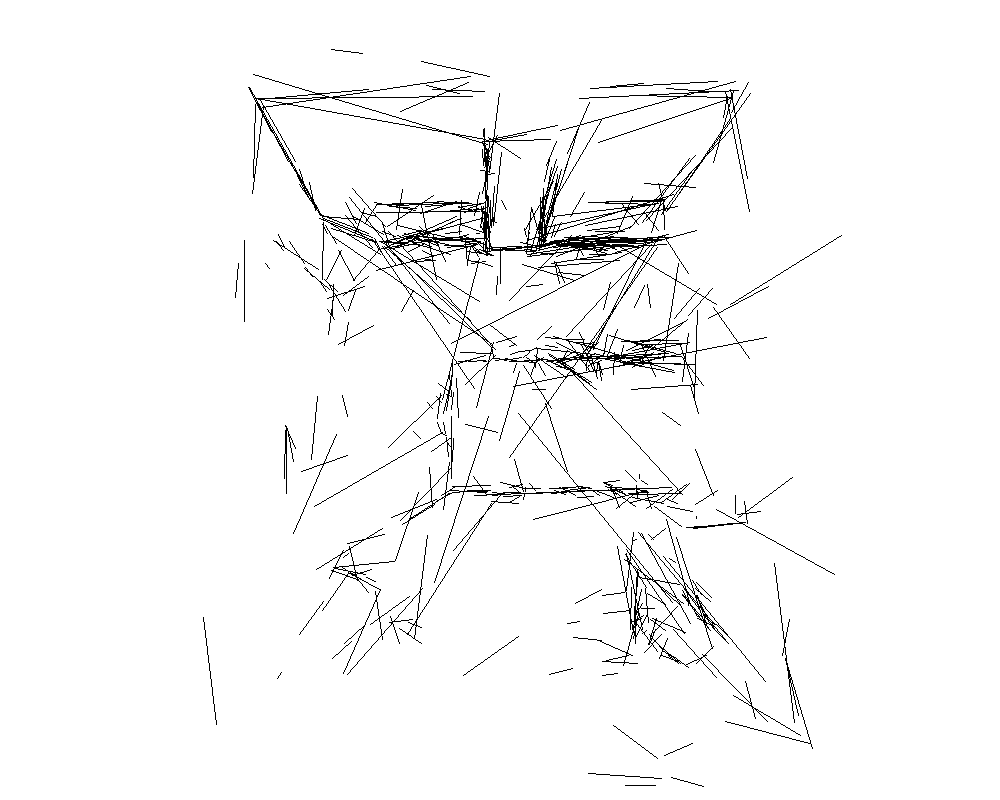}} &
        \parbox[c]{0.18\textwidth}{\centering\includegraphics[width=\linewidth]{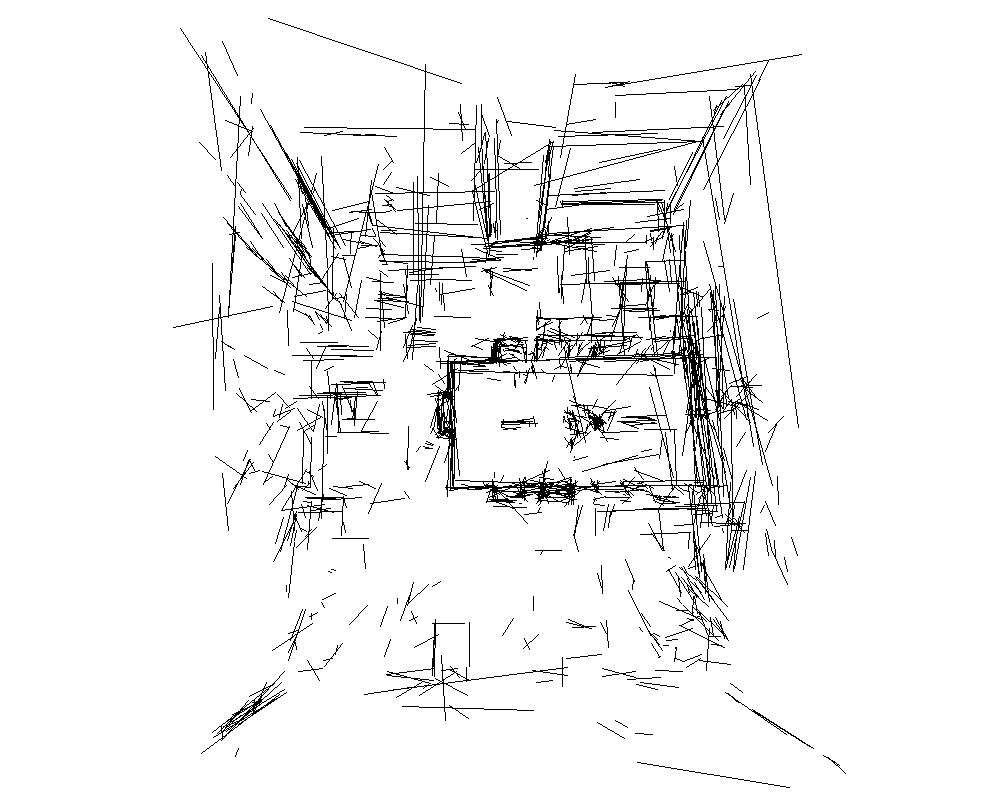}} &
        \parbox[c]{0.18\textwidth}{\centering\includegraphics[width=\linewidth]{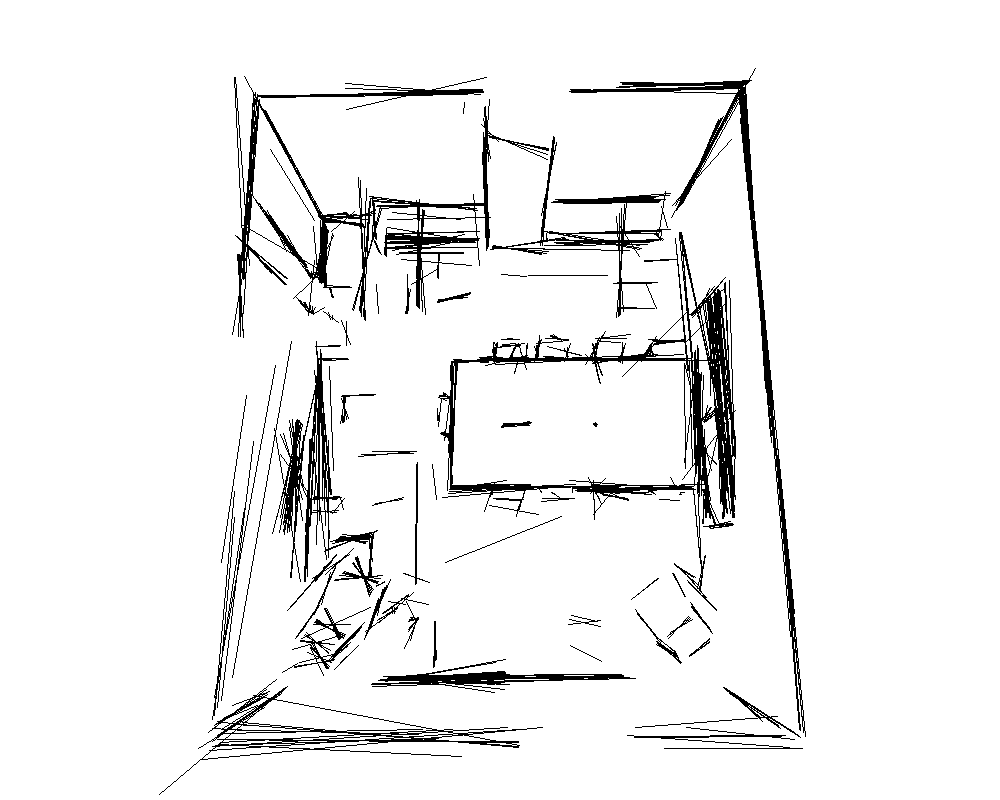}} &
        \parbox[c]{0.18\textwidth}{\centering\includegraphics[width=\linewidth]{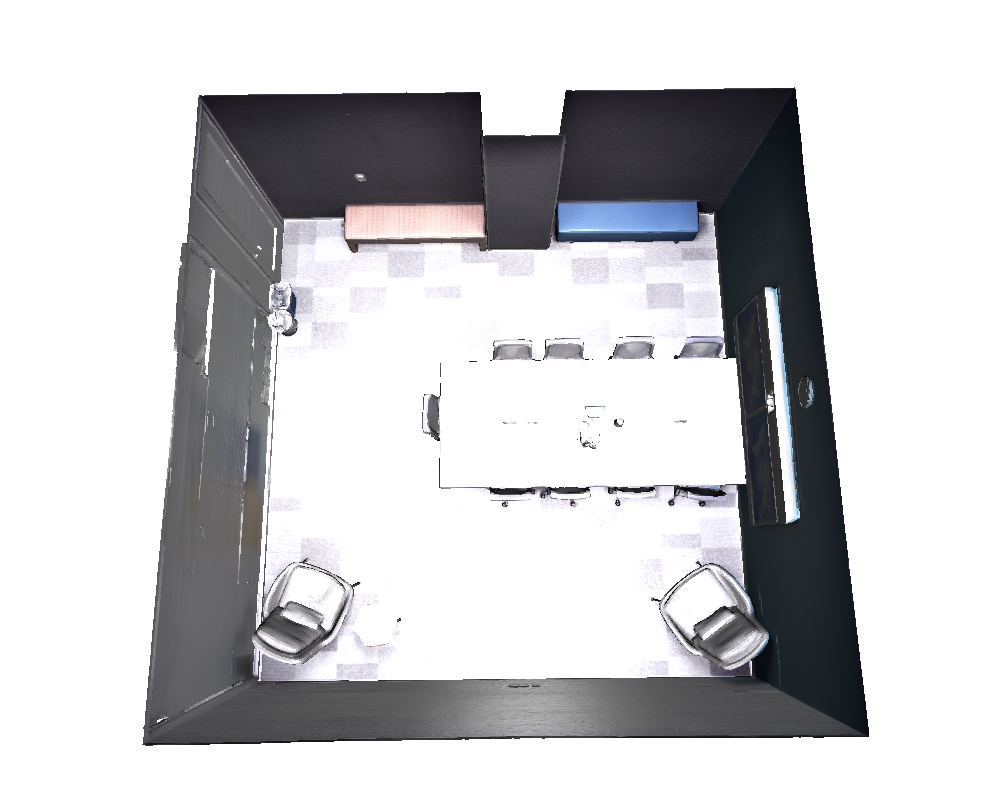}} \\

        \parbox[c]{0.02\textwidth}{\centering\rotatebox{90}{\texttt{office2}}} &
        \parbox[c]{0.18\textwidth}{\centering\includegraphics[width=\linewidth]{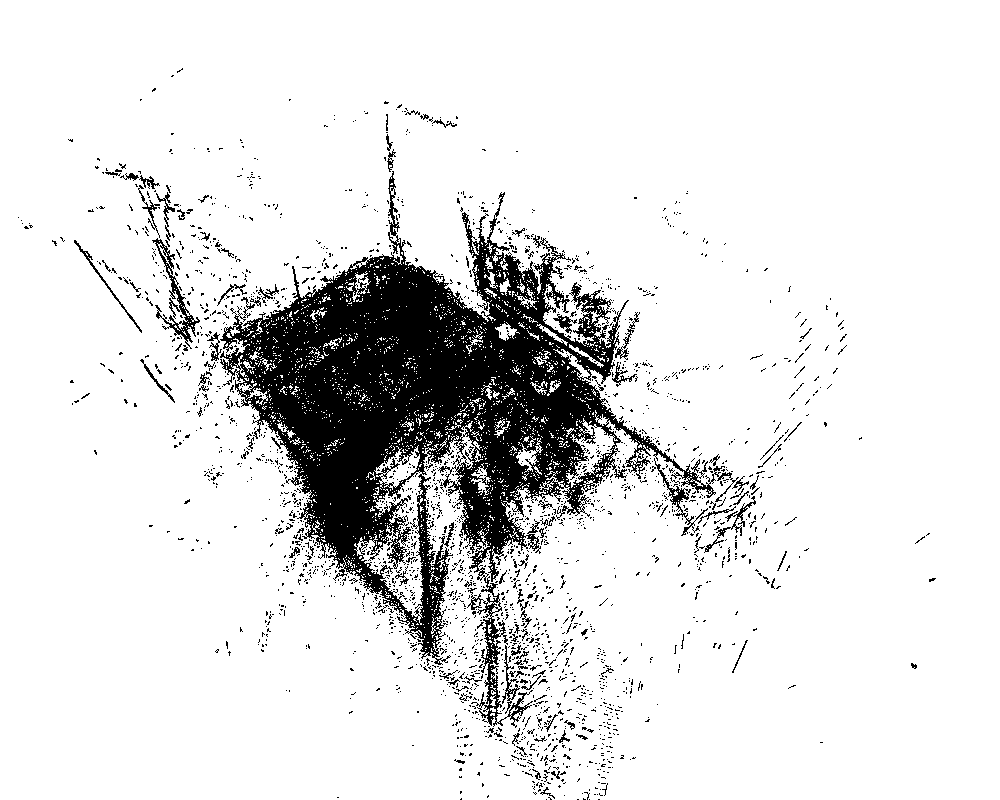}} &
        \parbox[c]{0.18\textwidth}{\centering\includegraphics[width=\linewidth]{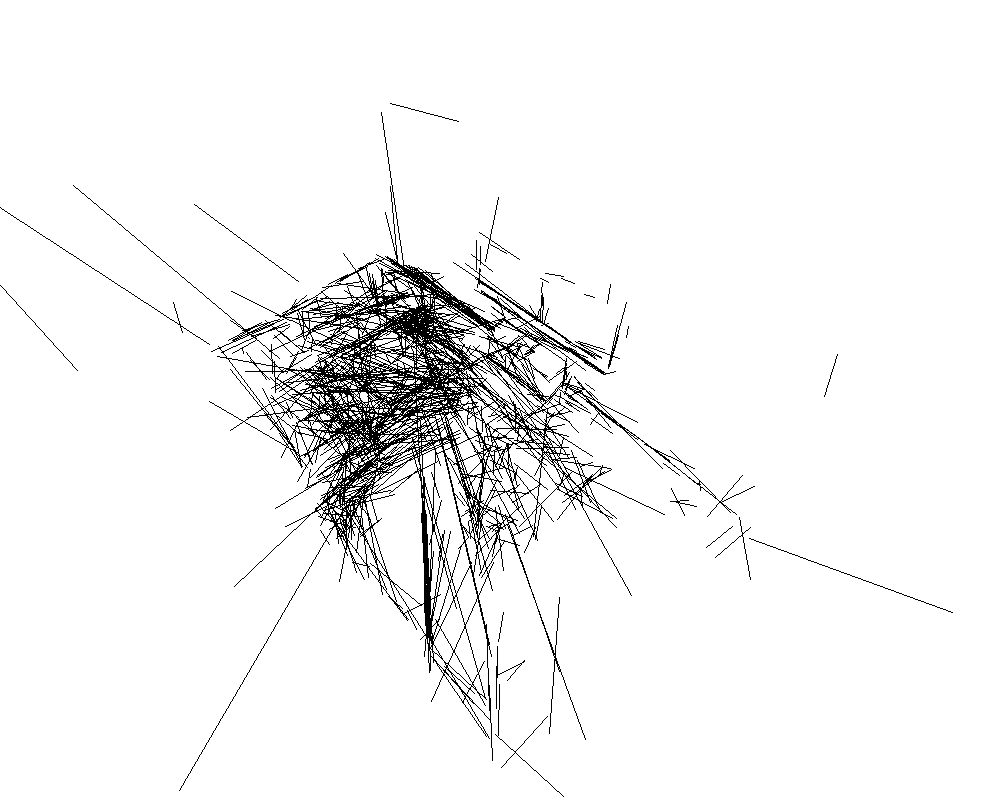}} &
        \parbox[c]{0.18\textwidth}{\centering\includegraphics[width=\linewidth]{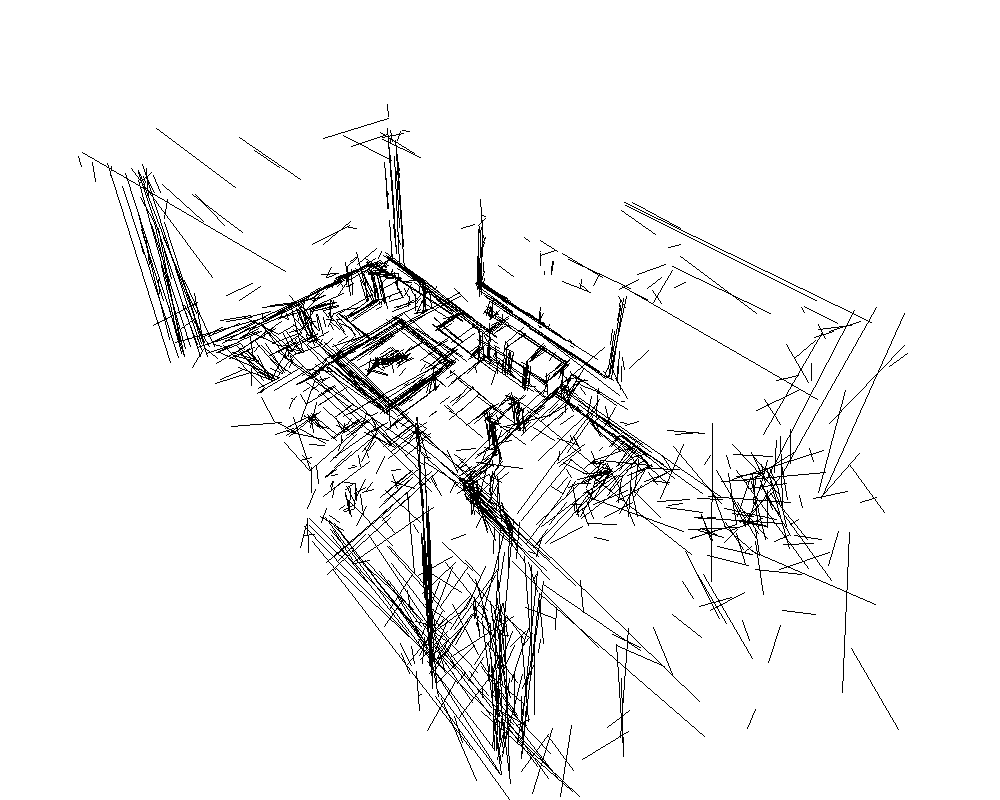}} &
        \parbox[c]{0.18\textwidth}{\centering\includegraphics[width=\linewidth]{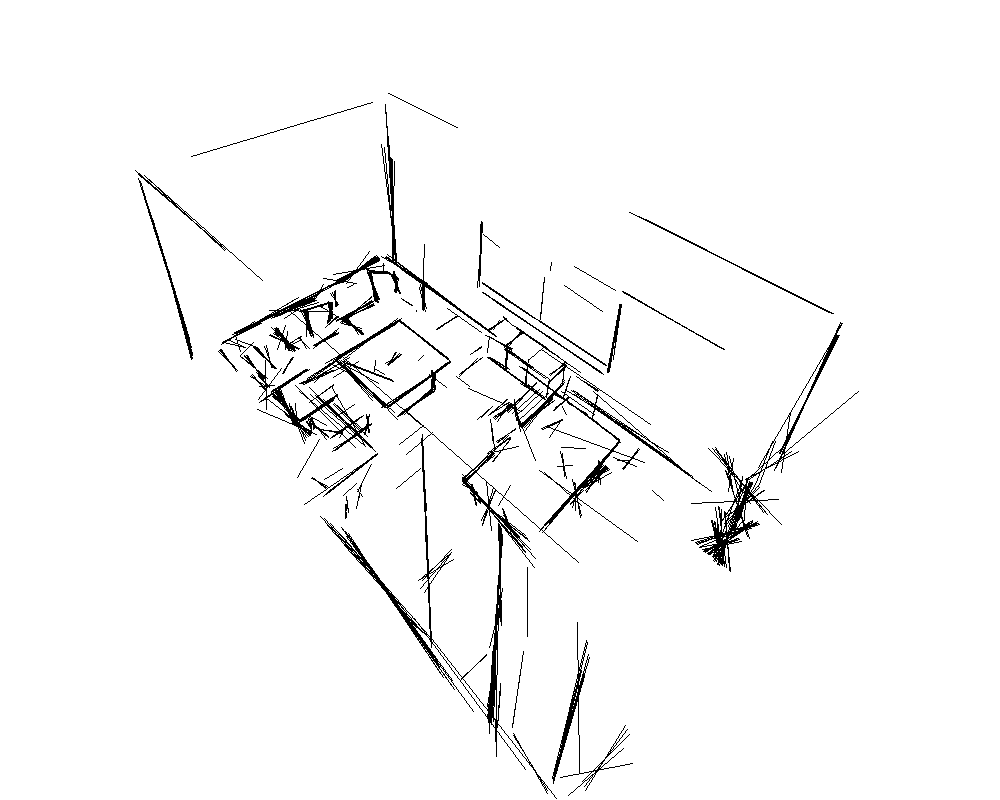}} &
        \parbox[c]{0.18\textwidth}{\centering\includegraphics[width=\linewidth]{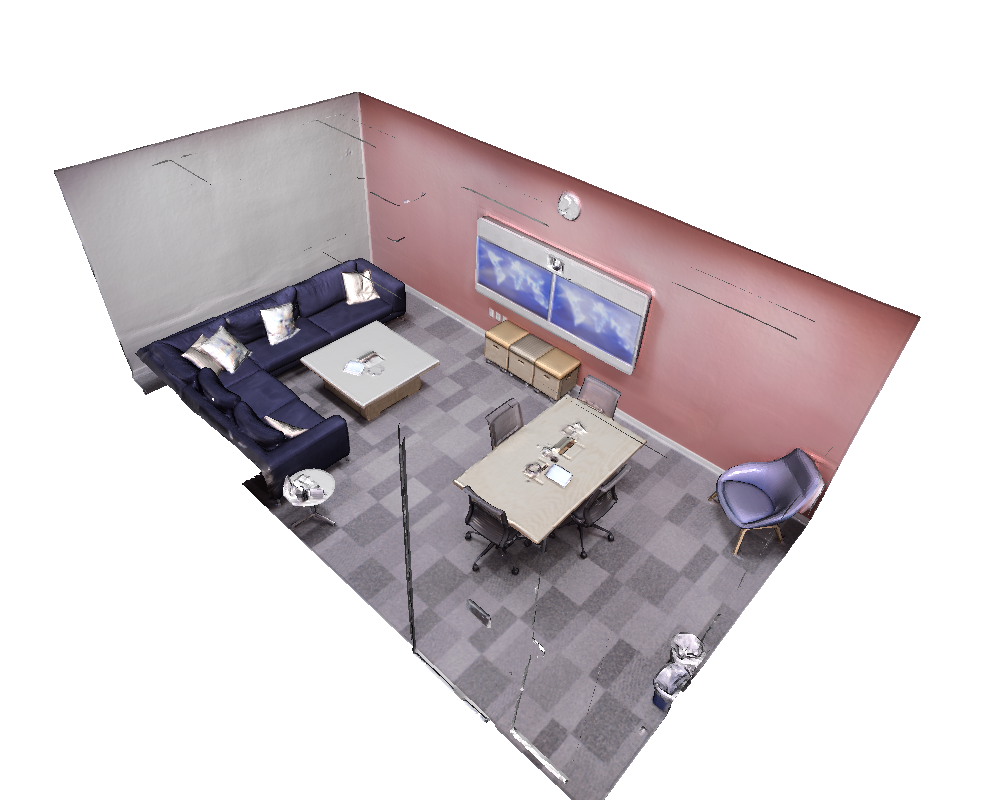}} \\

        \parbox[c]{0.02\textwidth}{\centering\rotatebox{90}{\texttt{office0}}} &
        \parbox[c]{0.18\textwidth}{\centering\includegraphics[width=\linewidth]{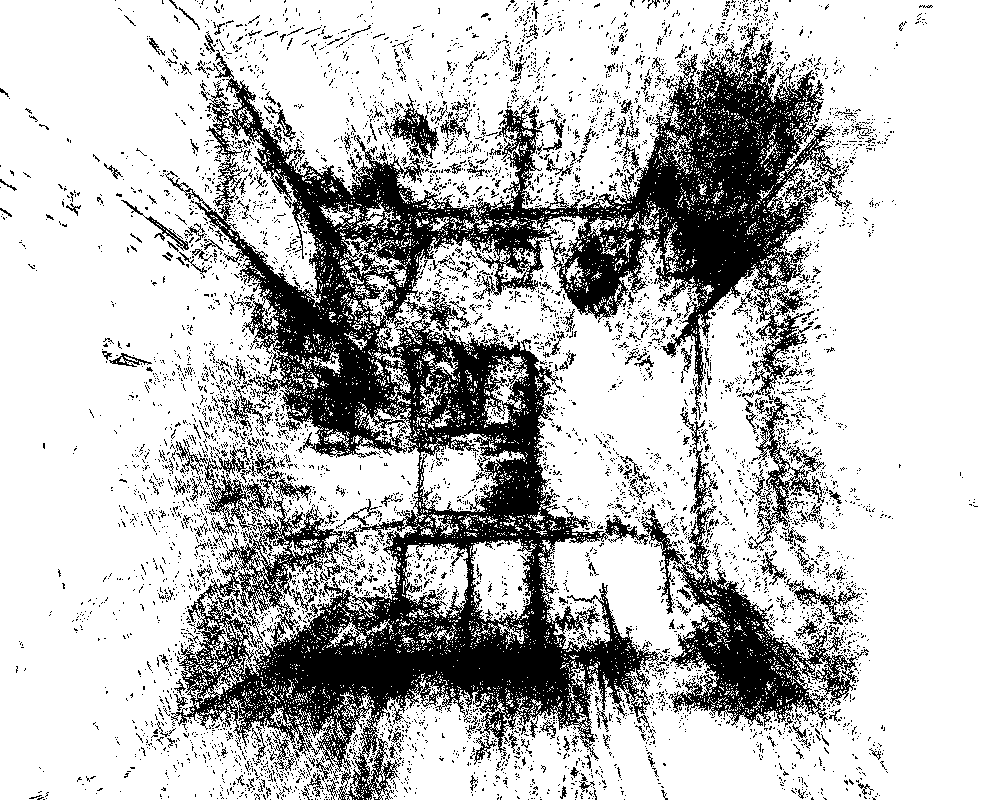}} &
        \parbox[c]{0.18\textwidth}{\centering\includegraphics[width=\linewidth]{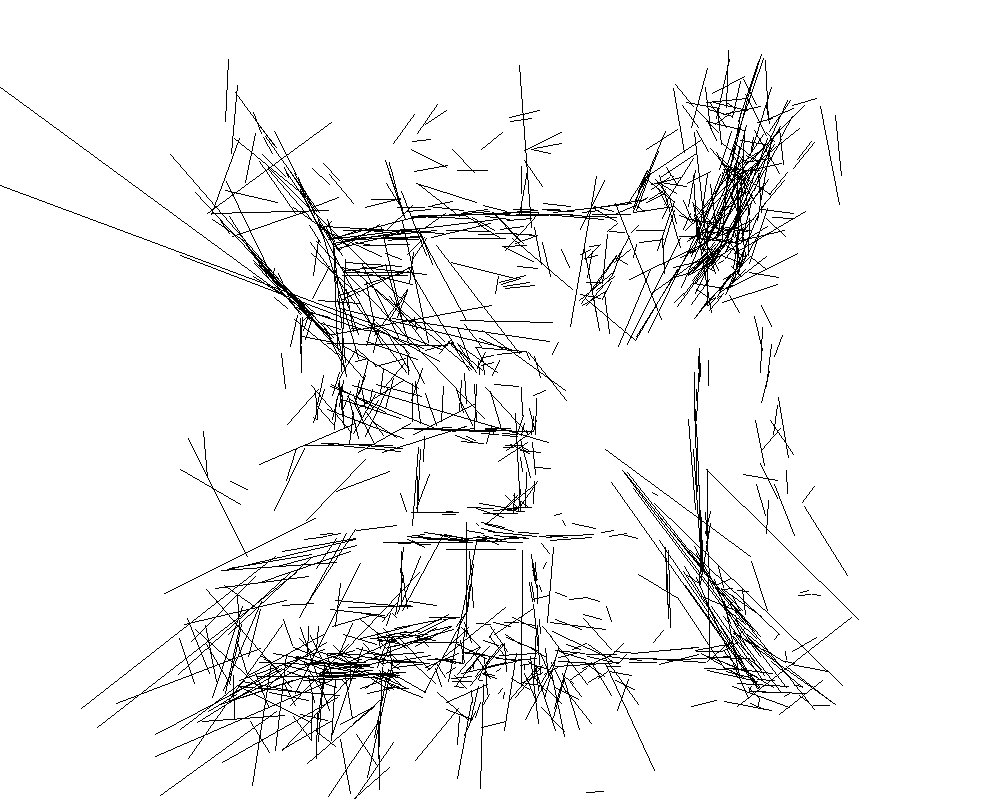}} &
        \parbox[c]{0.18\textwidth}{\centering\includegraphics[width=\linewidth]{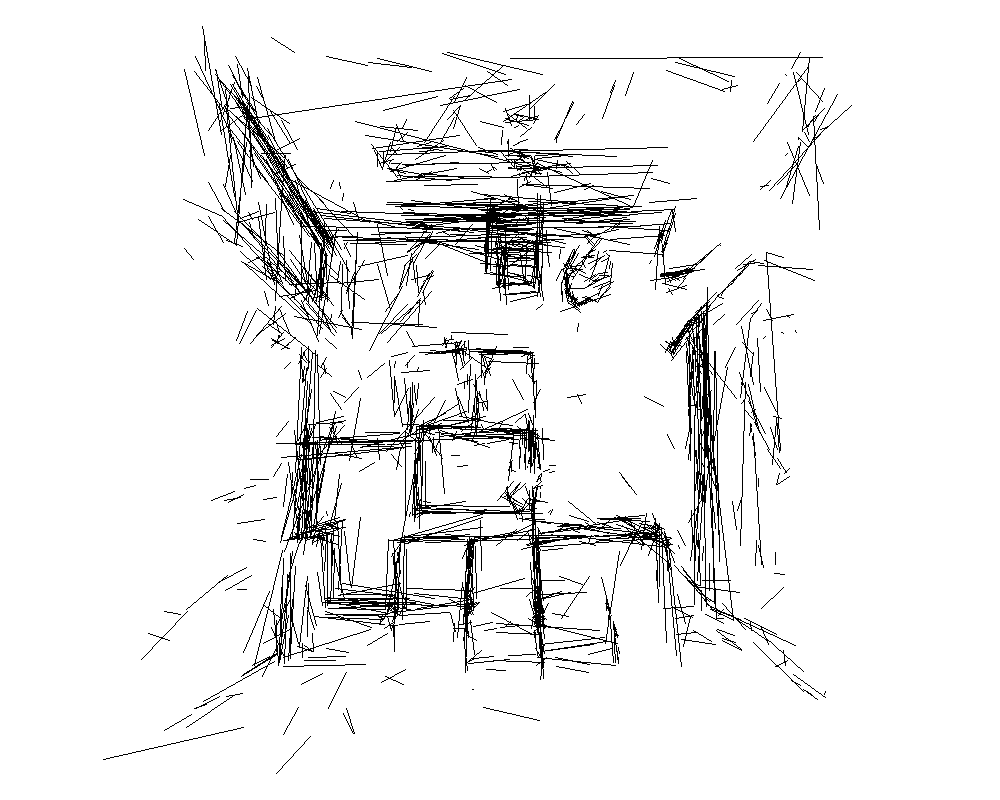}} &
        \parbox[c]{0.18\textwidth}{\centering\includegraphics[width=\linewidth]{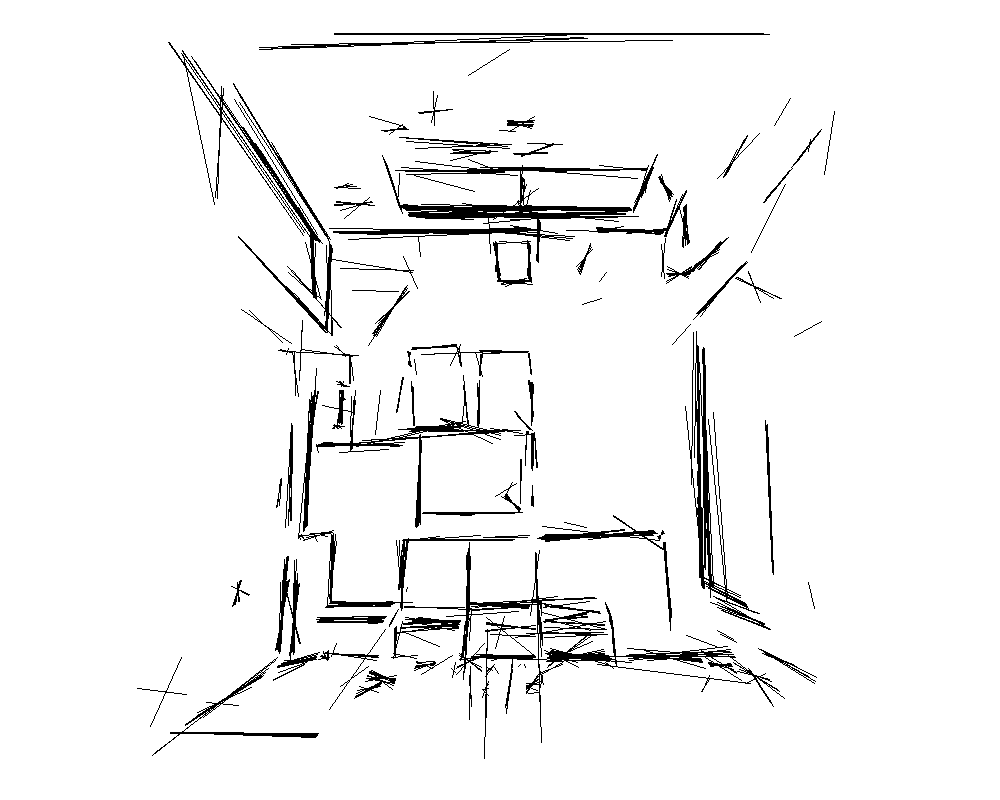}} &
        \parbox[c]{0.18\textwidth}{\centering\includegraphics[width=\linewidth]{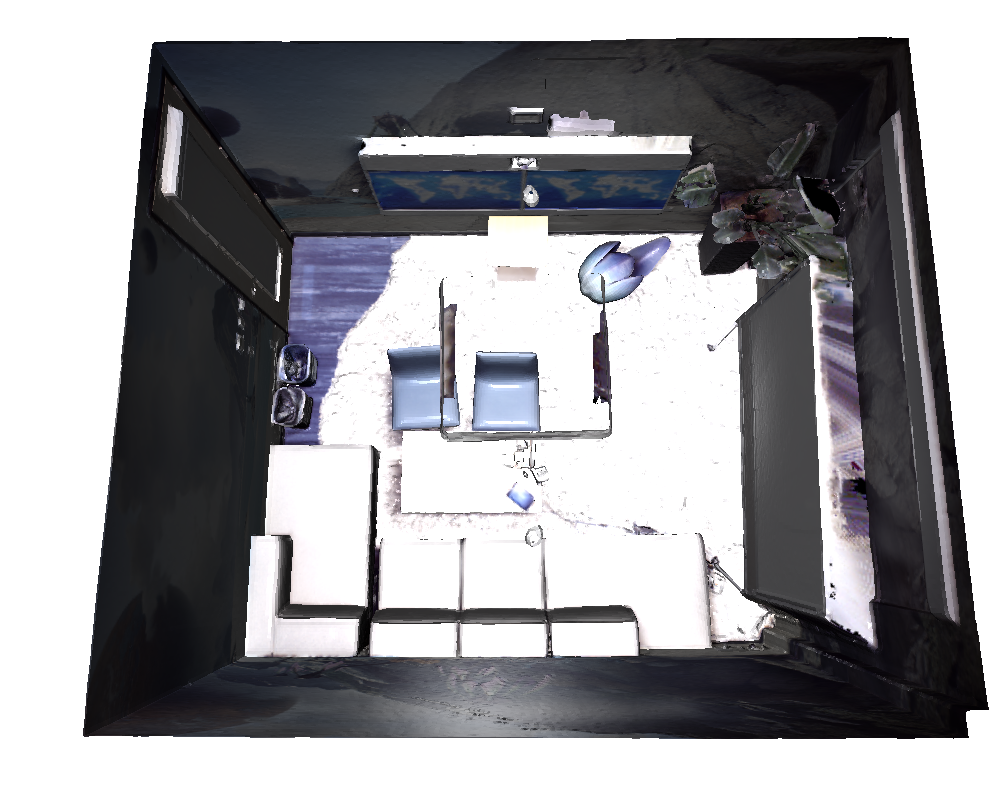}} \\

        & \makebox[0.02\textwidth][c]{EMVS~\cite{emvs}}
        & \makebox[0.18\textwidth][c]{EL-SLAM~\cite{elslam}}
        & \makebox[0.18\textwidth][c]{LIMAP~\cite{limap}}
        & \makebox[0.18\textwidth][c]{Ours}
        & \makebox[0.18\textwidth][c]{GT} \\
    \end{tabular}
    \caption{Qualitative comparison of reconstruction quality on the Replica~\cite{replica} dataset.}
    \label{fig:recon_replica}
\end{figure*}

%% file: figures/recon_tum.tex
\begin{figure*}[t]
    \centering
    \renewcommand{\arraystretch}{1.1}
    \begin{tabular}{@{}c@{\,}c@{\,}c@{\,}c@{\,}c@{}}

        \parbox[c]{0.02\textwidth}{\centering\rotatebox{90}{\texttt{desk2}}} &
        \parbox[c]{0.235\textwidth}{\centering\includegraphics[width=\linewidth]{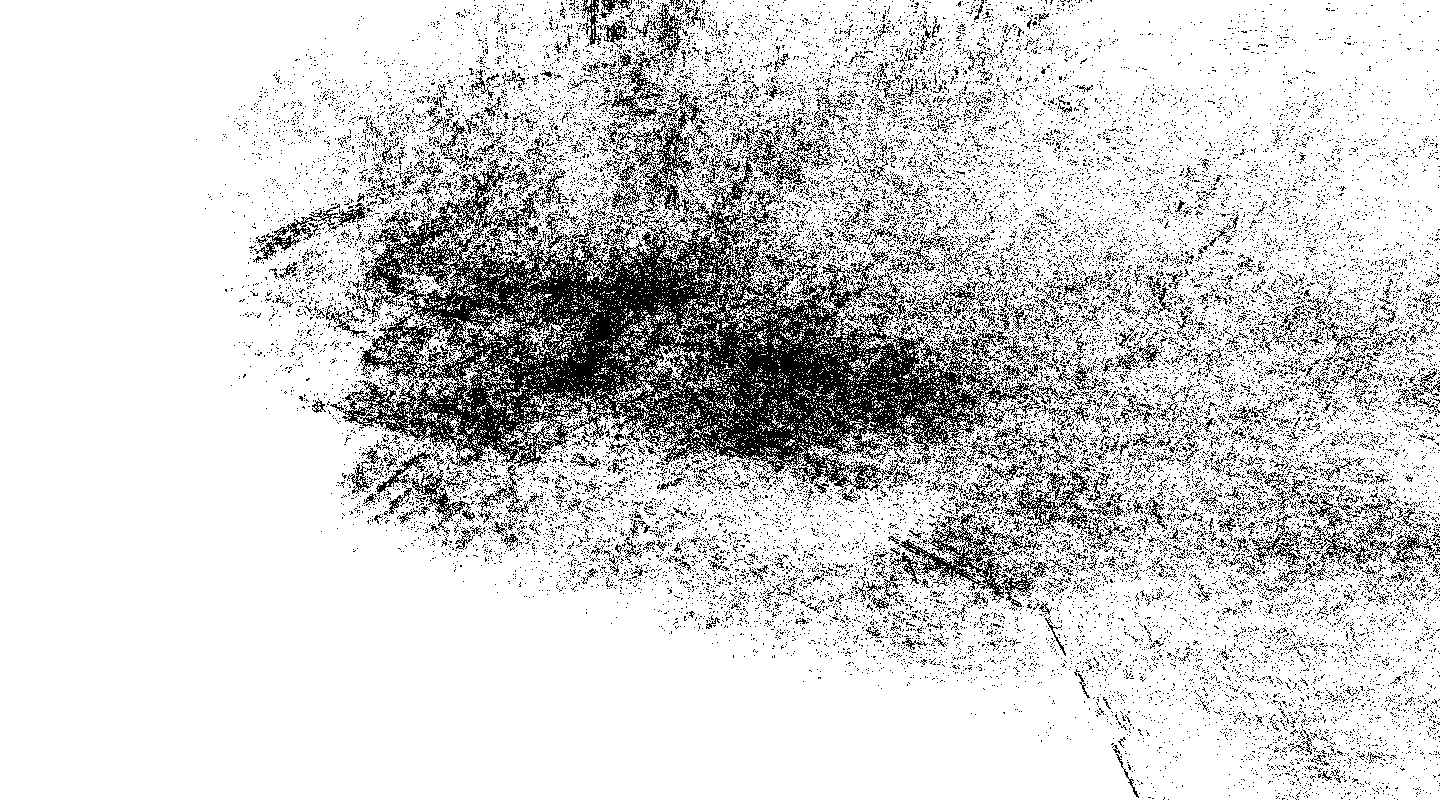}} &
        \parbox[c]{0.235\textwidth}{\centering\includegraphics[width=\linewidth]{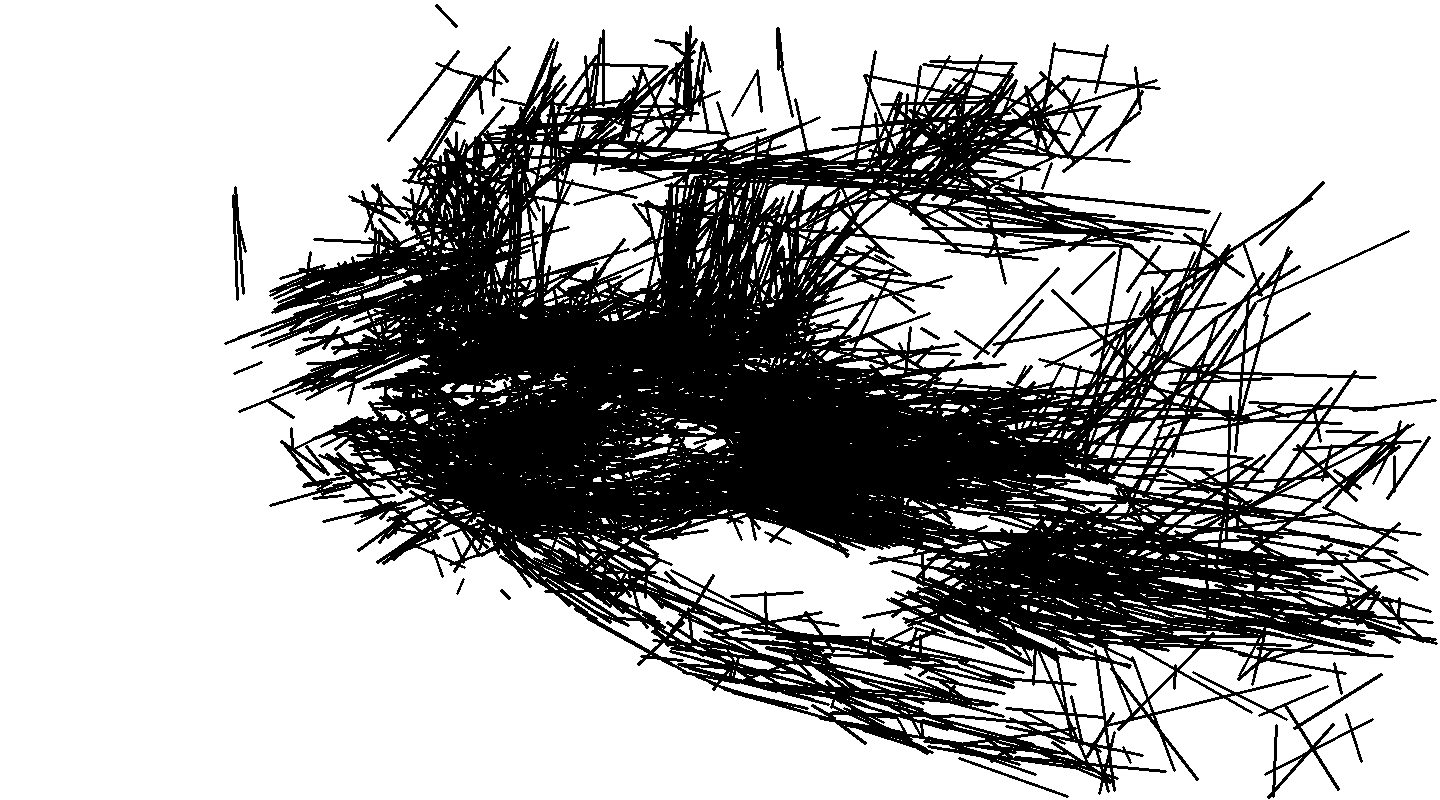}} &
        \parbox[c]{0.235\textwidth}{\centering\includegraphics[width=\linewidth]{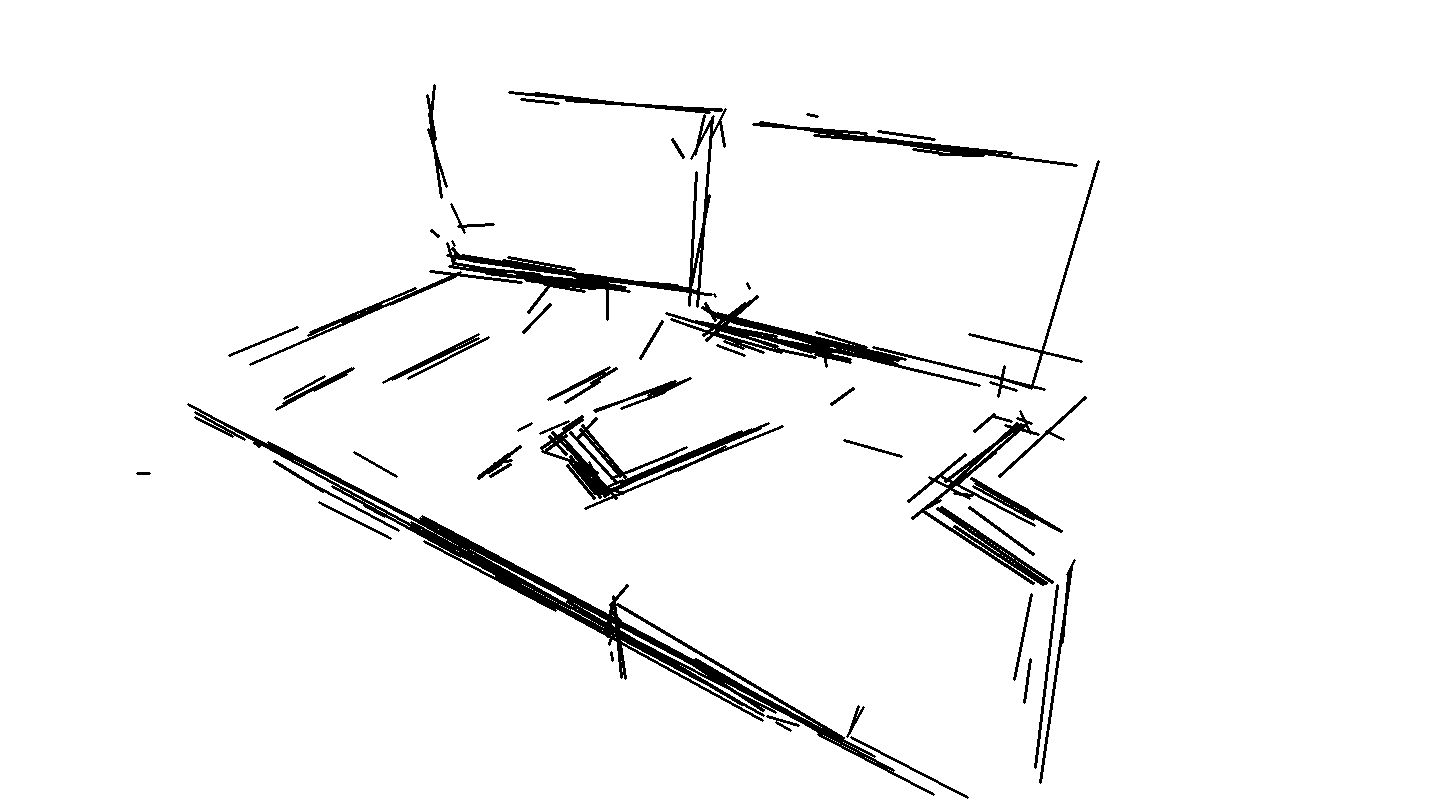}} &
        \parbox[c]{0.235\textwidth}{\centering\includegraphics[width=\linewidth]{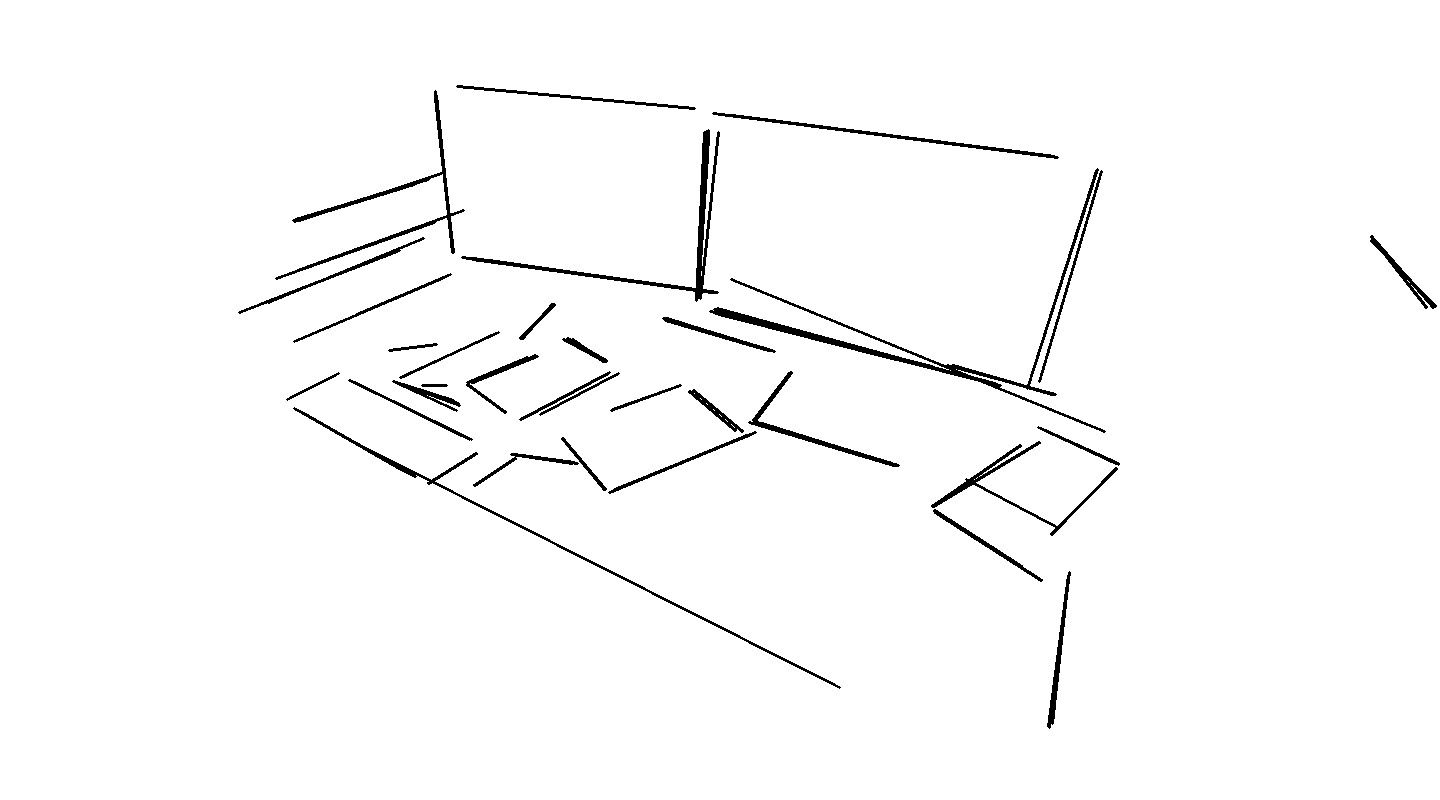}} \\

        \parbox[c]{0.02\textwidth}{\centering\rotatebox{90}{\texttt{desk}}} &
        \parbox[c]{0.235\textwidth}{\centering\includegraphics[width=\linewidth]{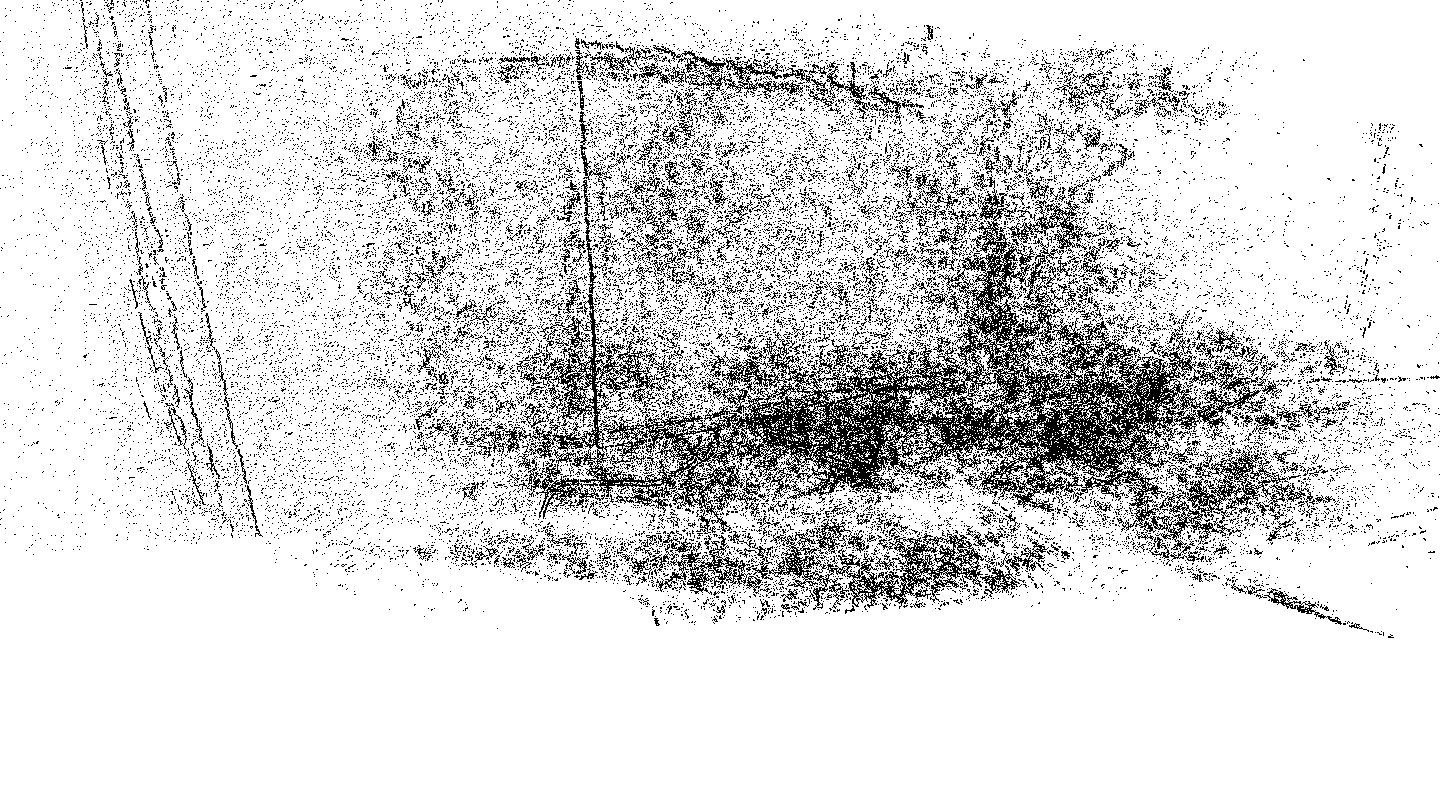}} &
        \parbox[c]{0.235\textwidth}{\centering\includegraphics[width=\linewidth]{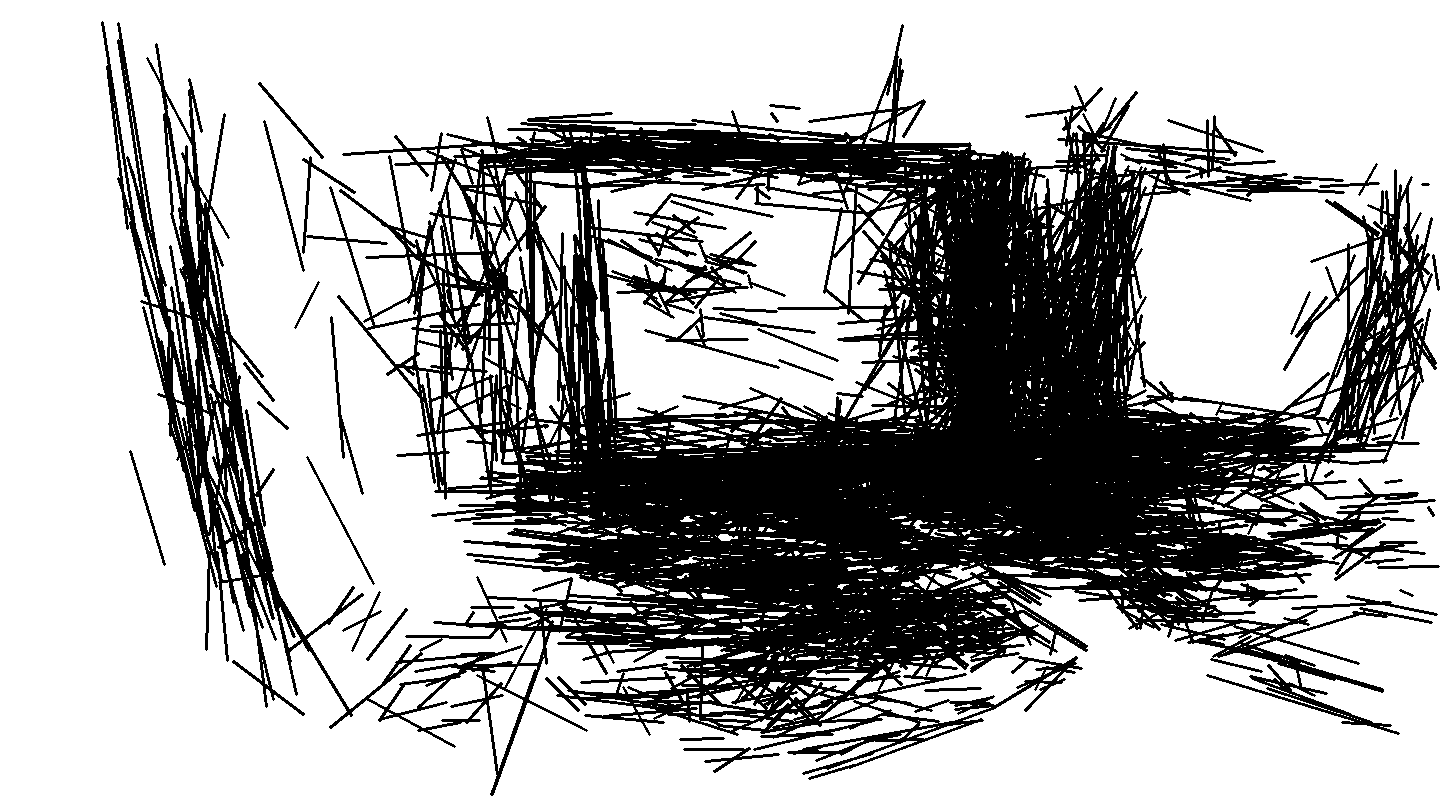}} &
        \parbox[c]{0.235\textwidth}{\centering\includegraphics[width=\linewidth]{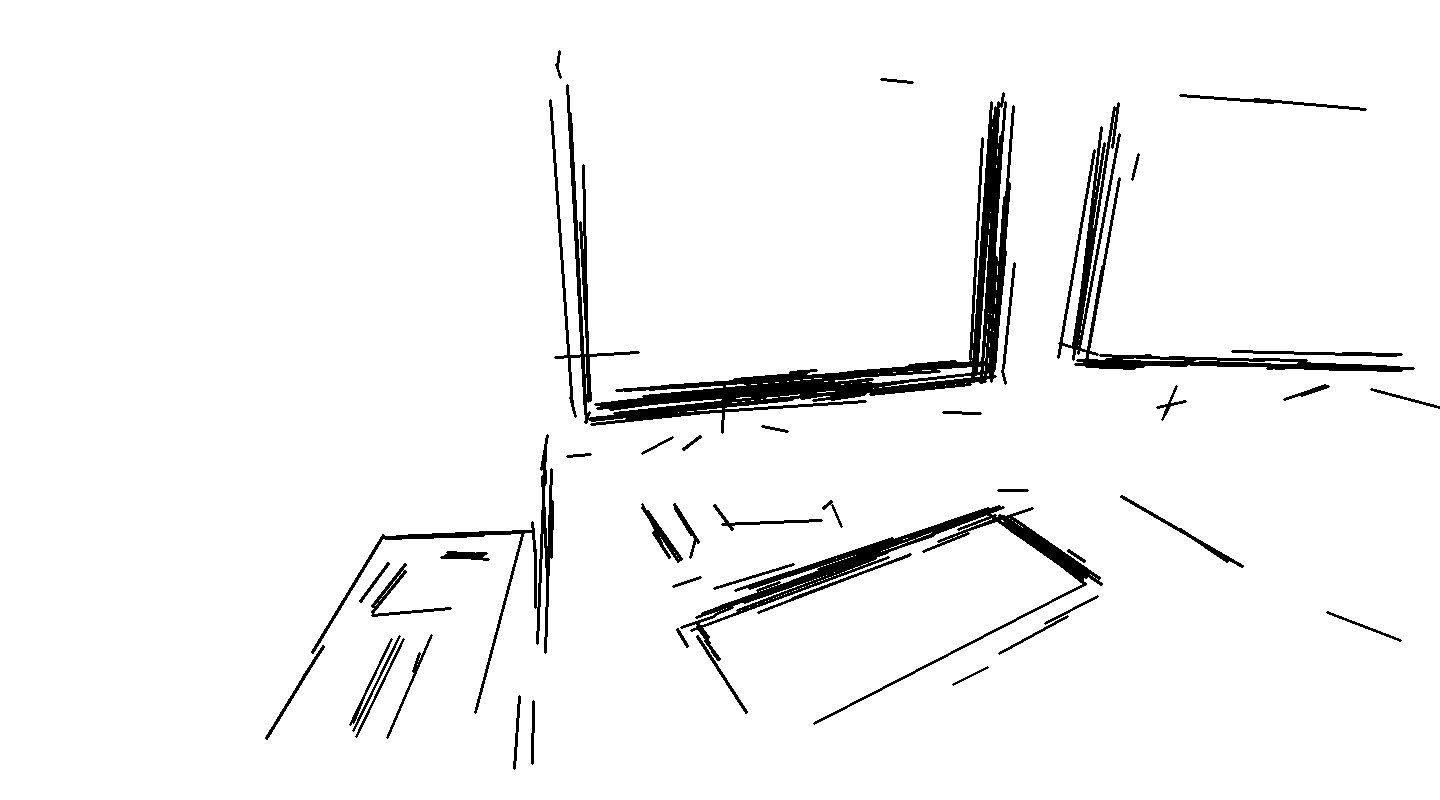}} &
        \parbox[c]{0.235\textwidth}{\centering\includegraphics[width=\linewidth]{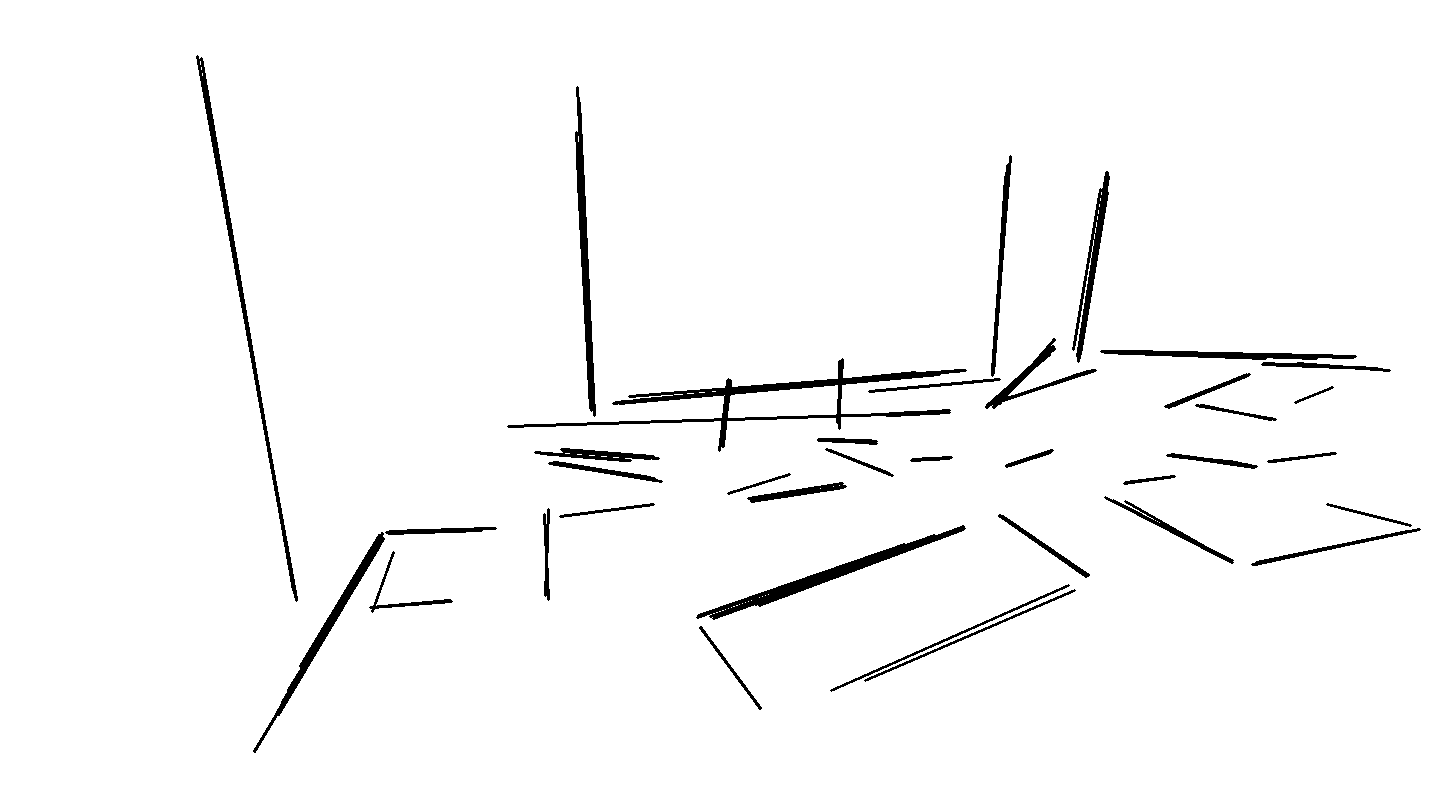}} \\

        & \makebox[0.235\textwidth][c]{EMVS~\cite{emvs}} 
        & \makebox[0.235\textwidth][c]{EL-SLAM~\cite{elslam}} 
        & \makebox[0.235\textwidth][c]{LIMAP~\cite{limap}} 
        & \makebox[0.235\textwidth][c]{Ours} \\
    \end{tabular}
    \caption{Qualitative comparison of reconstruction quality on the TUM-VIE~\cite{tum-vie} dataset.}
    \label{fig:recon_tum}
\end{figure*}

%% file: tables/recon_vector_transpose.tex
\begin{table*}[t]
\caption{Quantitative comparison of reconstruction quality on the VECtor~\cite{vector} dataset. Bold indicates best. \#Rep. indicates the number of lines or points. We report Accuracy and Precision using the ground-truth 3D point map, as we could not obtain a reliable ground-truth 3D edge map due to misalignment between the RGB and depth images.}
\label{tab:recon_vector_transpose}
\centering

\resizebox{0.75\textwidth}{!}{
\begin{tabular}{llcccccc}
\toprule

\textbf{Method} & \textbf{Scene} & \textbf{Acc.↓} & \textbf{P 5↑} & \textbf{P 10↑} & \textbf{P 20↑} & \makecell{\textbf{\#Rep.}} & \makecell{\textbf{Rep.}} \\
\midrule
\multirow{3}{*}[-0.45ex]{{EL-SLAM}~\cite{elslam}}
 & avg  & 221.85 & 0.021 & 0.069 & 0.140 & 1711 & \multirow[c]{3}{*}[-0.45ex]{Line} \\[0.3ex] 
\cdashline{2-7}[0.7pt/2pt]
\addlinespace[0.6ex]
 & \texttt{desk-fast} & 184.96 & 0.020 & 0.062 & 0.121 & 1953 \\
 & \texttt{sofa-fast} & 308.12 & 0.016 & 0.052 & 0.107 & 1654 \\
 & \texttt{desk-normal} & 109.17 & 0.028 & 0.096 & 0.197 &   436 \\
 & \texttt{sofa-normal} & 285.13 & 0.021 & 0.067 & 0.135 &  2799 \\
\midrule
\multirow{3}{*}[-0.45ex]{{LIMAP}~\cite{limap}}
 &  avg & 91.21 & 0.034 & 0.105 & 0.233 & 278 & \multirow{3}{*}[-0.45ex]{Line} \\[0.3ex] 
\cdashline{2-7}[0.7pt/2pt]
\addlinespace[0.6ex]
 & \texttt{desk-fast} & 85.28 & 0.031 & 0.099 & 0.206 & 110 \\
 & \texttt{sofa-fast} & 77.57 & 0.033 & 0.093 & 0.294 & 23 \\
 & \texttt{desk-normal} & 100.89 & \textbf{0.043} & 0.122 & 0.214 &   341 \\
 & \texttt{sofa-normal} & 101.10 & 0.029 & 0.104 & 0.218 &   637 \\
\midrule
\multirow{3}{*}[-0.45ex]{{EMVS}~\cite{emvs}}
 & avg  & 271.65 & 0.022 & 0.069 & 0.138 & 165313 & \multirow{3}{*}[-0.45ex]{Point}\\[0.3ex] 
\cdashline{2-7}[0.7pt/2pt]
\addlinespace[0.6ex]
 & \texttt{desk-fast} & 189.96 & 0.022 & 0.068 & 0.131 & 153856 \\
 & \texttt{sofa-fast} & 361.92 & 0.015 & 0.047 & 0.095 & 190062 \\
 & \texttt{desk-normal} & 182.56 & 0.029 & 0.095 & 0.185 &  64503 \\
 & \texttt{sofa-normal} & 352.17 & 0.021 & 0.068 & 0.140 & 252831 \\
\midrule
\multirow{3}{*}[-0.45ex]{Ours}
 & avg  & \textbf{58.46} & \textbf{0.049} & \textbf{0.173} & \textbf{0.359} & 835 & \multirow{3}{*}[-0.45ex]{Line} \\[0.3ex] 
\cdashline{2-7}[0.7pt/2pt]
\addlinespace[0.6ex]
 & \texttt{desk-fast} & \textbf{42.31} & \textbf{0.069} & \textbf{0.190} & \textbf{0.350} & 73  \\
 & \texttt{sofa-fast} & \textbf{58.20} & \textbf{0.044} & \textbf{0.186} & \textbf{0.471} & 31  \\
 & \texttt{desk-normal} & \textbf{58.20} & 0.041 & \textbf{0.131} & \textbf{0.252} &  1452 \\
 & \texttt{sofa-normal} & \textbf{75.14} & \textbf{0.043} & \textbf{0.183} & \textbf{0.361} &  1785 \\
 
\bottomrule
\end{tabular}}
\end{table*}

%% file: figures/recon_vector.tex
\begin{figure*}[t]
    \centering
    \renewcommand{\arraystretch}{1.1}
    \begin{tabular}{@{}c@{\,}c@{\,}c@{\,}c@{\,}c@{}}

        \includegraphics[width=0.19\textwidth]{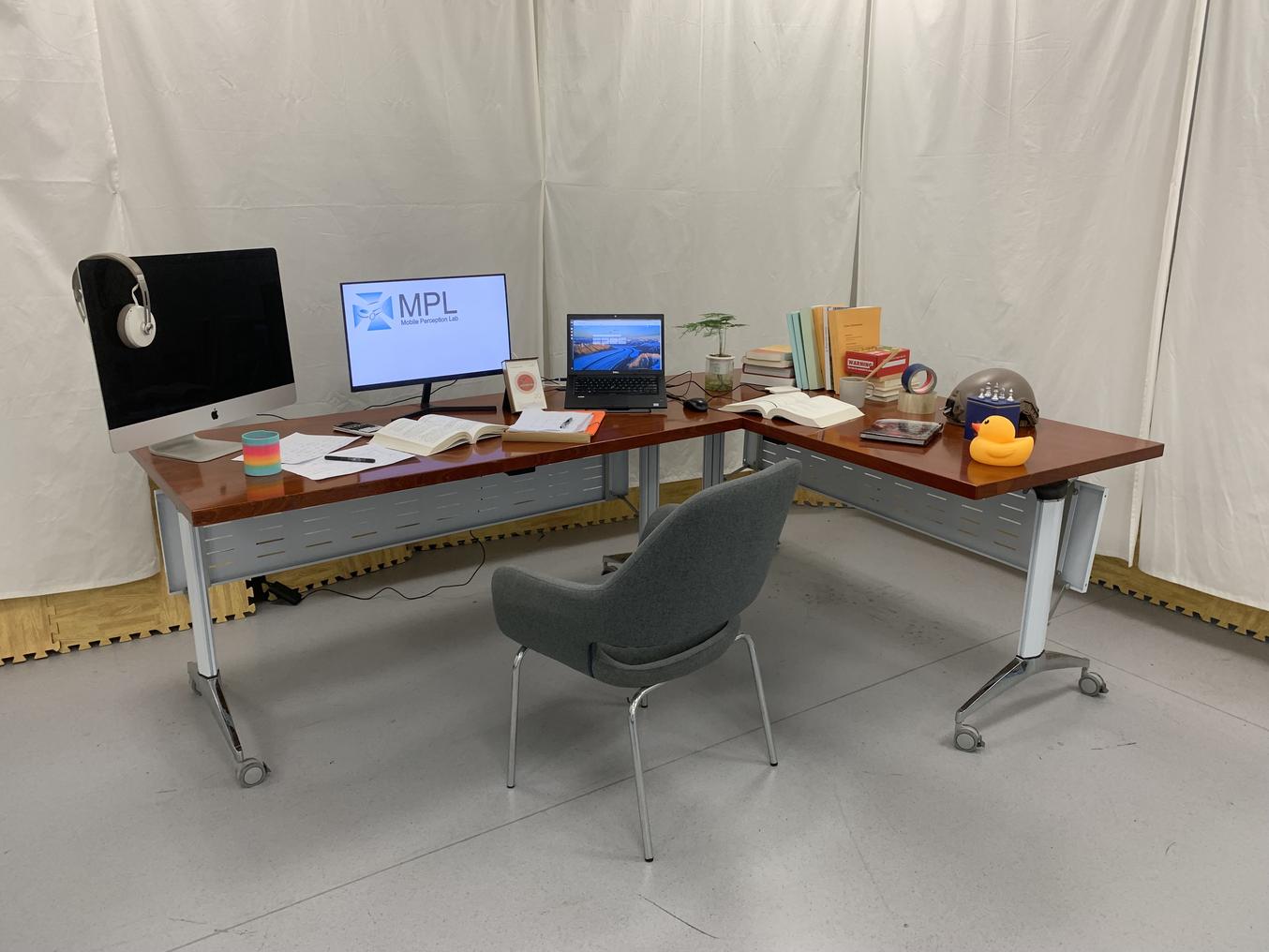} &
        \includegraphics[width=0.19\textwidth]{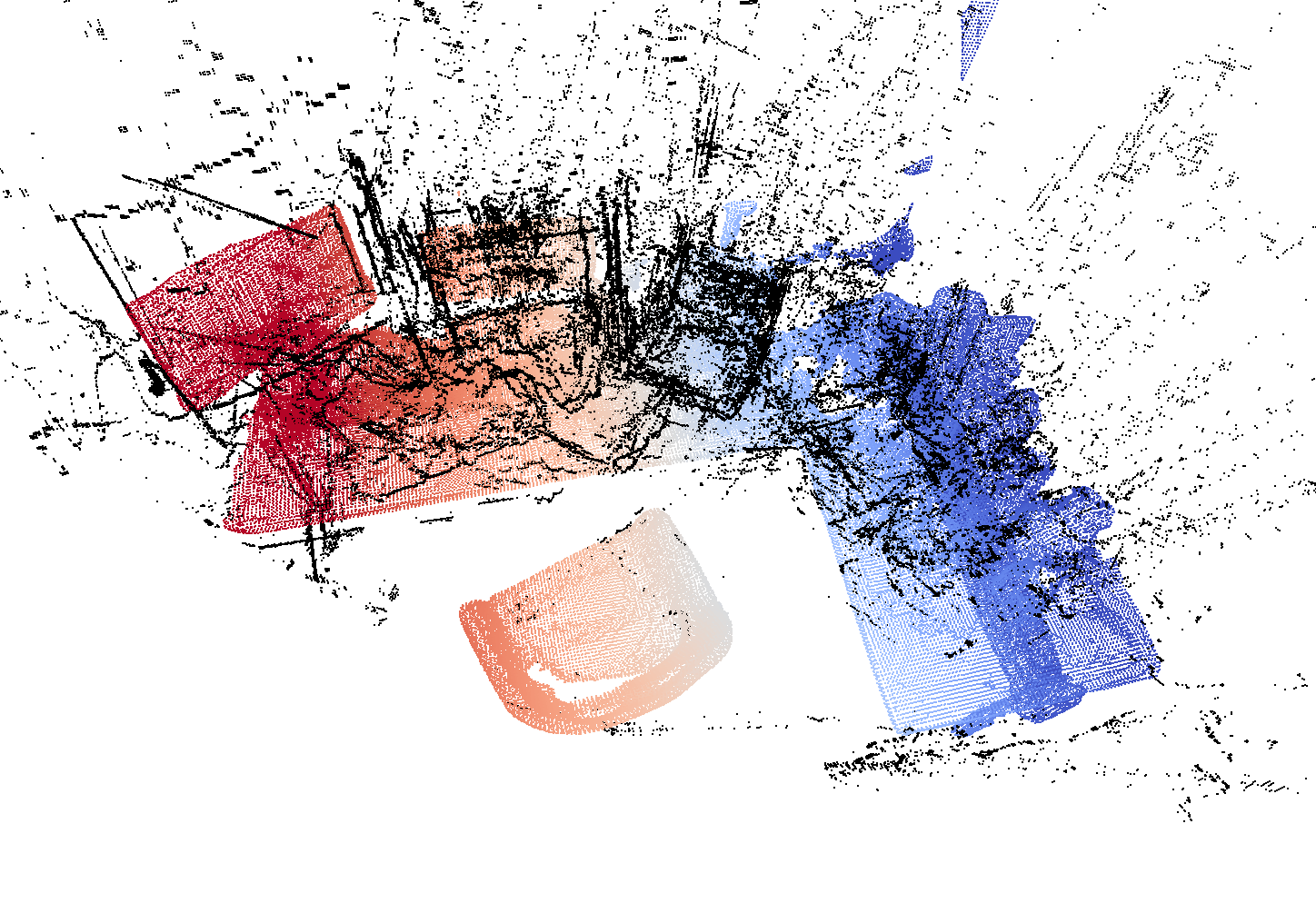} &
        \includegraphics[width=0.19\textwidth]{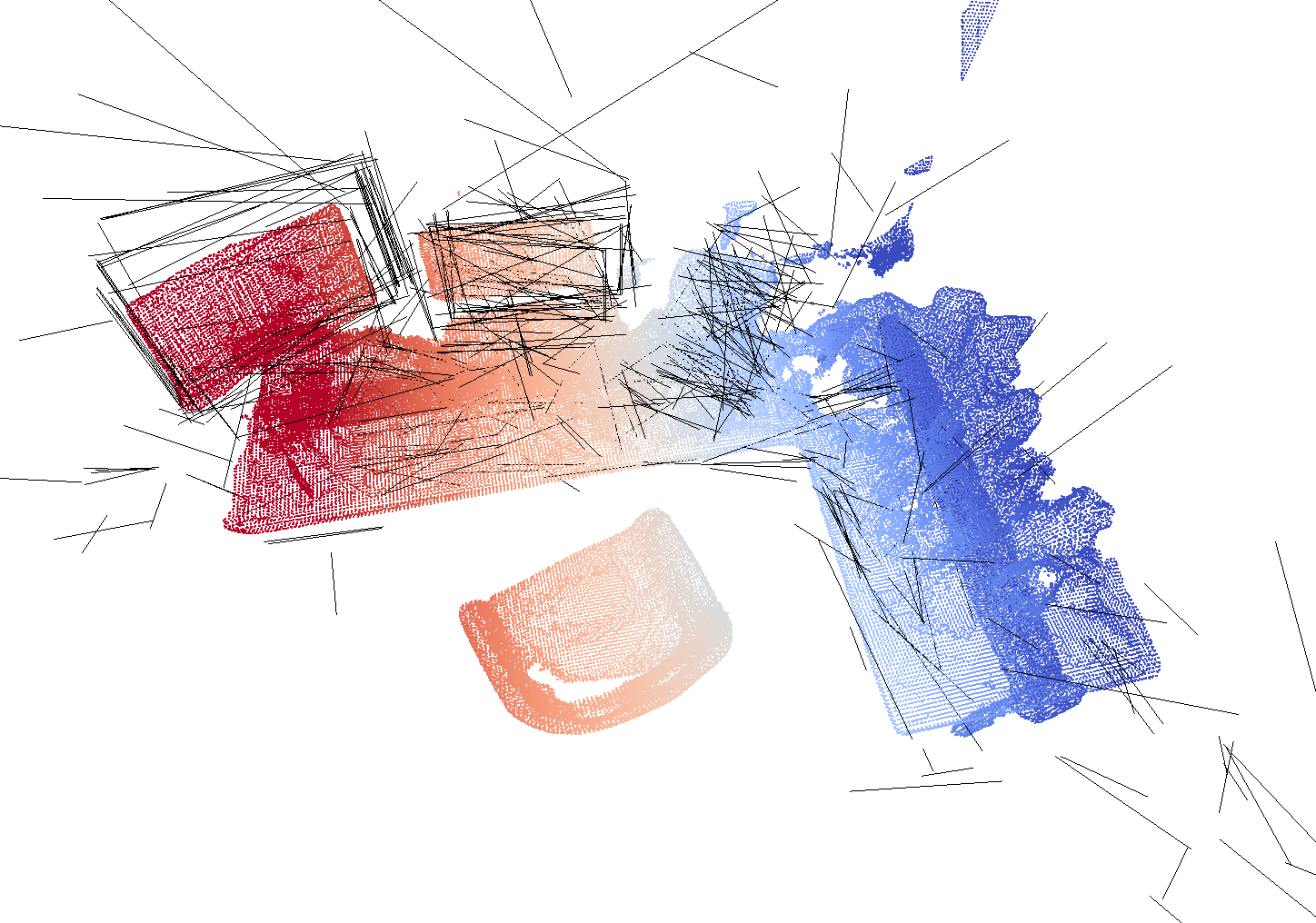} &
        \includegraphics[width=0.19\textwidth]{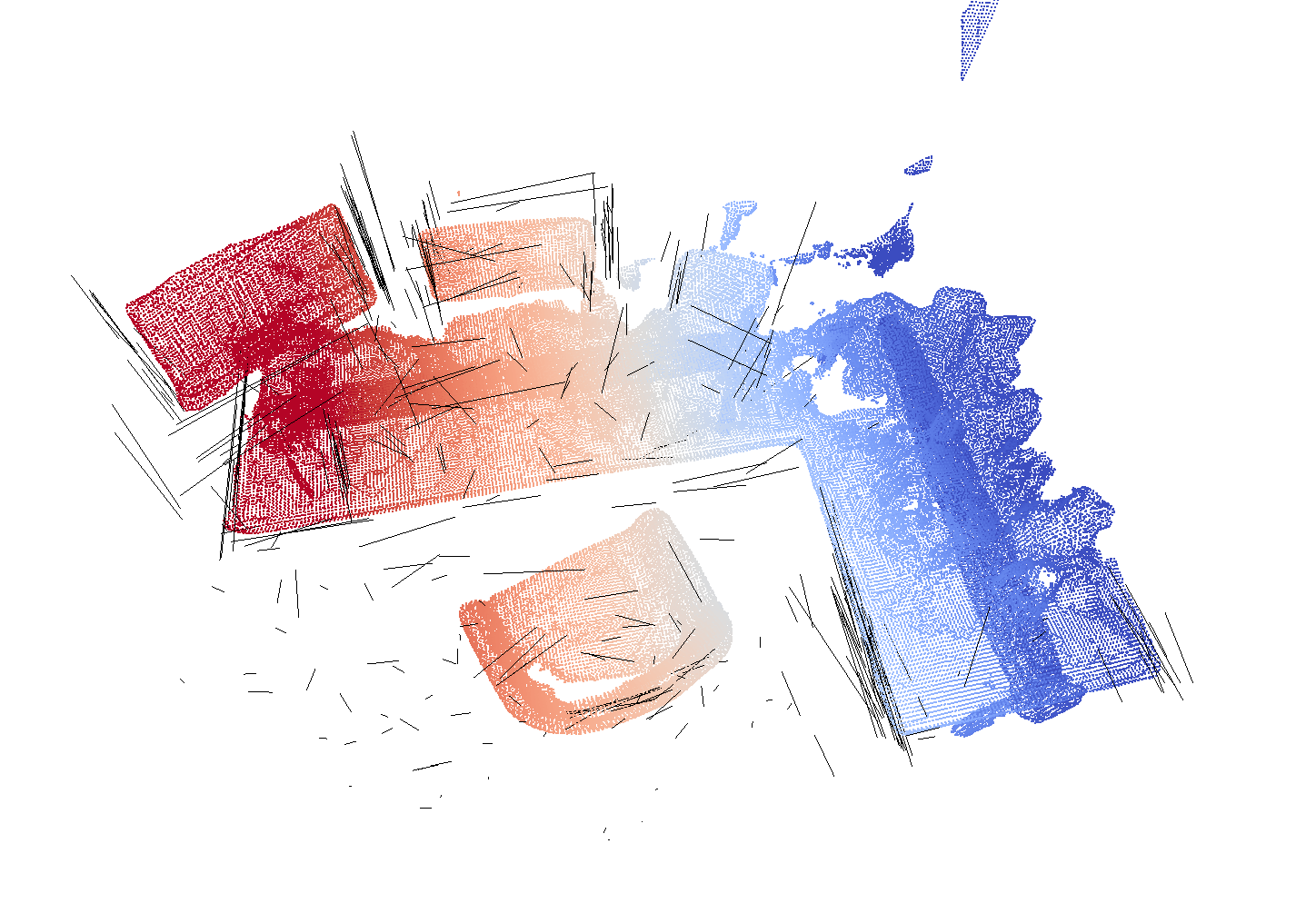} &
        \includegraphics[width=0.19\textwidth]{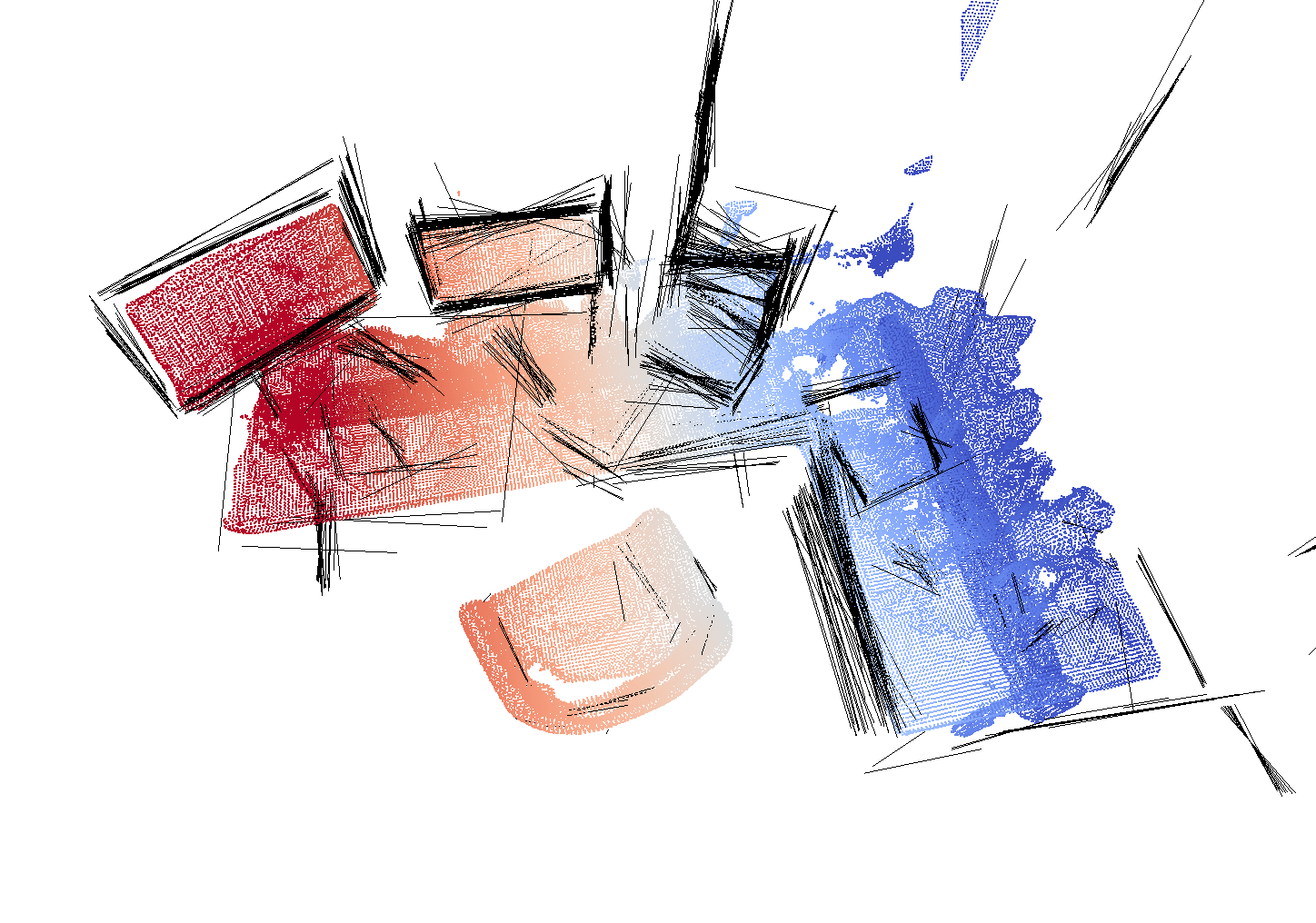} \\

        \makebox[0.19\textwidth][c]{Scene} &
        \makebox[0.19\textwidth][c]{EMVS~\cite{emvs}} &
        \makebox[0.19\textwidth][c]{EL-SLAM~\cite{elslam}} &
        \makebox[0.19\textwidth][c]{LIMAP~\cite{limap}} &
        \makebox[0.19\textwidth][c]{Ours} \\
    \end{tabular}
    \caption{Qualitative comparison of reconstruction quality on \texttt{desk-normal} of the VECtor~\cite{vector} dataset. The first column shows the input scene.
    The ground-truth map, in color, is overlaid on the reconstructed map, represented by black points or lines.
    }
    \label{fig:recon_vector}
    \vspace{-3mm}
\end{figure*}

%% file: figures/registration.tex
\begin{figure*}[!tbp]
    \centering
    \renewcommand{\arraystretch}{1.1}
    \begin{tabular}{@{}c@{\,}c@{\,}c@{\,}c@{}}

        \includegraphics[width=0.24\textwidth]{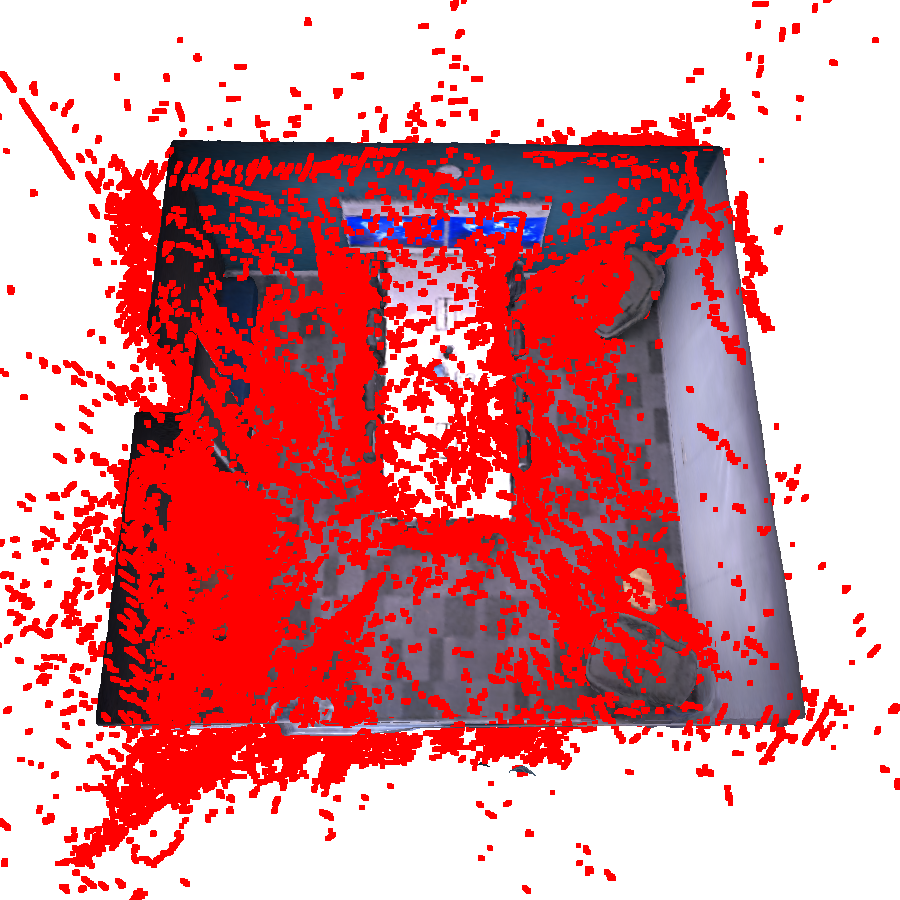} &
        \includegraphics[width=0.24\textwidth]{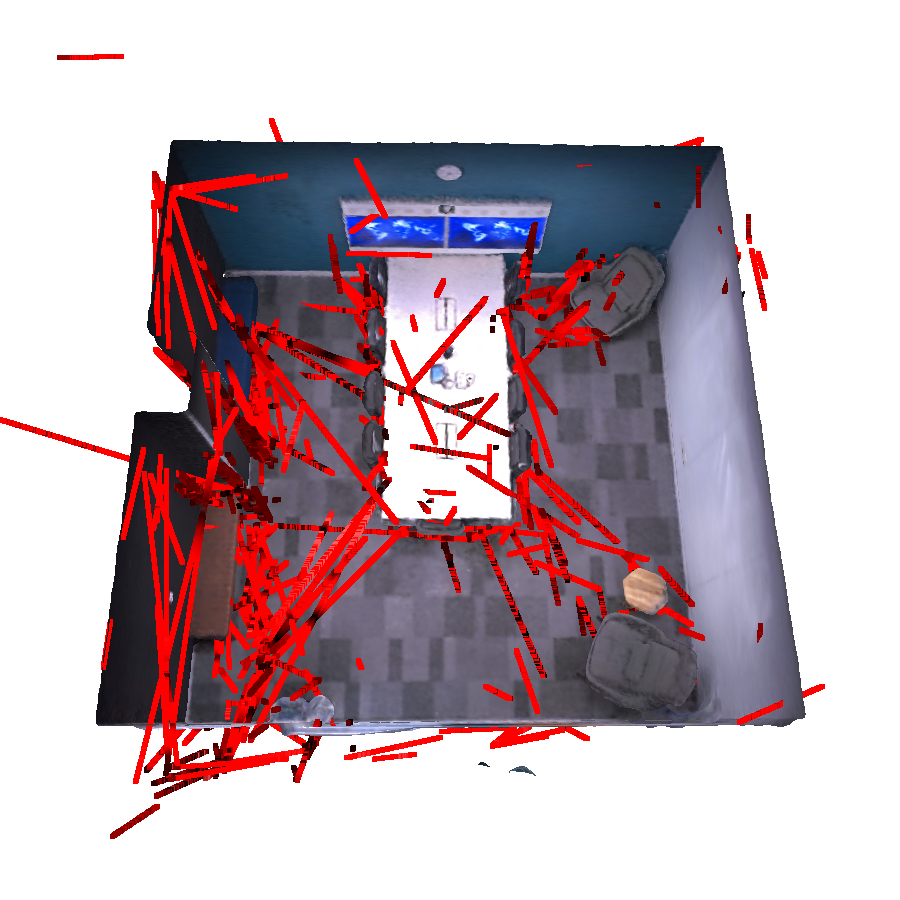} &
        \includegraphics[width=0.24\textwidth]{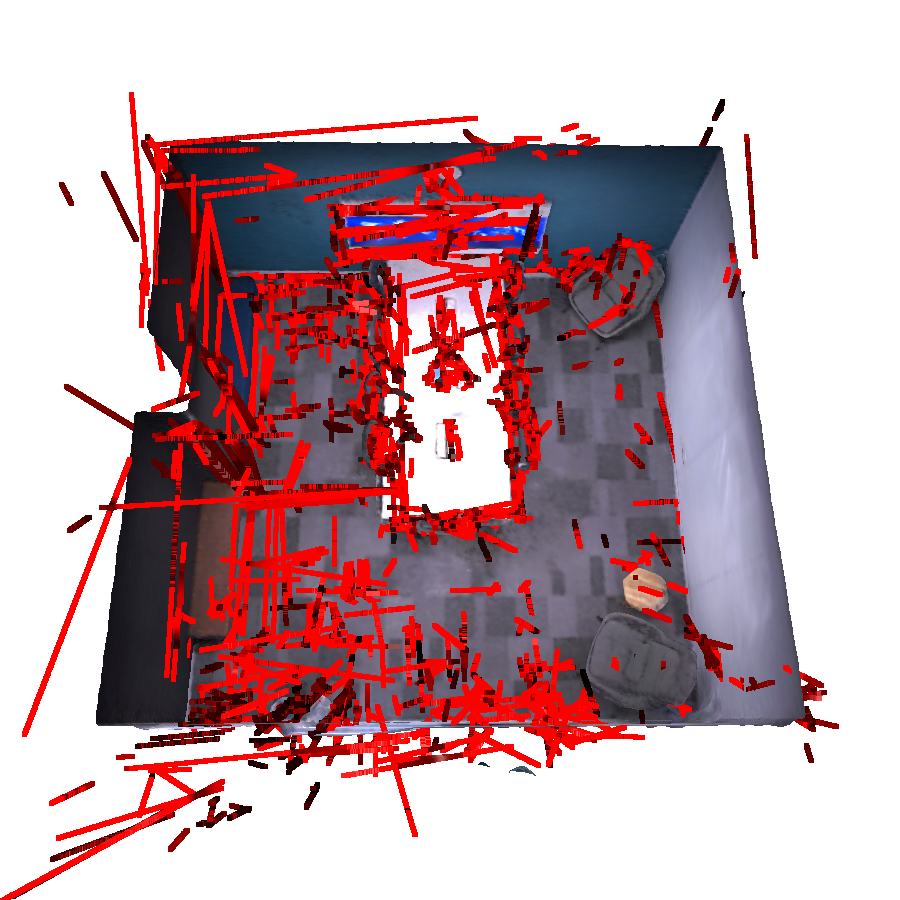} &
        \includegraphics[width=0.24\textwidth]{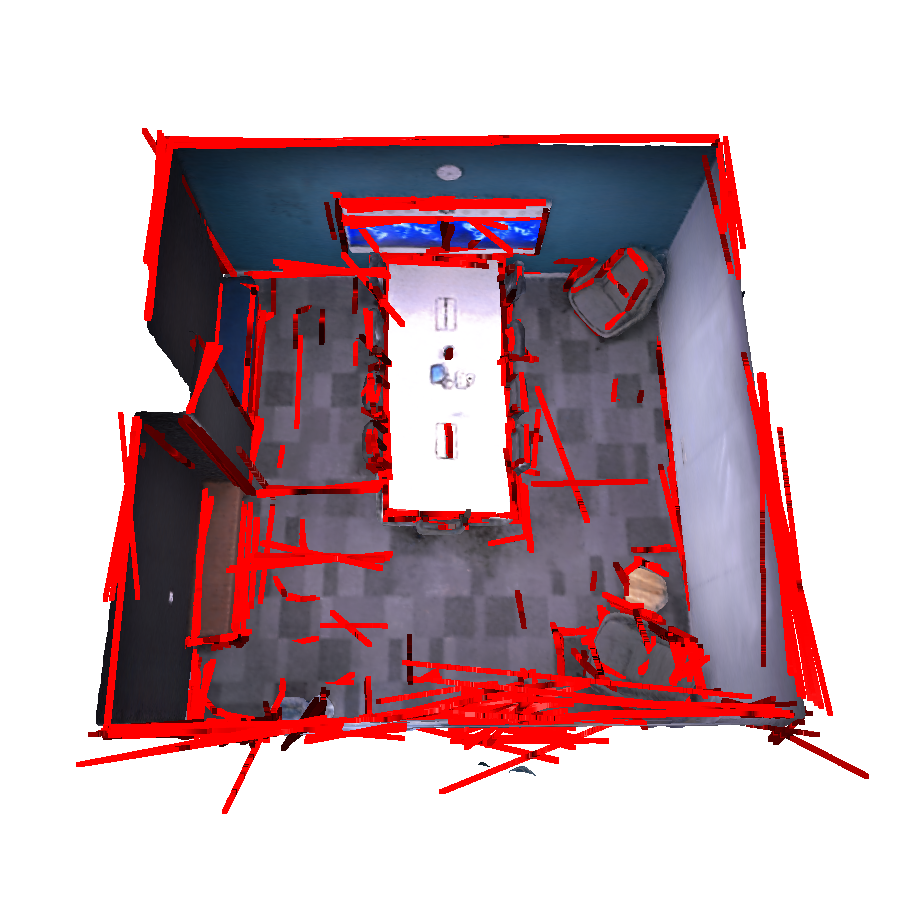} \\

        \makebox[0.24\textwidth][c]{EMVS~\cite{emvs}} &
        \makebox[0.24\textwidth][c]{EL-SLAM~\cite{elslam}} &
        \makebox[0.24\textwidth][c]{LIMAP~\cite{limap}} &
        \makebox[0.24\textwidth][c]{Ours} \\

    \end{tabular}
    \caption{Qualitative comparison of registration on \texttt{office4} of the Replica~\cite{replica} dataset. The background shows a dense RGB map reconstructed with RGB-D SLAM, and the red-colored map represents the reconstruction results after registration.}
    \label{fig:registration}
\end{figure*}

%% file: tables/registration_replica_transpose.tex
\begin{table}[t]
\caption{Quantitative comparison of registration performance of the Replica~\cite{replica} dataset. Bold indicates best.}
\label{tab:registration_replica_transpose}
\centering

\resizebox{0.4\textwidth}{!}{

\begin{tabular}{llcc}
\toprule
\textbf{Method} & \textbf{Scene} & \textit{R} error ($^\circ$) ↓ & \textit{t} error ($\mathrm{m}$) ↓ \\
\midrule
\multirow{9}{*}[-0.45ex]{EL-SLAM~\cite{elslam}}
 & avg      & 4.681 &  0.271 \\ 
\addlinespace[0.3ex]
\cdashline{2-4}[0.7pt/2pt]
\addlinespace[0.6ex]
 & \texttt{room0}    & 2.419 & 0.289 \\
 & \texttt{room1}    & 6.520 & 0.257 \\
 & \texttt{room2}    & 2.745 & 0.475 \\
 & \texttt{office0}  & 5.901 & 0.120 \\
 & \texttt{office1}  & 5.102 & 0.126 \\
 & \texttt{office2}  & 3.261 & 0.288 \\
 & \texttt{office3}  & 8.254 & 0.301 \\
 & \texttt{office4}  & 3.241 & 0.315 \\
\midrule
\multirow{9}{*}[-0.45ex]{{LIMAP}~\cite{limap}}
 & avg      & 1.623 & 0.076 \\
\addlinespace[0.3ex]
\cdashline{2-4}[0.7pt/2pt]
\addlinespace[0.6ex]
 & \texttt{room0}    & \textbf{0.776} & 0.103 \\
 & \texttt{room1}    & 1.243 & 0.060 \\
 & \texttt{room2}    & 3.810 & 0.136 \\
 & \texttt{office0}  & 1.089 & 0.032 \\
 & \texttt{office1}  & 2.275 & 0.042 \\
 & \texttt{office2}  & 1.113 & 0.105 \\
 & \texttt{office3}  & 1.038 & 0.027 \\
 & \texttt{office4}  & 1.642 & 0.105 \\
\midrule
\multirow{9}{*}[-0.45ex]{{EMVS}~\cite{emvs}}
 & avg      & 2.673 & 0.158 \\
\addlinespace[0.3ex]
\cdashline{2-4}[0.7pt/2pt]
\addlinespace[0.6ex]
 & \texttt{room0}    & 3.031 & 0.220 \\
 & \texttt{room1}    & 4.002 & 0.188 \\
 & \texttt{room2}    & 3.258 & 0.208 \\
 & \texttt{office0}  & 1.901 & 0.081 \\
 & \texttt{office1}  & 1.556 & 0.040 \\
 & \texttt{office2}  & 1.385 & 0.210 \\
 & \texttt{office3}  & 3.615 & 0.136 \\
 & \texttt{office4}  & 2.636 & 0.183 \\
\midrule
\multirow{9}{*}[-0.45ex]{Ours}
 & avg      & \textbf{0.714} & \textbf{0.019} \\
\addlinespace[0.3ex]
\cdashline{2-4}[0.7pt/2pt]
\addlinespace[0.6ex]
 & \texttt{room0}    & 0.882 & \textbf{0.029} \\
 & \texttt{room1}    & \textbf{0.816} & \textbf{0.014} \\
 & \texttt{room2}    & \textbf{1.942} & \textbf{0.047} \\
 & \texttt{office0}  & \textbf{0.346} & \textbf{0.014} \\
 & \texttt{office1}  & \textbf{0.980} & \textbf{0.016} \\
 & \texttt{office2}  & \textbf{0.279} & \textbf{0.009} \\
 & \texttt{office3}  & \textbf{0.217} & \textbf{0.010} \\
 & \texttt{office4}  & \textbf{0.254} & \textbf{0.016} \\
\bottomrule
\end{tabular}}
\end{table}

%% file: sec/04-6_registration.tex
\subsection{Registration}

\label{subsec:registration}

\input{tables/registration_vector_transpose}

\subsubsection{Setup}
To demonstrate the usefulness of our reconstructed line maps and evaluate the performance in a downstream task, we perform cross-modal registration between our line maps and maps reconstructed from RGB-D inputs.
For the Replica~\cite{replica} dataset, we use Co-SLAM~\cite{co-slam}, a neural SLAM system that operates on RGB-D images, to obtain neural implicit representations of the scene.
We convert the implicit representations into meshes by using matching cubes~\cite{marching-cubes}, and densely sample points from the meshes to construct a reference point map.

From our 3D line reconstructions, we sample points along each line at $5mm$ intervals to build a point map.
We then apply point-to-point Iterative Closest Point (ICP)~\cite{icp} to register our sampled point map to the map generated by Co-SLAM.
For the EMVS, we directly use the reconstructed point map.
For EL-SLAM and LIMAP, we adopt the same point sampling procedure from line maps as in our method to ensure fair comparison.
For each scene, we perform registration with 10 randomly perturbed initial poses and report the average error.
We added zero-mean Gaussian noise with a standard deviation of 10° sequentially along the yaw, pitch, roll to perturb the rotation. We then applied additional zero-mean Gaussian translational perturbations with a standard deviation of 0.02m.
The translation error is calculated by $\left\| \hat{\mathbf{t}} - \mathbf{t}_{gt} \right\|_2 $ and the rotation error is evaluated by $\cos^{-1}\left( (\operatorname{tr}(\hat{\mathbf{R}}^\top \mathbf{R}_{gt}) - 1) / 2 \right)$, where $\hat{\mathbf{t}}$ and $\hat{\mathbf{R}}$ denote the estimated translation and rotation, and $\mathbf{t}_{gt}$ and $\mathbf{R}_{gt}$ represent the ground-truth translation and rotation, respectively.

We also evaluate registration performance on real-world scenes using VECtor~\cite{vector} dataset.
Since the VECtor dataset does not include RGB images aligned with depth, we do not employ RGB-D SLAM. Instead, we use the ground-truth depth and poses to directly build a TSDF volume via TSDF-Fusion~\cite{tsdf-fusion, 3dmatch}, which is the same map used for evaluation in~\Cref{subsec:recon}.
We then extract a surface mesh using marching cubes and sample a dense point map from it.
Following the same procedure as in Replica, we apply 10 random perturbations per scene and evaluate registration using the same metrics.

\subsubsection{Results}

Quantitative registration results on the Replica~\cite{replica} dataset are shown in~\Cref{tab:registration_replica_transpose}.
Here, \textit{t} error is measured in meters, and \textit{R} error is in degrees.
Except for the rotation error in \texttt{room0}, our method consistently achieves the lowest registration error across all metrics and scenes.
This highlights the effectiveness of our line maps as a compact yet sufficient representation for registering with dense point maps.
It also shows that our reconstructed line map can serve as an effective mid-level representation for the cross-modal registration task.

\Cref{tab:registration_vector_transpose} presents the results on the VECtor~\cite{vector} dataset.
On average, our method outperforms all baselines in both \textit{R} error and \textit{t} error.
Due to the lower resolution and challenging scenarios in the VECtor dataset, the errors are generally higher than those observed in Replica.

Qualitative results in~\Cref{fig:registration} illustrate registration on the \texttt{office4} scene from the Replica dataset.
Our method achieves the best alignment with the reference map reconstructed with RGB-D inputs, thanks to its low noise and high completeness.
In particular, successful registration can be observed at the ceiling boundaries and around the chair region.

%% file: tables/registration_vector_transpose.tex
\begin{table}[t]
\caption{Quantitative comparison of registration performance on the VECtor~\cite{vector} dataset. Bold indicates best.}
\label{tab:registration_vector_transpose}
\centering
\scriptsize
\resizebox{0.50\textwidth}{!}{%
\begin{tabular}{llcc}
\toprule
\textbf{Method} & \textbf{Scene} & \textit{R} error ($^\circ$) ↓ & \textit{t} error ($\mathrm{m}$) ↓ \\
\midrule
\multirow{3}{*}[-0.45ex]{{EL-SLAM}~\cite{elslam}}
 & avg  & 15.82 & 0.62 \\
\addlinespace[0.3ex]
\cdashline{2-4}[0.7pt/2pt]
\addlinespace[0.6ex]
 & \texttt{desk-fast} & 21.28 & 0.91 \\
 & \texttt{sofa-fast} & 26.09 & 0.88 \\
 & \texttt{desk-normal} & \textbf{6.42} & 0.28 \\
 & \texttt{sofa-normal} & 9.49  & 0.41 \\
\midrule
\multirow{3}{*}[-0.45ex]{{LIMAP}~\cite{limap}}
 & avg  & 9.42 & 0.29 \\
\addlinespace[0.3ex]
\cdashline{2-4}[0.7pt/2pt]
\addlinespace[0.6ex]
 & \texttt{desk-fast} & \textbf{8.43} & 0.26 \\
 & \texttt{sofa-fast} & 12.08 & \textbf{0.31} \\
 & \texttt{desk-normal} & 9.11  & 0.36 \\
 & \texttt{sofa-normal} & 8.05  & 0.22 \\
\midrule
\multirow{3}{*}[-0.45ex]{{EMVS}~\cite{emvs}}
 & avg  & 16.34 & 0.56 \\
\addlinespace[0.3ex]
\cdashline{2-4}[0.7pt/2pt]
\addlinespace[0.6ex]
 & \texttt{desk-fast} & 17.92 & 0.73 \\
 & \texttt{sofa-fast} & 28.74 & 0.89 \\
 & \texttt{desk-normal} & 12.12 & 0.31 \\
 & \texttt{sofa-normal} & \textbf{6.60} & 0.32 \\
\midrule
\multirow{3}{*}[-0.45ex]{Ours}
 & avg  & \textbf{8.86}  & \textbf{0.21} \\ 
\addlinespace[0.3ex]
\cdashline{2-4}[0.7pt/2pt]
\addlinespace[0.6ex]
 & \texttt{desk-fast} & 10.61 & \textbf{0.25} \\
 & \texttt{sofa-fast} & \textbf{9.80} & 0.33 \\
 & \texttt{desk-normal} & 6.75  & \textbf{0.11} \\
 & \texttt{sofa-normal} & 8.30  & \textbf{0.16} \\
\bottomrule
\end{tabular}}
\end{table}

%% file: sec/04-7_localization.tex
\subsection{Panoramic localization}
\label{subsec:localization}

\subsubsection{Setup}

In addition to registration, we conduct cross-modal panoramic localization to further evaluate how effectively our line maps can support downstream applications.
We adopt FGPL~\cite{fgpl} for this task, which estimates the camera pose of a query RGB panoramic image on a pre-built 3D line map.
FGPL~\cite{fgpl} extracts 2D line segments from the panoramic image and performs coarse pose search followed by pose refinement, relying solely on geometric line features without any visual descriptors.

We compare our method with line-based baselines: EL-SLAM~\cite{elslam} and LIMAP~\cite{limap}.
Following FGPL, we evaluate localization performance using median rotation error (\textit{R} error) and translation error (\textit{t} error), along with success rates.
The success rates are defined as the percentage of images whose pose errors fall within thresholds of 0.1m / $5^\circ$ and 0.3m / $15^\circ$, respectively.

We use the \texttt{room0} and \texttt{office4} scene from the Replica~\cite{replica} dataset and generate a total of 100 query panoramic images for each scene.
Each image is synthesized by rendering six cube faces with Habitat simulator~\cite{savva2019habitat} and converting them into an equirectangular panoramic image.

\input{figures/localization}

\subsubsection{Results}
Using 3D lines as mid-level representations, our method successfully performs panoramic localization between event-based maps and RGB panoramic images.
From the quantitative results in~\Cref{tab:localization}, our method achieves a success rate of 81.5\% under the 0.3m / $15^\circ$ threshold, significantly outperforming all baselines.

Compared to EL-SLAM~\cite{elslam} and LIMAP~\cite{limap}, our approach yields the lowest median rotation and translation errors, as well as the highest success rates across all thresholds.
These results demonstrates the robustness and accuracy of our line-based mapping for panoramic localization.
Since we report median errors, EL-SLAM, which suffers from a low success rate, results in a large reported error.

\Cref{fig:localization} presents qualitative results of panoramic localization on the \texttt{room0} scene.
The background shows the RGB panoramic query image, while red lines indicate the projections of the 3D line map at the estimated camera pose.
As illustrated, our line map contains less noise and accurately reconstructs structures such as windows, doors, and picture frames, which are useful for localization.
In the zoomed-in regions marked by blue boxes, our map is cleaner and more complete, resulting in more accurate localization.

\input{tables/localization}

%% file: figures/localization.tex
\begin{figure}[t]
    \centering
    \renewcommand{\arraystretch}{1.1}
    \begin{tabular}{c}

        \includegraphics[width=0.9\linewidth]{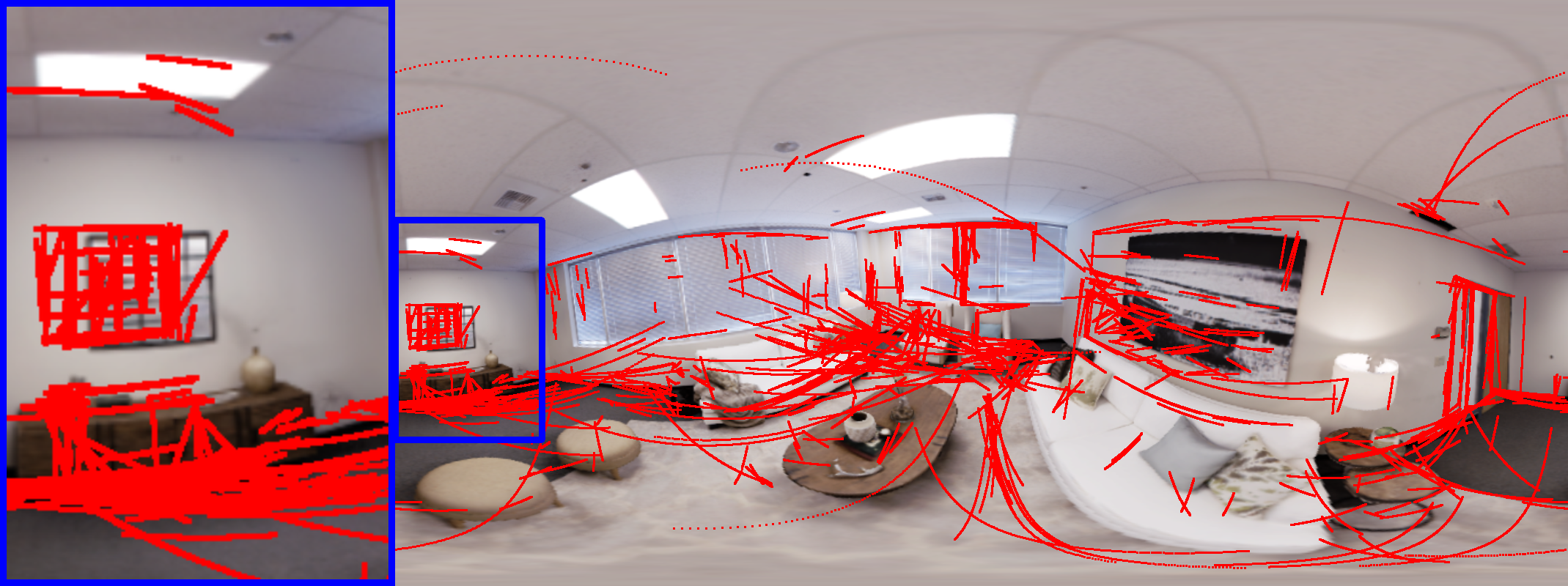} \\
        \makebox[\linewidth][c]{EL-SLAM~\cite{elslam}} \\[1mm]

        \includegraphics[width=0.9\linewidth]{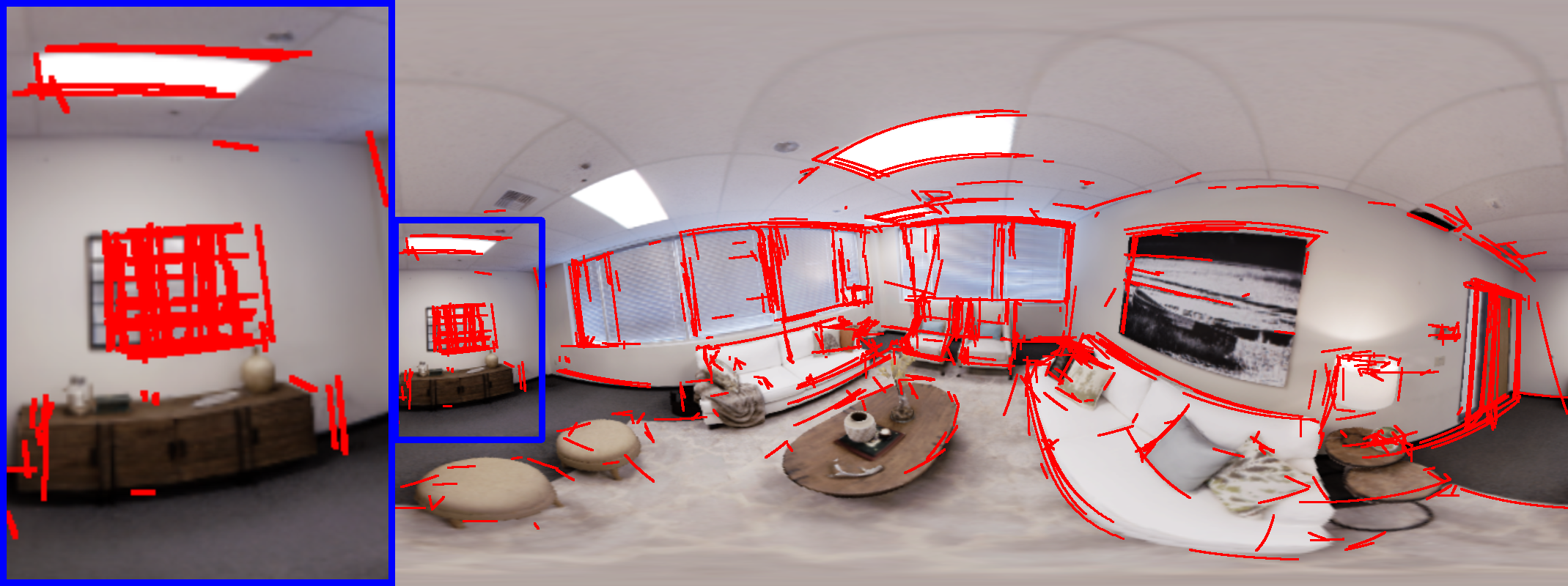} \\
        \makebox[\linewidth][c]{LIMAP~\cite{limap}} \\[1mm]

        \includegraphics[width=0.9\linewidth]{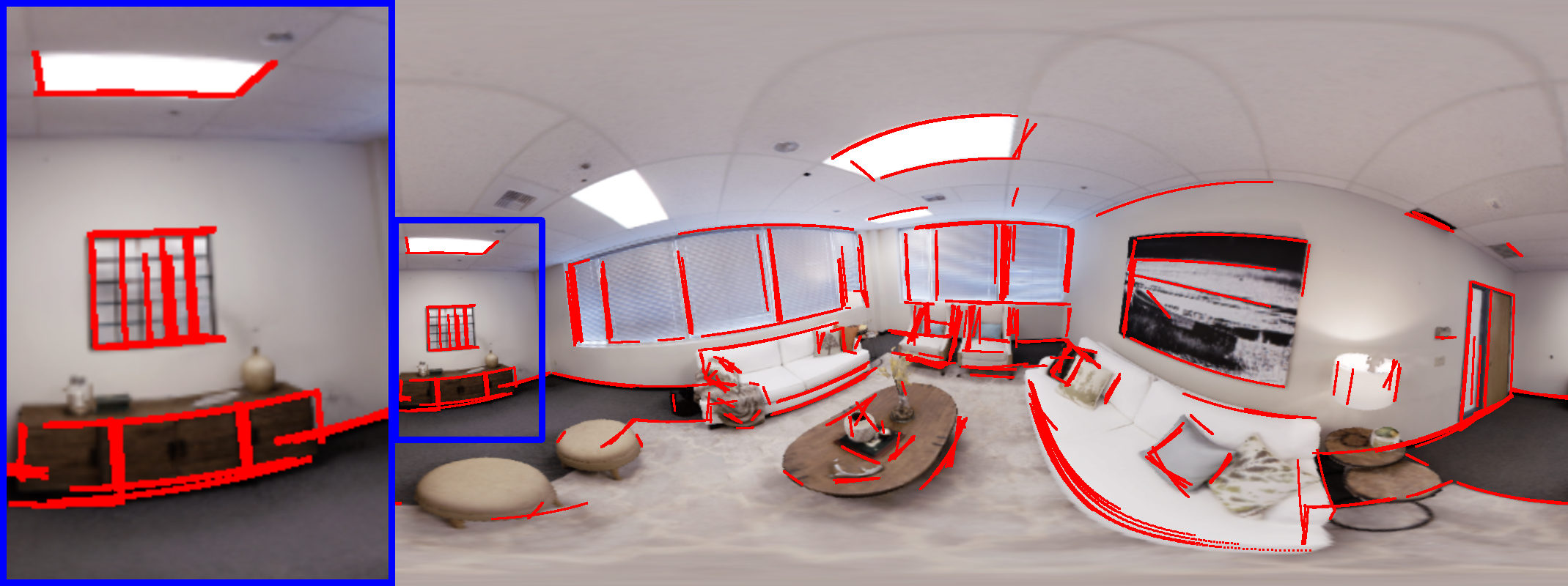} \\
        \makebox[\linewidth][c]{Ours} \\

    \end{tabular}
    \caption{Qualitative comparison of panoramic localization performance on \texttt{room0} of the Replica~\cite{replica} dataset. The background shows a query RGB panoramic image, and the red lines represent the projected 3D line map at the estimated camera pose. The blue boxes indicate the zoom-in regions.}
    \label{fig:localization}
\end{figure}

%% file: tables/localization.tex
\begin{table}[t]
\caption{Quantitative comparison of panoramic localization performance on the Replica~\cite{replica} dataset. Bold indicates best.}
\label{tab:localization}
\centering
\large
\resizebox{0.5\textwidth}{!}{
\begin{tabular}{lcccc}
\toprule

\textbf{Method} & \makecell{\textit{R} error ($^\circ$) ↓} & \makecell{\textit{t} error ($\mathrm{m}$) ↓} & \makecell{Success\\ (0.1$\mathrm{m}$/5$^\circ$) ↑} & \makecell{Success\\ (0.3$\mathrm{m}$/15$^\circ$) ↑} \\
\midrule
{EL-SLAM}~\cite{elslam} & 104.30 & 2.37 & 10.5 & 30.5 \\
{LIMAP}~\cite{limap}  & 5.81   & 0.08 & 37.0 & 61.0 \\
Ours  & \textbf{5.18} & \textbf{0.04} & \textbf{45.5} & \textbf{81.5} \\
\bottomrule
\end{tabular}
}
\end{table}

%% file: sec/04-5_pose.tex
\subsection{Pose Refinement}
\label{subsec:pose}

\input{tables/pose}
\input{figures/pose}

\subsubsection{Setup}
Our optimization module in~\Cref{subsubsec:opt} jointly refines both the 3D line map and the noisy camera poses.
We optimize the camera poses by minimizing the cost function defined in~\Cref{eq:opt_cost}.
The rotational component of each pose is represented in the $\mathfrak{so}(3)$ Lie algebra using the logarithmic mapping.
Since we use a monocular camera setup, the first camera pose is fixed, and only the remaining poses are optimized.
The optimization is performed a batch manner, where the cost terms for all lines in all frames are calculated and all camera poses are updated simultaneously.

We demonstrate the effectiveness of pose refinement on both synthetic and real-world data.
To isolate the effect of pose refinement from 3D line reconstruction accuracy, we first use a synthetic dataset.
We simulate event data by capturing a simple cube-shaped synthetic object divided into $2 \times 2$ surface grids and convert it to events using VID2E~\cite{vid2e}.
The synthetic dataset is unit-less, with a cube of side length 10 observed from a distance of 25 units.
In contrast, real-world results follow the metric scale (in meters).
For clarity, we report errors in 1/100 scale in~\Cref{tab:pose}.
Using this dataset, we run our pipeline up to the line triangulation step in~\Cref{subsubsec:triang}, using ground-truth poses to generate 3D lines.
We then add Gaussian noise to the camera positions and Euler angles and refine the poses using the optimization with~\Cref{eq:opt_cost}.
For real-world evaluation, we use the TUM-VIE~\cite{tum-vie} dataset.
In particular, we analyze the performance on the \texttt{desk} scene, where the camera poses provided by DEVO show non-negligible errors.
We evaluate how much these errors are reduced through our optimization, using the Absolute Trajectory Error (ATE) metric~\cite{sturm2012benchmark}.

To further investigate the contribution of each component, we also conduct an ablation study on the pose refinement.
We compare two cases: (1) using only the 2D line-3D line Grassmann cost, and (2) using~\Cref{eq:opt_cost} with both the 2D line-3D line and event-3D line Grassmann cost.
We also compare our cost function, which is defined in the 3D space, with the reprojection error used in LIMAP.

\subsubsection{Results}
Our method successfully refines the camera poses and achieves better performance than the projection-based cost function of LIMAP~\cite{limap}.
As shown in~\Cref{tab:pose}, even when using only our 2D line-based cost function, the refined poses outperform those obtained using LIMAP.
Unlike projection-based costs, our Grassmann distance-based cost function is defined in 3D space, preserving depth-related information.
When the event-based cost function is also included, we observe a significant performance improvement.
The event-based term is less sensitive to errors in 2D line detection and leverages the advantages of direct methods by associating events with lines and selectively using inlier events.

Qualitative results in~\Cref{fig:pose} visualize the camera trajectories on the synthetic dataset.
Ground-truth trajectories are shown as black dashed lines, while the noisy and optimized poses are shown in blue lines.
When using LIMAP's cost function, the poses are slightly refined but still noisy.
In contrast, our cost function results in smooth and accurate trajectories.

%% file: tables/pose.tex
\begin{table}[t]
\caption{Pose refinement performance analysis. We compare LIMAP's projection-based line cost and our Grassmann-based line and event cost functions.}
\label{tab:pose}
\centering
\resizebox{\linewidth}{!}{
\begin{tabular}{ccc|cc}
\toprule
\makecell{LIMAP~\cite{limap}} & \makecell{Ours\\Line Grass.} & \makecell{Ours\\Event Grass.} & \makecell{ATE ↓\\(Synthetic)} & \makecell{ATE ↓\\(Real-world)} \\
\midrule
-      & -      & -      & 33.25 & 0.89 \\
\cmark & -      & -      & 32.05 & 0.76 \\
-      & \cmark & -      & 27.95 & 0.73 \\
-      & \cmark & \cmark & \textbf{6.92} & \textbf{0.67} \\
\bottomrule
\end{tabular}
}
\end{table}

%% file: figures/pose.tex
\begin{figure}[t]
    \centering
    \renewcommand{\arraystretch}{1.1}
    \begin{tabular}{@{}c@{\,}c@{\,}c@{}}

        \includegraphics[width=0.33\linewidth]{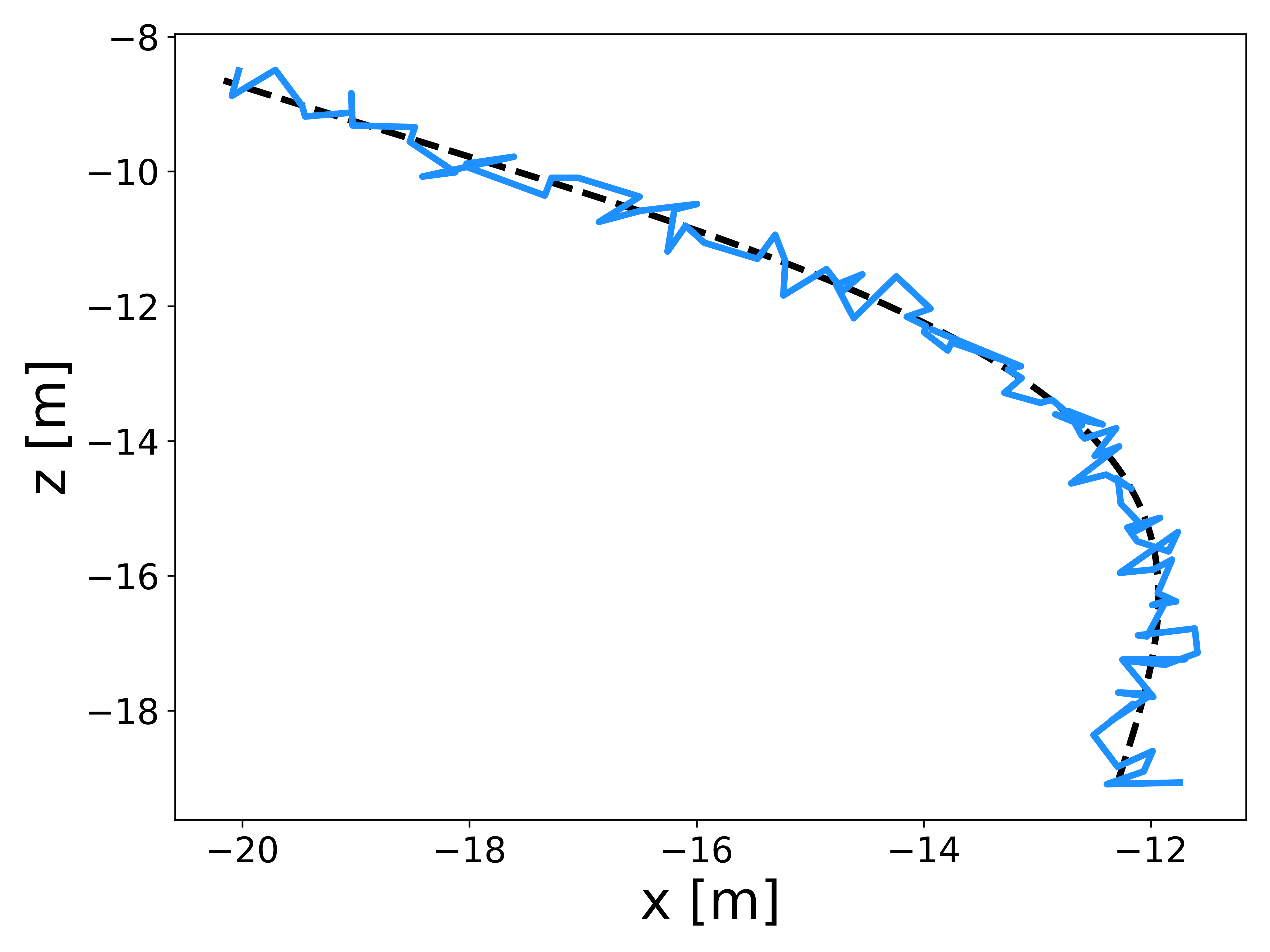} &
        \includegraphics[width=0.33\linewidth]{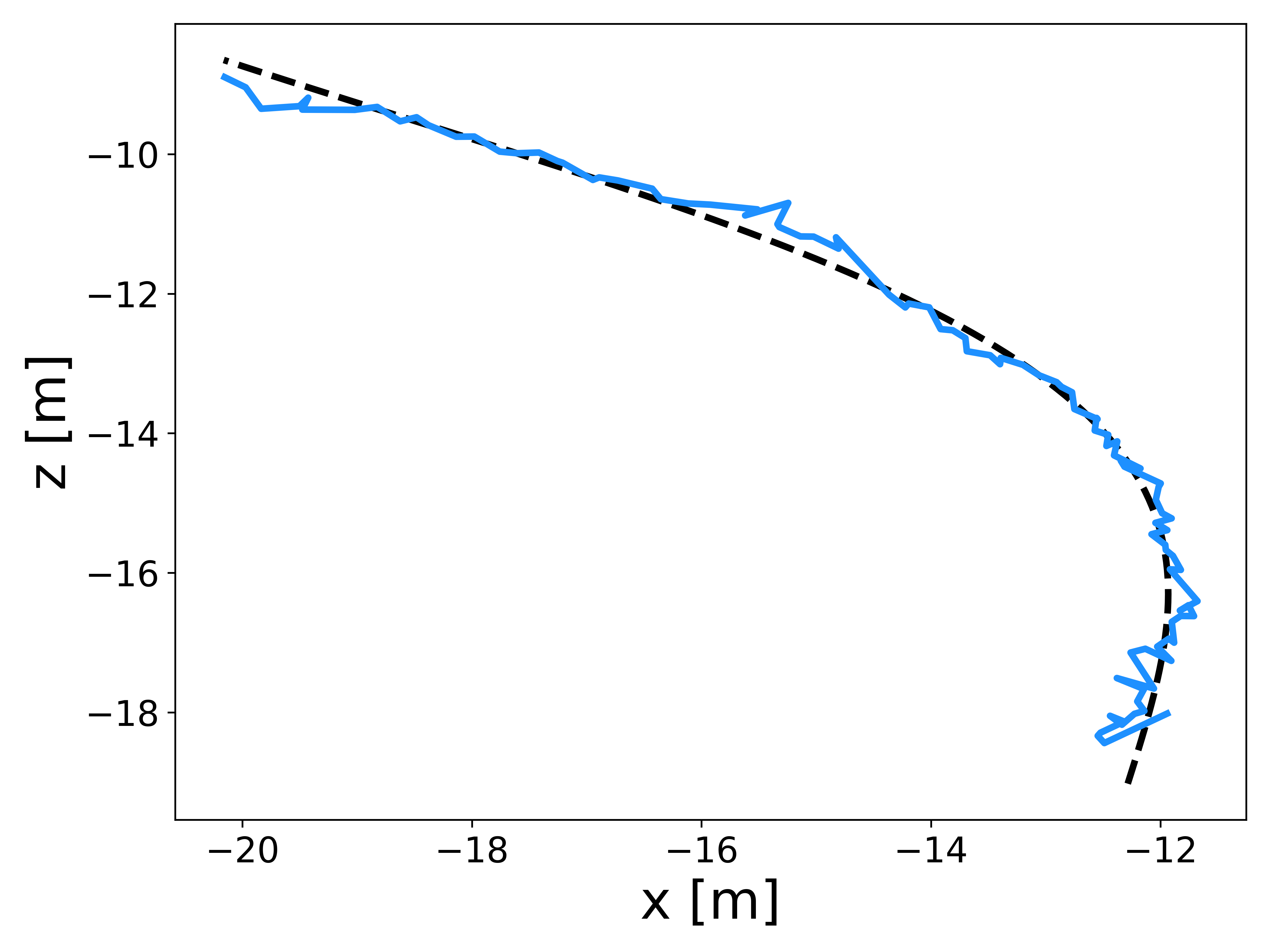} &
        \includegraphics[width=0.33\linewidth]{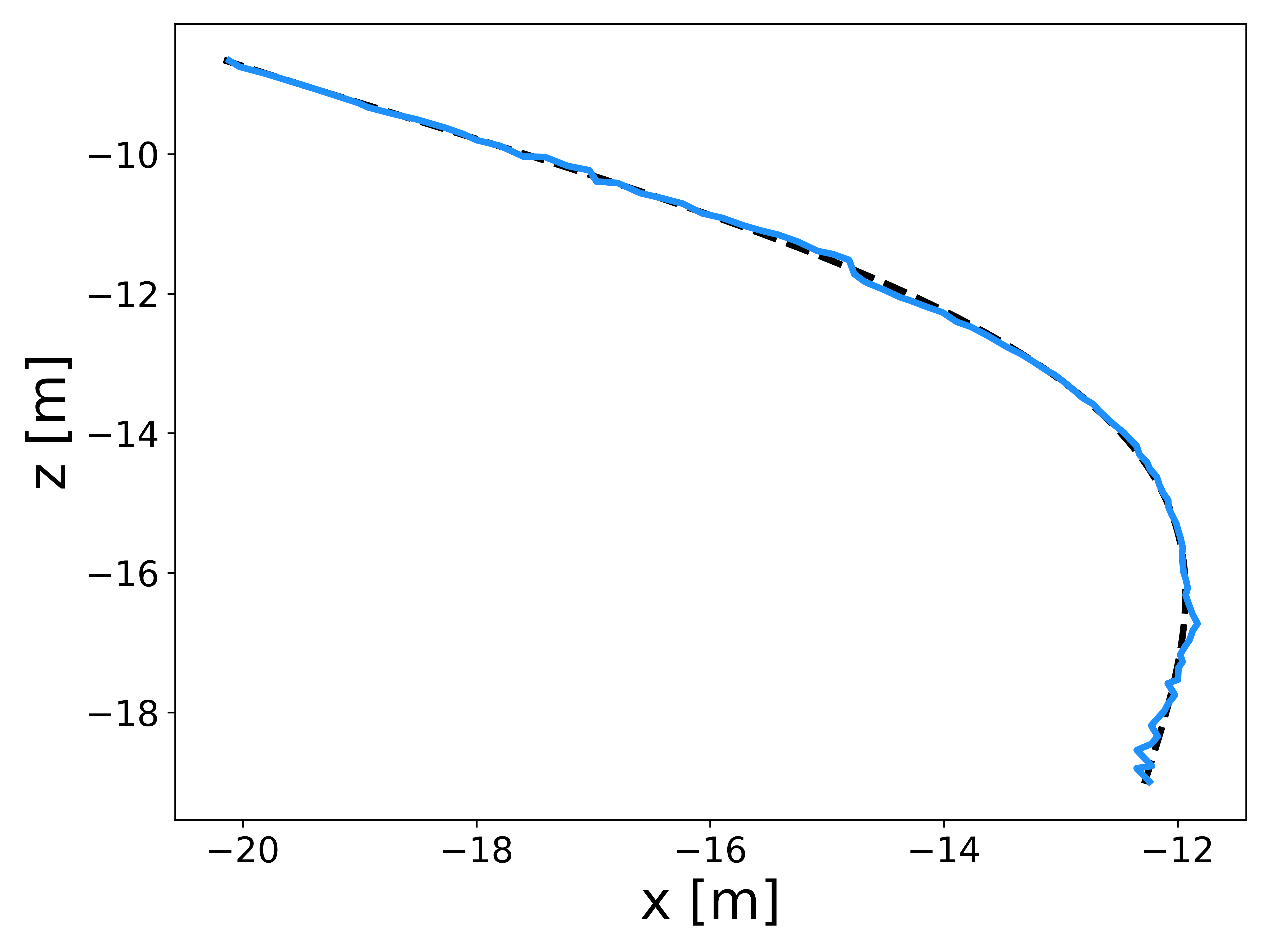} \\

        \makebox[0.33\linewidth][c]{Noisy Poses} &
        \makebox[0.33\linewidth][c]{LIMAP~\cite{limap}} &
        \makebox[0.33\linewidth][c]{Ours} \\

    \end{tabular}
    \caption{Qualitative comparison of pose refinement performance. Black dashed lines indicates the ground-truth camera trajectory, and the noisy and optimized poses are shown in blur lines.}
    \label{fig:pose}
\end{figure}

%% file: sec/04-8_challenging.tex
\subsection{Robustness Assessment in Challenging Conditions}
\label{subsec:challenging}

\input{figures/challenging}

\subsubsection{Setup}
We evaluate the robustness of our event-based line mapping method under challenging visual conditions where conventional frame-based cameras often fail.
Specifically, we compare our method with RGB-based line mapping, LIMAP~\cite{limap}, under high-speed motion and extreme lighting scenarios.
To ensure a realistic yet controlled simulation, we use Blender software to generate data with identical scenes and camera trajectories.
Following the motion blur and auto-exposure simulation method used in $I^2$-SLAM~\cite{i2slam}, we simulate image corruptions in the \texttt{italian-flat} scene.
Three types of RGB image sequences are rendered: sharp RGB images, RGB images with motion blur, and RGB images with both motion blur and underexposure.
For each of these sequences, we reconstruct 3D line maps from RGB images using LIMAP.
To simulate event data, we render sharp RGB frames at five times higher temporal resolution and generate events using VID2E~\cite{vid2e}.
While event cameras are generally robust under high-speed motion and challenging lighting conditions, their measurements can degrade in dark environments due to increased shot noise~\cite{guo2022low, hu2021v2e}.
To account for this effect, we additionally synthesize noisy events by following the noise simulation method in~\cite{hu2021v2e}.
Specifically, we inject noise events corresponding to 15\% of the clean event count.
While our synthetic data enables controlled comparisons across capture conditions, it may introduce a domain gap relative to real sensor artifacts.

\subsubsection{Results}
Qualitative results in \Cref{fig:challenging} show that RGB-based line mapping severely degrades under motion blur and underexposure.
Line structures corresponding to windows and sofas are well reconstructed when using sharp RGB images, but largely disappear when motion blur and underexposure are introduced.
In contrast, our event-based line mapping successfully reconstructs most of the window boundaries and a significant portion of the sofa structure.
Moreover, our method reconstructs ceiling boundaries that are not recovered even when using sharp RGB images.
Under these challenging visual conditions, our event-based line mapping method demonstrates better robustness compared to frame-based approaches.
We further observe that the performance of our method yields similar line mapping results when using noisy events that mimic shot noise in dark environments.
However, due to the inherent noise in event data, the reconstructions tend to be less detailed compared to those generated from sharp RGB images.

%% file: figures/challenging.tex
\begin{figure*}[t]
    \centering
    \renewcommand{\arraystretch}{1.1}
    \begin{tabular}{@{}c@{\,}c@{\,}c@{\,}c@{\,}c@{\,}c@{}}

        \includegraphics[width=0.15\textwidth]{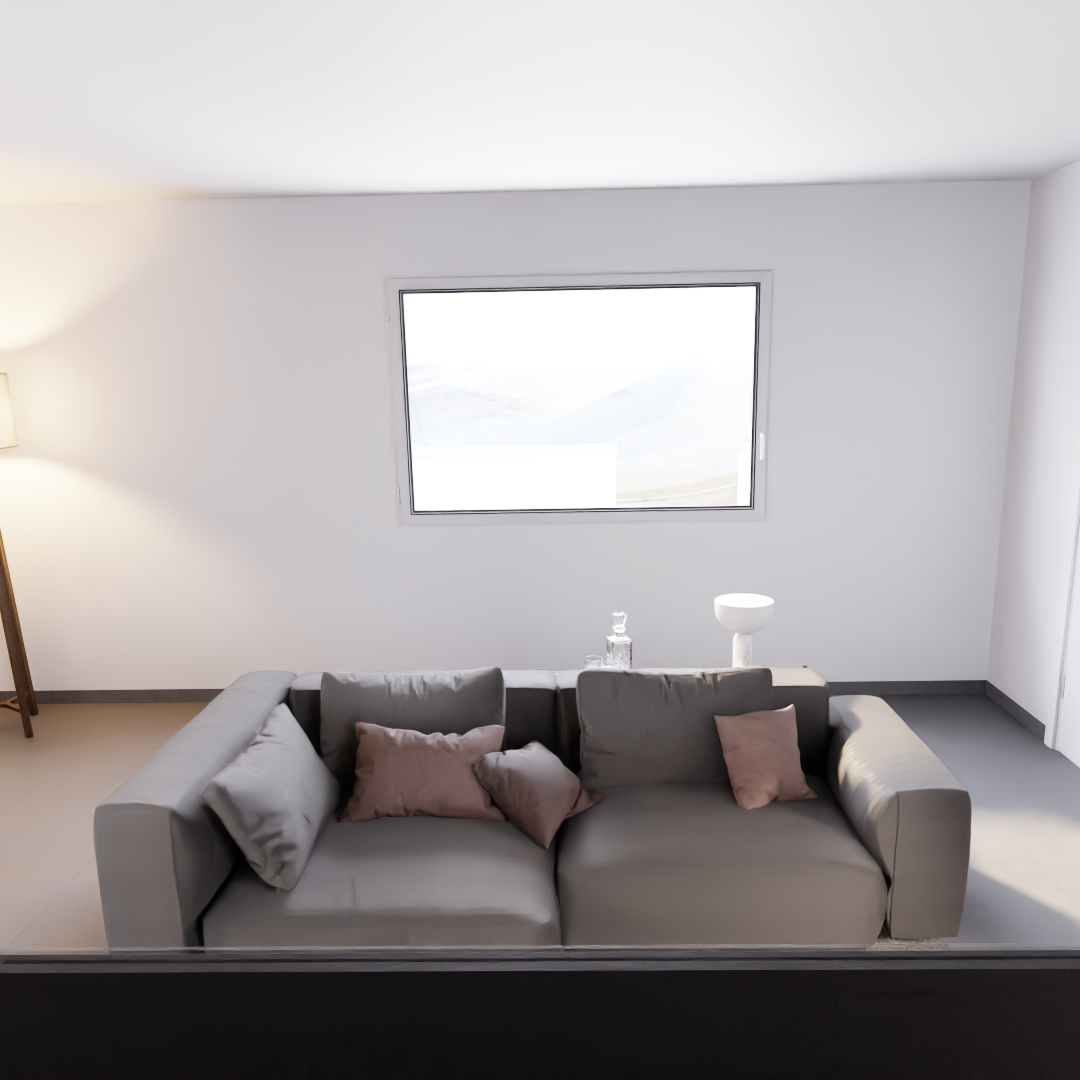} &
        \includegraphics[width=0.15\textwidth]{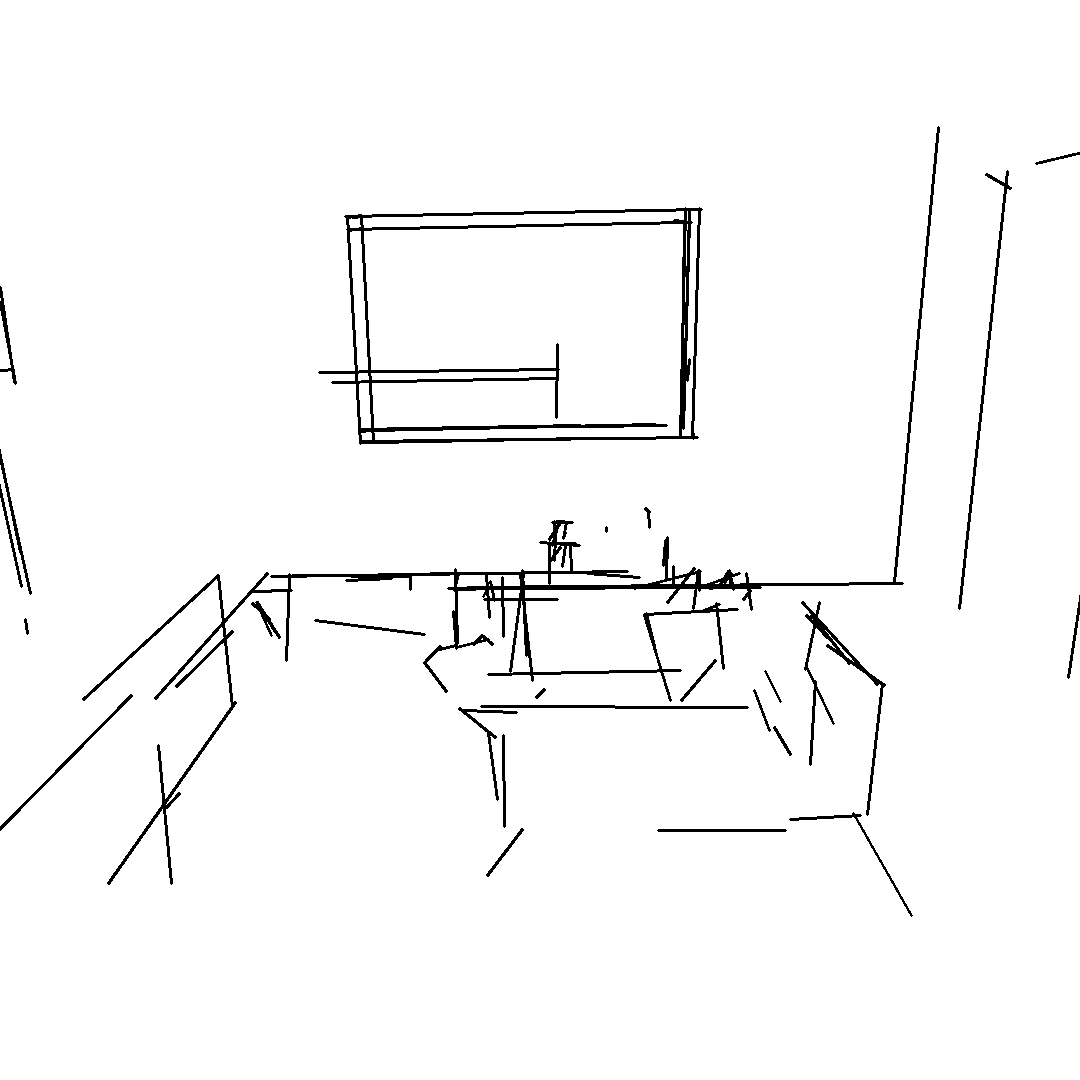} &
        \includegraphics[width=0.15\textwidth]{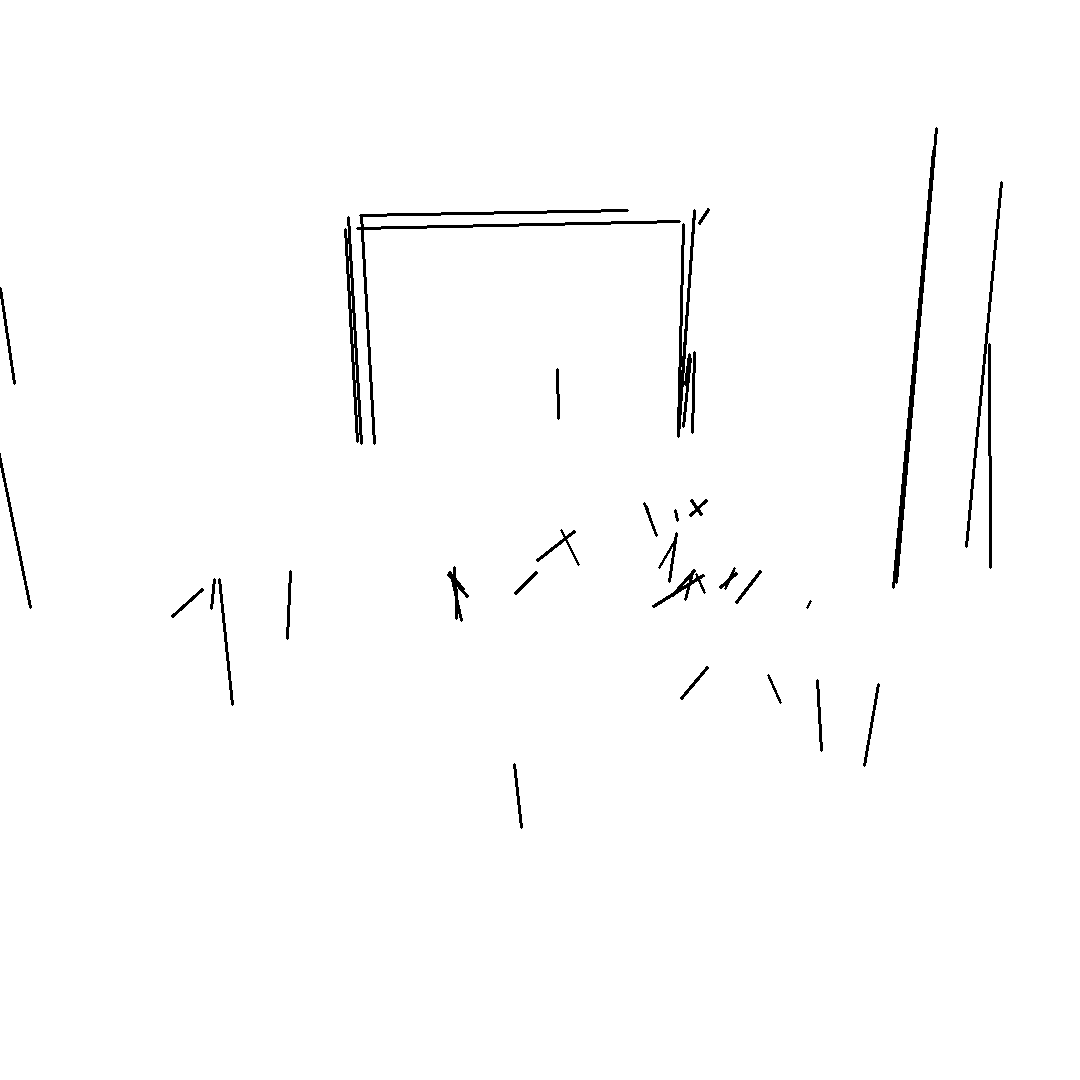} &
        \includegraphics[width=0.15\textwidth]{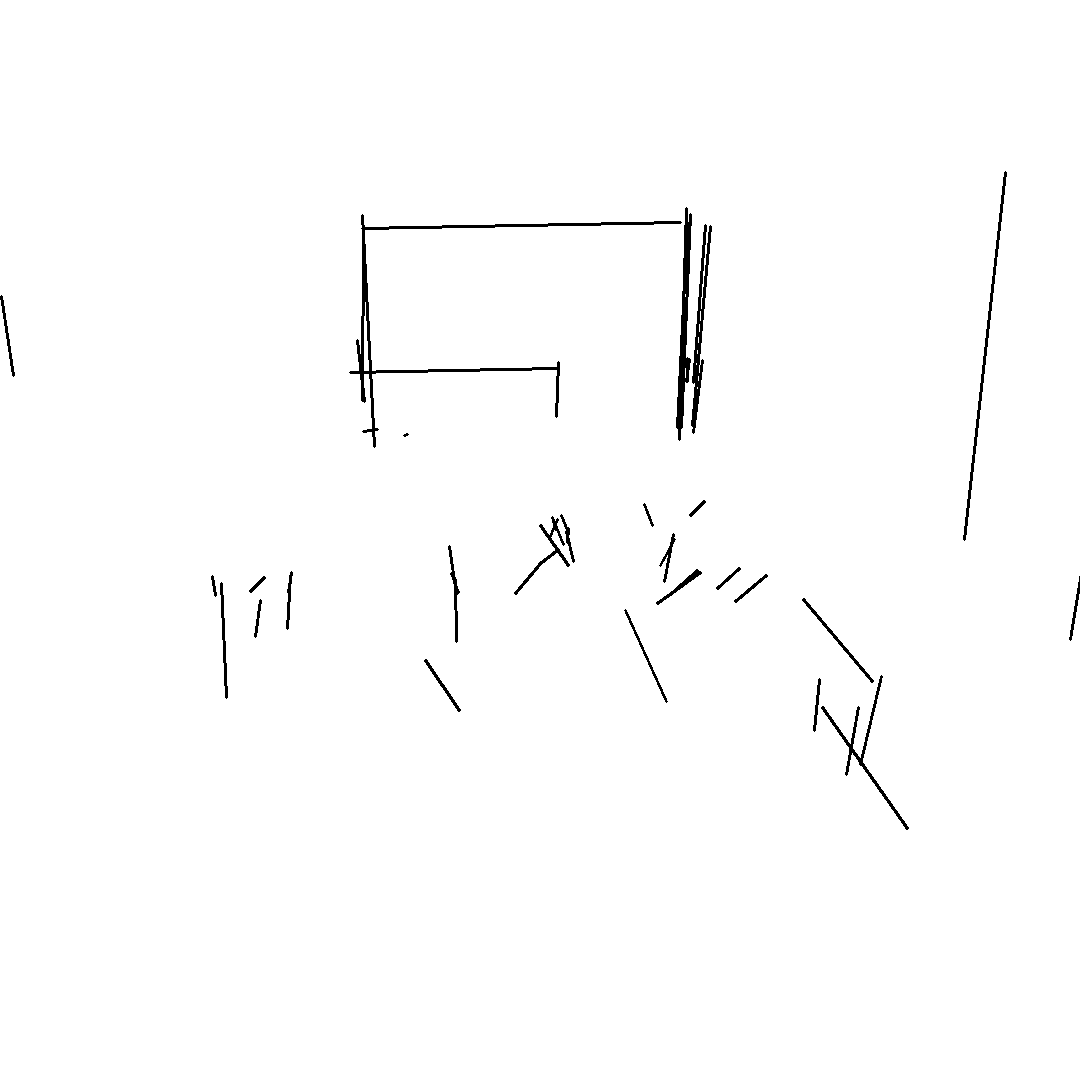} &
        \includegraphics[width=0.15\textwidth]{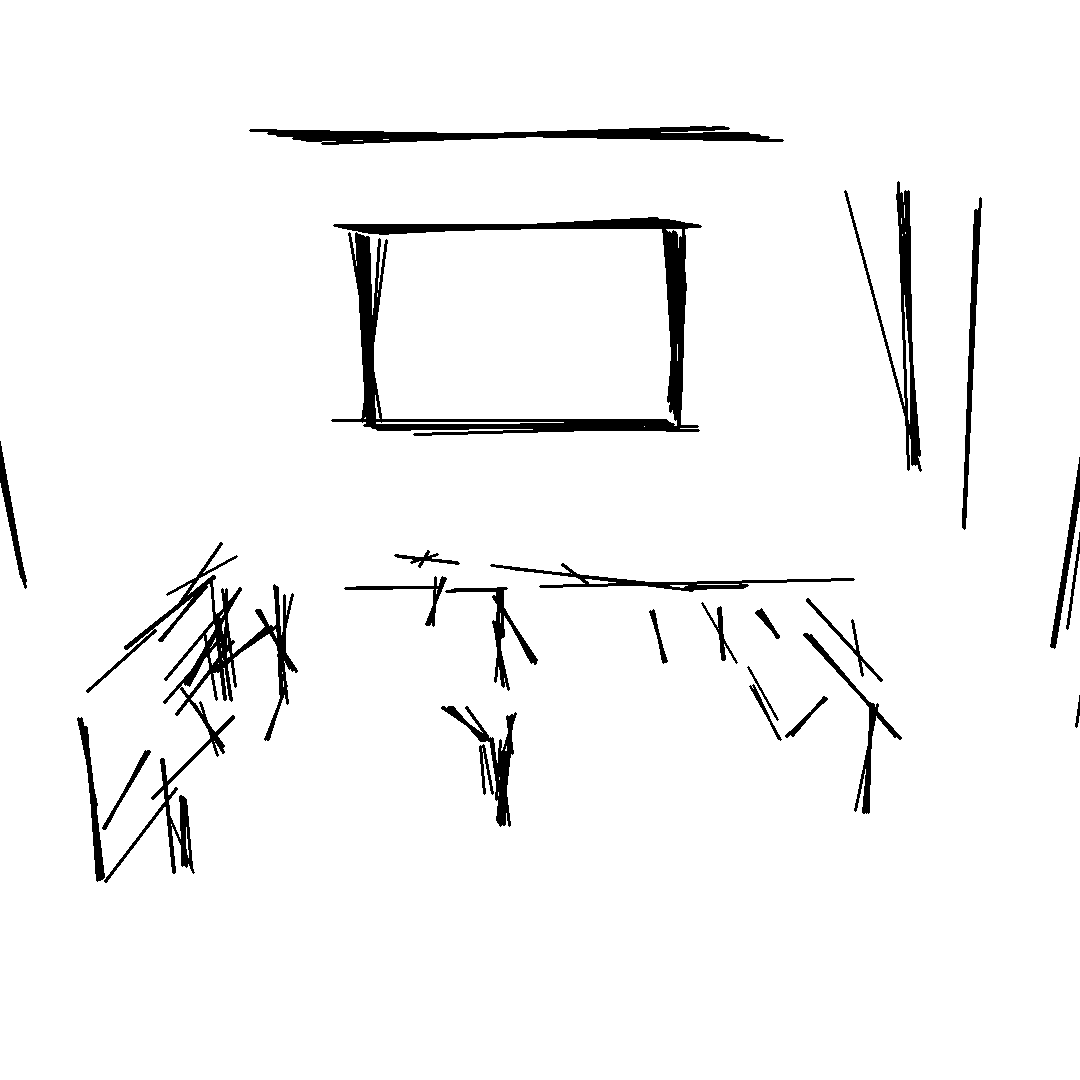} &
        \includegraphics[width=0.15\textwidth]{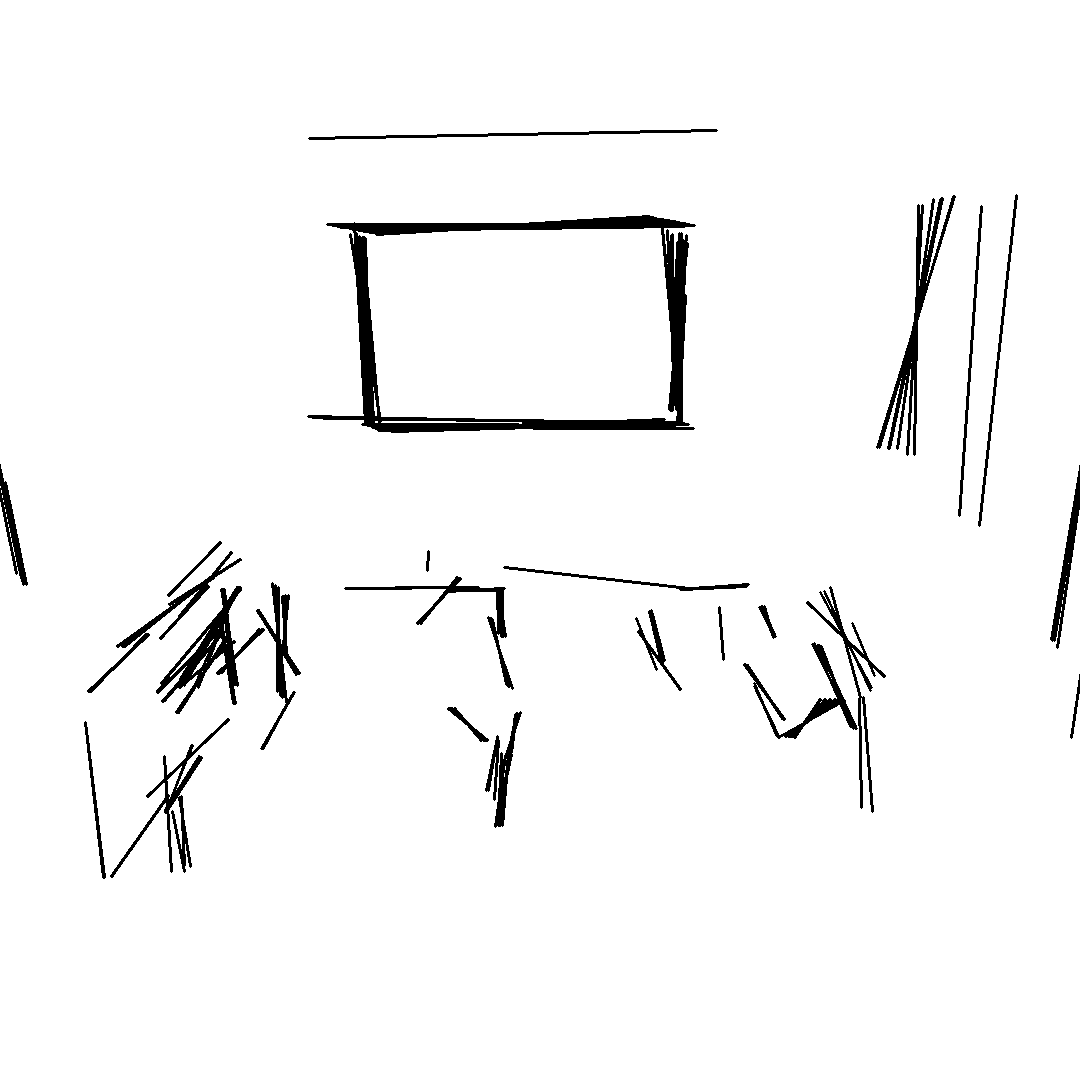} \\

        \parbox[c]{0.16\textwidth}{\centering Scene} &
        \parbox[c]{0.16\textwidth}{\centering Sharp\\RGB} &
        \parbox[c]{0.16\textwidth}{\centering Blurry\\RGB} &
        \parbox[c]{0.16\textwidth}{\centering Blurry, Underexp.\\RGB} &
        \parbox[c]{0.16\textwidth}{\centering Ours\\Events} &
        \parbox[c]{0.16\textwidth}{\centering Ours\\Noisy Events} \\
        
    \end{tabular}
    \caption{Comparison between frame-based  line mapping with~\cite{limap} and our event-based line mapping under high speed motion and underexposure. Sharp image in favorable condition, blurry image in high speed motion, blurry and underexposed image in high speed motion are used for comparison. Noisy events are synthesized to simulate low-light shot noise by injecting noise events into the clean event stream.}
    \label{fig:challenging}
\end{figure*}

%% file: sec/05_analysis.tex
\section{Component-wise Analysis}
\label{sec:analysis}
Our method introduces several components that collectively enable correspondence-based line mapping for event camera.
In this section, we conduct a component-wise analysis of our reconstruction pipeline to examine the contribution and behavior of each stage.
We first perform an ablation study to quantify the effect of each proposed component on the overall reconstruction performance in~\Cref{subsec:analysis_ablation}.
We then evaluate how the detection-guided space-time plane fitting serves as a robust solution for inlier event association and line refinement in~\Cref{subsec:analysis_plane}.
Furthermore, in~\Cref{subsec:analysis_line}, we analyze how our multi-window, multi-representation line detection enables reliable line detection from event data.
Finally, we report the runtime of each component to assess the computational efficiency in~\Cref{subsec:analysis_runtime}.

\input{sec/05-1_ablation}
\input{sec/05-2_plane_fitting}
\input{sec/05-3_line_detection}
\input{sec/05-4_runtime}

%% file: sec/05-1_ablation.tex
\subsection{Ablation Studies}
\label{subsec:analysis_ablation}

\input{tables/ablation}

\subsubsection{Setup}
To understand the individual contributions of each component in our pipeline, we perform ablation studies on five key elements.
We analyze the following five components: 1) Multi-window, multi-representation (MWMR) line detection, 2) Space-time plane fitting for 2D line refinement and event-line association, 3) Optimization with projection-based cost function used in LIMAP~\cite{limap}, 4) Optimization with our 2D line-based Grassmann distance, 5) Optimization with our event-based Grassmann distance
The experiments are conducted on the \texttt{office4} scene from the Replica dataset.
We evaluate performance using the metrics \textit{Accuracy}, \textit{Completion}, and \textit{$IoU_{20}$}, as defined in~\Cref{subsec:recon}.

\subsubsection{Results}
As shown in~\Cref{tab:ablation}, the \textit{IoU} metric steadily improves as each component is incrementally added, demonstrating the effectiveness of each component.
In particular, we observe a significant performance gain when applying space-time plane fitting and when incorporating the event-based Grassmann distance, highlighting the benefits of components specifically designed for the event data modality.
Although applying MWMR line detection alone leads to a slight drop in \textit{IoU}, it substantially improves \textit{Completion}, indicating a successful reduction in false negatives.
Refining these noisy yet abundant 2D lines with space-time plane fitting results in consistent improvements across all metrics.
Applying plane fitting without subsequent optimization yields the highest \textit{Completion} but suffers from low \textit{Accuracy} and \textit{IoU}, as many noisy 3D lines still remain prior to optimization.
When replacing our cost with the projection-based cost function used in LIMAP~\cite{limap}, we observe some improvement in \textit{IoU}.
However, using our line-based Grassmann distance further improves performance, and the best results are achieved when both our line-based and event-based Grassmann cost terms are jointly used in the optimization.

%% file: tables/ablation.tex
\begin{table*}[t]
\caption{Ablation study. MWMR refers to the multi-window, multi-representation line detection module. Plane fitting denotes the 2D line refinement step. LIMAP~\cite{limap}
's projection-based line cost and our projection-free line and event cost defined on the Grassmann manifold are also analyzed. When MWMR is not used, we report the average performance over all combinations.}
\label{tab:ablation}
\centering
\resizebox{0.88\textwidth}{!}{
\begin{tabular}{ccccc|ccc}
\toprule
\textbf{MWMR} & \makecell{\textbf{Plane}\\\textbf{Fitting}} & \makecell{\textbf{Opt.}\\\textbf{Line LIMAP~\cite{limap}}} & \makecell{\textbf{Opt.}\\\textbf{Line Grass.}} & \makecell{\textbf{Opt.}\\\textbf{Event Grass.}} 
& \textbf{Acc. ↓} & \textbf{Comp. ↓} & \textbf{IoU 20 ↑} \\
\midrule
-      & -      & -      & -      & -      & 65.98 & 104.93 & 0.068 \\
\cmark & -      & -      & -      & -      & 68.00 & 51.17 & 0.053 \\
\cmark & \cmark & -      & -      & -      & 47.94 & \textbf{44.09} & 0.089 \\
\cmark & \cmark & \cmark & -      & -      & 46.57 & 58.36 & 0.092 \\
\cmark & \cmark & -      & \cmark & -      & 39.81 & 58.82 & 0.097 \\
\cmark & \cmark & -      & \cmark & \cmark & \textbf{28.63} & 55.20 & \textbf{0.120} \\
\bottomrule
\end{tabular}
}
\end{table*}

%% file: sec/05-2_plane_fitting.tex
\subsection{Detection-guided Space-time Plane Fitting}
\label{subsec:analysis_plane}

\input{tables/analysis_plane}

The key feature of our space-time plane fitting lies in using line detection to guide candidate event clustering.
A key requirement for successful plane fitting is to provide reliable candidate event clusters for each line.
Previous works~\cite{everding2018low, liu2024line, dietsche2021powerline}, typically rely on spatio-temporal proximity, such as point cloud segmentation or region growing in space-time volume, which yields suboptimal results on sparse and noisy event data.
In contrast, our approach leverages mature frame-based line detectors through MWMR line detection, enabling accurate candidate event clustering even in complex scenes.
We then refine the detected 2D lines using spatio-temporal plane fitting.
Our method achieves robust line correspondence search by synergistically combining line detection and spatio-temporal plane fitting: line detection for finding 2D lines, and plane fitting for locating them accurately in space–time.

We conduct a quantitative comparison between our line-detection-guided plane fitting and a direct plane fitting variant.
The variant replaces~\Cref{eq:neighbor} with a DBSCAN~\cite{dbscan}-based clustering method for candidate event selection, while all other modules remain identical.
Because this baseline lacks the refined candidate selection step, we use ten times more RANSAC iterations to achieve stable performance.
For 2D endpoint trimming, we follow the same procedure as in~\cite{everding2018low}.
In this analysis, our method use the multi-window, multi-representation line detection model for candidate generation.

\begin{figure}[t]
    \centering
    \includegraphics[width=0.95\linewidth]{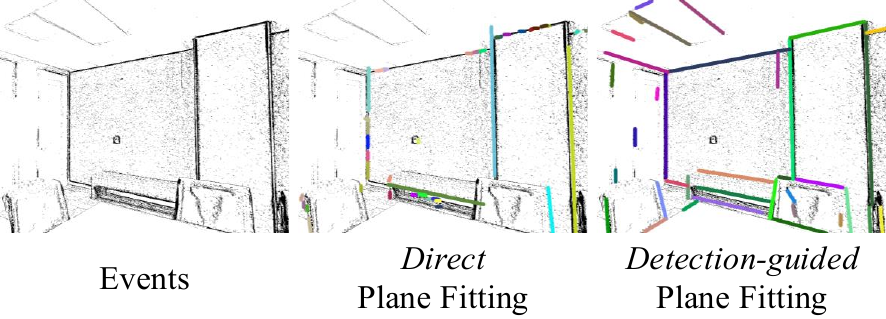}
    \caption{Qualitative analysis on plane fitting strategies. The fitted planes are sliced to visualize the resulting line segments.}
    \label{fig:analysis_plane}
\end{figure}

For quantitative evaluation, both approaches are tested on a line detection benchmark.
We obtain 2D lines by slicing the fitted planes with~\Cref{eq:slicing}.
Following the evaluation protocols in~\cite{huang2018learning, zhou2019end}, we measure the F-score against a ground-truth line heatmap and additionally report precision and recall for a more comprehensive analysis.
The ground-truth heatmap is generated by projecting the 3D edge map (used in~\Cref{eq:metric_completion}) onto the image plane.
All frames from the \texttt{office4} scene of the synthetic dataset are used for evaluation.

As shown in the quantitative results in~\Cref{tab:analysis_plane}, our detection-guided plane fitting achieves a higher F-score than the direct plane fitting baseline.
The direct variant suffers from low recall due to its instability in forming consistent event clusters.
Qualitative results in~\Cref{fig:analysis_plane} further demonstrate that the direct plane fitting approach often fails: missing lines in sparse regions (e.g., ceiling lights) or fragmenting one line into multiple clusters.
In contrast, our detection-guided strategy, which integrates frame-based line detection with multi-window and multi-representation accumulation, consistently segments events into accurate and geometrically coherent line structures.

%% file: tables/analysis_plane.tex
\renewcommand{\arraystretch}{1.1}
\begin{table}[t]
\centering
\caption{Quantitative analysis on plane fitting strategies.}
\label{tab:analysis_plane}
\large
\resizebox{0.495\textwidth}{!}{
\begin{tabular}{cccc}
\toprule
\textbf{Plane Fitting Method} & \textbf{Precision↑} & \textbf{Recall↑} & \textbf{F-score↑} \\ 
\midrule
Direct Plane Fitting & 0.603 & 0.228 & 0.331 \\ 
Detection-guided Plane Fitting & 0.535 & 0.526 & 0.530 \\ 
\bottomrule
\end{tabular}
}
\end{table}
\renewcommand{\arraystretch}{1/1.1}

%% file: sec/05-3_line_detection.tex
\subsection{Multi-window, Multi-representation Line Detection}
\label{subsec:analysis_line}

\input{tables/analysis_line}
\input{tables/analysis_sensitivity}
\input{tables/analysis_velocity}

In the line detection stage, we propose a strategy to effectively apply mature frame-based line detectors to event data, termed multi-window, multi-representation (MWMR) line detection.
We perform multiple detections using various temporal windows and event-image representations, and subsequently merge the predicted lines to remove redundant lines.
This simple yet effective approach enables robust line detection from sparse and noisy event data.

We conduct an in-depth quantitative analysis of our MWMR line detection strategy to examine its process and effectiveness.
In~\Cref{tab:analysis_line}, we report the 2D line detection performance following the same evaluation procedure described in~\Cref{subsec:analysis_plane}.
In addition, we report the average number of detected lines per frame to assess the effect of multiple predictions and line merging.
We compare three models: (1) single window, single representation (averaged over all possible combinations), (2) multiple predictions without line merging, and (3) multiple predictions with line merging (our MWMR line detection).
Without line merging, combining all predictions significantly increases the number of detected lines.
While this improves recall and F-score, precision drops due to redundant detections.
After applying line merging, redundant short segments are suppressed and only the longest representative line is retained, resulting in fewer lines and a higher overall F-score, demonstrating the effectiveness of the merging strategy.

We also perform a sensitivity analysis on the temporal window size and image representation selection.
As shown in~\Cref{tab:analysis_sensitivity}, the performance remains stable when varying the temporal window size by up to five times, indicating robustness to this parameter.
A tenfold change causes moderate degradation, suggesting that employing a wider range of window sizes could further enhance robustness.
Similarly, when replacing our selected image representations~\cite{cohen2018spatial, park2016performance} with either TOS~\cite{glover2021luvharris} or SITS~\cite{manderscheid2019speed, acin2023vk}, the results show minimal variation, confirming the robustness of the MWMR design to the choice of image representation.
Additionally, we compare the MWMR with velocity-invariant representations~\cite{glover2021luvharris, manderscheid2019speed, acin2023vk, glover2024edopt}.
As seen in~\Cref{tab:analysis_velocity}, our MWMR line detection achieves superior performance compared with TOS and SITS, particularly in recall.
We attribute the lower recall of velocity-invariant representations to the noise frequently introduced during their processing.
However, when TOS or SITS was integrated into our MWMR framework along with other image representations, the performance improved, as shown in~\Cref{tab:analysis_sensitivity}.

%% file: tables/analysis_line.tex
\renewcommand{\arraystretch}{1.1}
\begin{table}[t]
\centering
\caption{Quantitative analysis across stages of multi-window, multi-representation line detection. When Multiple Prediction is not used, we report the average performance over all predictions.}
\label{tab:analysis_line}
\large
\resizebox{0.495\textwidth}{!}{
\begin{tabular}{cccccc}
\toprule
\makecell{\textbf{Multiple}\\\textbf{Prediction}} & 
\makecell{\textbf{Line}\\\textbf{Merging}} & 
\textbf{\#Line} & 
\textbf{Precision↑} & 
\textbf{Recall↑} & 
\textbf{F-score↑} \\ 
\midrule
- & - & 31  & 0.612 & 0.344 & 0.434 \\ 
\cmark & - & 185 & 0.359 & 0.758 & 0.487 \\ 
\cmark & \cmark & 55  & 0.530 & 0.523 & 0.526 \\ 
\bottomrule
\end{tabular}
}
\end{table}
\renewcommand{\arraystretch}{1/1.1}

%% file: tables/analysis_sensitivity.tex
\renewcommand{\arraystretch}{1.25}
\begin{table}[t]
\centering
\caption{Sensitivity analysis of MWMR line detection with respect to temporal window size and image representations: 
TOS~\cite{glover2021luvharris} and SITS~\cite{manderscheid2019speed, acin2023vk}.}
\label{tab:analysis_sensitivity}
\resizebox{0.43\textwidth}{!}{
\begin{tabular}{cc|cc}
\hline
\textbf{Window Size} & \textbf{F-score↑} & \textbf{Image Rep.} & \textbf{F-score↑} \\ 
\hline
1/10 & 0.368 & \cite{cohen2018spatial}+TOS & 0.502 \\ 
1/5  & 0.523 & \cite{park2016performance}+TOS & 0.523 \\ 
1    & 0.526 & \cite{cohen2018spatial}+\cite{park2016performance} & 0.526 \\ 
5    & 0.484 & \cite{cohen2018spatial}+SITS & 0.499 \\ 
10   & 0.443 & \cite{park2016performance}+SITS & 0.496 \\ 
\hline
\end{tabular}
}
\end{table}
\renewcommand{\arraystretch}{0.8}

%% file: tables/analysis_velocity.tex
\renewcommand{\arraystretch}{1.1}
\begin{table}[t]
\centering
\caption{Comparison of velocity-invariant representations and MWMR line detection.}
\label{tab:analysis_velocity}
\resizebox{0.48\textwidth}{!}{
\begin{tabular}{ccccc}
\toprule
\textbf{Image Rep.} & \textbf{\#Line} & \textbf{Precision↑} & \textbf{Recall↑} & \textbf{F-score↑} \\ 
\midrule
TOS & 42 & 0.527 & 0.380 & 0.441 \\ 
SITS & 62 & 0.443 & 0.412 & 0.427 \\ 
MWMR & 55 & 0.530 & 0.523 & 0.526 \\ 
\bottomrule
\end{tabular}
}
\end{table}
\renewcommand{\arraystretch}{1/1.1}

%% file: sec/05-4_runtime.tex
\subsection{Runtime Analysis}
\label{subsec:analysis_runtime}

\input{tables/analysis_runtime}

We conduct a detailed runtime analysis of the main stages in our pipeline in~\Cref{tab:analysis_runtime}, although our method operates offline.
The measurements are performed on the \texttt{office4} sequence of the synthetic dataset which has $1200 \times 680$ resolution.
Runtime is measured on a workstation equipped with an AMD Ryzen 9 3900X CPU (12 cores, 3.8 GHz) and an NVIDIA GeForce RTX 3090 GPU.
Since longer sequences naturally require more processing time, we report the average runtime per frame by dividing the total runtime by the number of frames.
Here, ‘per frame’ denotes the processing time for all events within a single frame (pose) interval.
As seen in~\Cref{tab:analysis_runtime}, the local line matching stage is highly efficient, showing the effectiveness of our mutual nearest neighbor-based approach for short-term associations.
Despite using a multi-window, multi-representation strategy, the line detection stage remains efficient thanks to the use of a lightweight detector~\cite{elsed}.
In contrast, the global line matching stage incurs the largest computational cost.
We believe that employing faster matchers could further improve efficiency in future work.
The runtime of the plane fitting stage is approximately proportional to the number of input events.
Thus, effective event downsampling is expected to reduce computational cost.

%% file: tables/analysis_runtime.tex
\renewcommand{\arraystretch}{1.1}
\begin{table}[t]
\centering
\caption{Runtime analysis. The reported values indicate the processing time per frame (in seconds).}
\label{tab:analysis_runtime}
\large
\resizebox{0.24\textwidth}{!}{
\begin{tabular}{cc}
\toprule
\textbf{Stage} & \textbf{Time (s)} \\ 
\midrule
Line detection   & 0.064 \\
Plane fitting    & 0.358 \\
Local matching   & 0.001 \\
Global matching  & 0.663 \\
Triangulation    & 0.120 \\
Optimization     & 0.351 \\
\bottomrule
\end{tabular}
}
\end{table}
\renewcommand{\arraystretch}{1/1.1}

%% file: sec/06_limitation.tex
\section{Limitation and Future Work}
\label{sec:limitation}

\begin{figure}[t]
    \centering
    \includegraphics[width=0.95\linewidth]{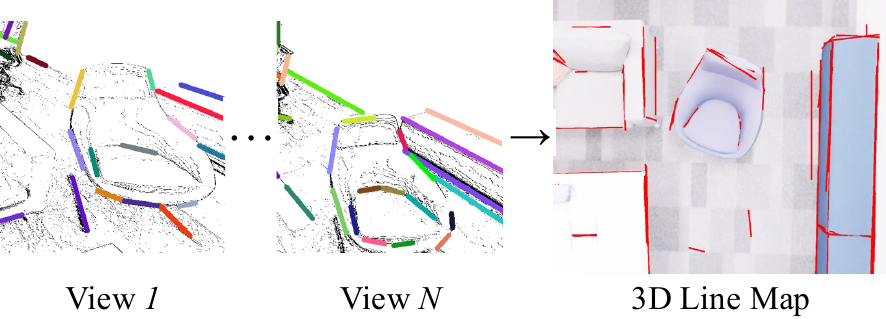}
    \caption{Example of our method applied to a curved-shape chair. The 2D line detection results from two viewpoints and the reconstructed 3D line map are shown.}
    \label{fig:limitation_curved}
\end{figure}

Although our method reconstructs compact and accurate 3D line maps from event data, it has some limitations.
We provide an in-depth analysis of how our current pipeline behaves under identified limitations.
Furthermore, we discuss possible directions to address these limitations and improve the overall system in future work.

First, the current pipeline operates offline.
Applying it to real-time scenarios would require further research on algorithmic efficiency and incremental optimization.
As analyzed in~\Cref{subsec:analysis_runtime}, developing more efficient matchers for global line matching would be an important direction for future work.
The plane parameters obtained from the space–time plane fitting stage could provide rich geometric cues and potentially serve as geometric descriptors for efficient line matching.

Second, line-based representations inherently struggle to represent 3D curved edges.
Extending event-based 3D mapping to support general edge structures, including curves, would be an interesting direction for future work.
For this purpose, we analyze how our method performs on curved surfaces to examine its current limitations.
\Cref{fig:limitation_curved} presents an example of our method applied to a chair with curved structures, showing the 2D line detection results from two viewpoints and the final 3D line reconstruction.
As illustrated, our method tends to approximate portions of curved boundaries using
multiple short line segments.
Among these short segments, only a subset is successfully reconstructed in the 3D line map, while many are discarded due to insufficient multi-view consistency, which is an
inherent limitation when approximating curved edges with straight lines.
We believe this can be addressed in future work by extending our representation to more general curve models such as 3D Bézier curves, superquadric contours~\cite{choi20243doodle}, which can capture smooth curvature while maintaining geometric consistency across views.

Third, in the space–time plane fitting, the planarity assumption may not strictly hold under highly non-linear motion or rapid rotations.
If plane fitting fails and only the initial line detections and line matches are available, the line triangulation and the Grassmann cost for line observations can still be applied.
However, the event-level Grassmann cost cannot be directly used because the inlier event association becomes unreliable.
To address this, we think that two possible directions can be considered.
First, a piecewise-planar fitting strategy could be adopted, where the temporal dimension is divided into several regions, and plane fitting is performed for each segment while maintaining connectivity across adjacent planes.
This approach could partially handle non-planar motion.
Second, inspired by the eventail~\cite{gao20235, gao2024n}, which explicitly models non-planar event traces, more advanced inlier event association methods could be developed to handle complex motion patterns.

%% file: sec/07_conclusion.tex
\section{Conclusion}
\label{sec:conclusion}

We present RoEL, a robust pipeline for reconstructing 3D line maps from event data.
To overcome noise sensitivity of prior event-based mapping methods that directly use all events, we leverage line correspondences to enable noise-robust 3D line reconstruction.
Since extracting line correspondences from event data is inherently challenging due to its sparse, asynchronous, and noisy nature, we present a series of event-specific techniques for robust and accurate correspondence search.
In addition, we introduce geometrically accurate cost functions defined on the Grassmann manifold to reconstruct 3D line maps by jointly utilizing multi-view line and event observations.

Our resulting 3D line maps accurately and efficiently represent events in indoor environments, where the majority of events correspond to line structures.
Extensive evaluation on both newly generated synthetic datasets and real-world sequences demonstrates that our line maps are robust and compact yet complete compared to existing event-based mapping methods.
Furthermore, RoEL outperforms a line mapping approach that does not account for the event modality, producing more complete and precise 3D reconstructions.
We also show that our reconstructed line maps not only enable accurate reconstruction and pose refinement but also serve as effective mid-level representations in downstream applications such as cross-modal registration and panoramic localization, thereby demonstrating their practical applicability.